\documentclass[10pt,twocolumn,letterpaper]{article}

\usepackage[pagenumbers]{cvpr} 

\definecolor{cvprblue}{rgb}{0.21,0.49,0.74}
\usepackage[pagebackref,breaklinks,colorlinks,allcolors=cvprblue]{hyperref}

\usepackage[accsupp]{axessibility}
\usepackage{algorithm}
\usepackage{booktabs}       
\usepackage{multirow} 
\usepackage{amsfonts}       
\usepackage{amsmath}        
\usepackage{tikz}

\usepackage[table]{xcolor}
\usepackage{soul}
\usepackage[noend]{algpseudocode}

\usepackage{capt-of} 

\usepackage{siunitx}
\usepackage{colortbl}

\newcommand{\smallfont}{\fontsize{7pt}{8pt}\selectfont}

\newcommand{\imgwidth}{0.135\linewidth}

\def\paperID{29833} 
\def\confName{CVPR}
\def\confYear{2026}

\title{SGI: Structured 2D Gaussians for Efficient and Compact \\ Large Image Representation}

\newcommand*\samethanks[1][\value{footnote}]{\footnotemark[#1]}
\author{
Zixuan Pan$^1$\thanks{Equal contribution.}, Kaiyuan Tang$^1$\samethanks, Jun Xia$^1$, Yifan Qin$^1$, \\ Lin Gu$^2$,  Chaoli Wang$^1$, Jianxu Chen$^3$, Yiyu Shi$^1$\\
$^1$University of Notre Dame \qquad $^2$Tohoku University \qquad \\ $^3$Leibniz-Institut für Analytische Wissenschaften – ISAS – e.V.\\
}

\begin{document}
\maketitle
\begin{abstract}
2D Gaussian Splatting has emerged as a novel image representation technique that can support efficient rendering on low-end devices.
However, scaling to high-resolution images requires optimizing and storing millions of unstructured Gaussian primitives independently, leading to slow convergence and redundant parameters.
To address this, we propose structured Gaussian image (SGI), a compact and efficient framework for representing high-resolution images. 
SGI decomposes a complex image into multi-scale local spaces defined by a set of seeds. Each seed corresponds to a spatially coherent region and, together with lightweight multi-layer perceptrons (MLPs), generates structured implicit 2D neural Gaussians. This seed-based formulation imposes structural regularity on otherwise unstructured Gaussian primitives, which facilitates entropy-based compression at the seed level to reduce the total storage.
However, optimizing seed parameters directly on high-resolution images is a challenging and non-trivial task. 
Therefore, we designed a multi-scale fitting strategy that refines the seed representation in a coarse-to-fine manner, substantially accelerating convergence. 
Quantitative and qualitative evaluations demonstrate that SGI achieves up to 7.5$\times$ compression over prior non-quantized 2D Gaussian methods and 1.6$\times$ over quantized ones, while also delivering 1.6$\times$ and 6.5$\times$ faster optimization, respectively, without degrading, and often improving, image fidelity.
Code is available at \url{https://github.com/zx-pan/SGI}.
\end{abstract}    
\section{Introduction}

\begin{figure}[!h]
\centering
\includegraphics[width=\linewidth]{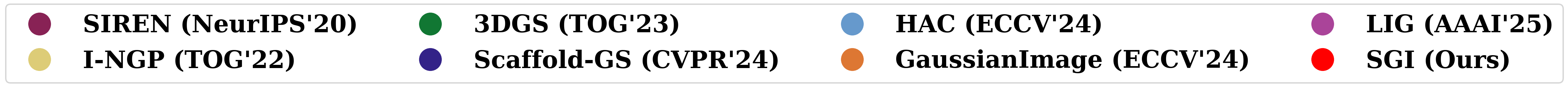}
\hspace*{-0.05\linewidth}
\begin{tabular}{c@{\hspace{0.05in}}c}
    \includegraphics[height=1.2in]{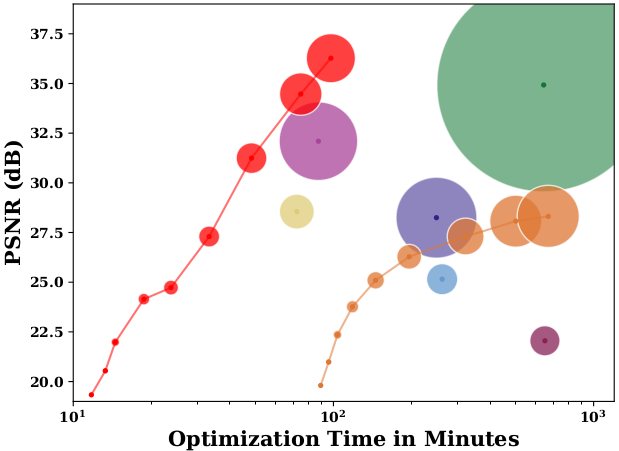} &
    \includegraphics[height=1.2in]{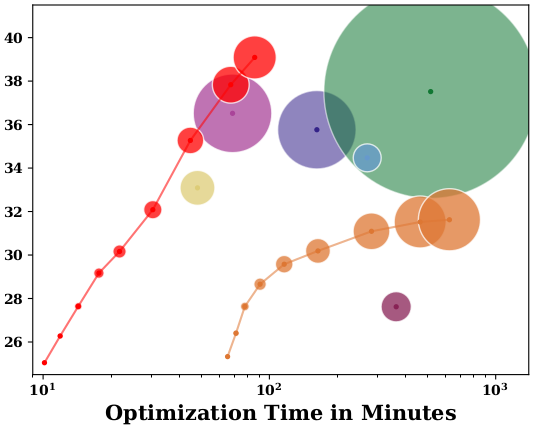} \\
    \footnotesize (a) FGF2 & \footnotesize (b) ICB \\
\end{tabular}
\vspace{-0.1in}
\caption{
Image representation results on the FGF2~\cite{Li-IAC15} and ICB~\cite{image-dataset} datasets. The x-axis (log scale) denotes optimization time in minutes, and the y-axis shows PSNR (dB). Each point represents a specific model configuration, with the area of the circle indicating its storage size. For GaussianImage and our SGI, we plot performance curves obtained by varying the number of Gaussian primitives.
Our SGI consistently achieves a favorable trade-off between fidelity, compactness, and optimization time.
}
\label{fig:lineplots}
\vspace{-0.2in}
\end{figure}

Image representation is a fundamental problem in computer vision, with wide-ranging applications such as compression~\cite{Guo-NIPS23}, editing~\cite{Zhang-NIPS23}, and super-resolution~\cite{Yang-NIPS21}.
Traditional image representations, such as grid graphics or transform-based methods (e.g., discrete cosine transform~\cite{Ahmed-TOC06} or wavelet transform~\cite{Heil-SIAM89}), are often limited in their ability to model visual signals with high fidelity.
Recent work has explored implicit neural representations (INRs)~\cite{Sitzmann-NeurIPS20, Saragadam-CVPR23}, which encode images using multi-layer perceptrons (MLPs) that learn the mapping from pixel coordinates to pixel values. 
Although INRs offer continuous, resolution-independent modeling, representing large, high-resolution images with INRs remains challenging, as faithfully capturing fine spatial details typically requires deep MLPs. 
This results in significant computation and memory overhead when processing millions of pixels, leading to slow encoding/decoding and limited feasibility on low-end devices. 
To address this issue, recent methods based on 2D Gaussian Splatting~\cite{Zhang-ECCV24, Zhu-AAAI25} represent images as sets of explicit Gaussian primitives. 
This formulation enables fast encoding and decoding without requiring slow network inference. 
Moreover, the memory-efficient splatting rasterization enables high-resolution optimization even on resource-constrained devices.

Despite the efficiency of 2D Gaussian splatting, existing 2D Gaussian representation methods~\cite{Zhang-ECCV24,Zhu-AAAI25} optimize each Gaussian independently without leveraging spatial locality, the property that nearby pixels tend to share similar colors, textures, and structures, resulting in significant parameter redundancy across adjacent primitives. This issue becomes more severe in high-resolution settings, where millions of Gaussians are required to cover the spatial content, leading to (1) \emph{large model sizes} due to storing vast numbers of primitives, and (2) \emph{substantial optimization overhead}, especially when quantization-aware fine-tuning is needed for compression~\cite{Zhang-ECCV24}.

To address these limitations, we present structured Gaussian image (SGI), a compact representation tailored for high-resolution image modeling. SGI partitions the image into a collection of multi-scale local regions, each associated with a seed point. For every seed, a pair of lightweight MLPs predicts the attributes of the Gaussian primitives within its region, converting the formerly unstructured set of Gaussians into a coherent, seed-organized representation that preserves local details. 
The structural regularity introduced by seed-based 2D neural Gaussians allows us to further compress residual spatial redundancy at the seed level.
Specifically, a context model composed of a binary hash grid estimates explicit distributions of seed attributes, enabling adaptive bit allocation and effectively compressing residual spatial redundancy.
While seed-based compression alleviates the model-size issue, directly optimizing seed parameters at full resolution, however, is difficult and computationally costly. The adaptive bit allocation also further exacerbates the challenge of long optimization time. To mitigate this, we adopt a coarse-to-fine multi-scale fitting strategy, where SGI is first optimized on a low-resolution approximation and then progressively refined at higher resolutions. This hierarchical optimization significantly improves both convergence speed and stability.

As shown in Figure~\ref{fig:lineplots}, SGI provides consistently better trade-offs between fidelity, compactness, and optimization efficiency compared to prior 2D Gaussian and INR-based methods.
Our contributions are summarized as follows:
\begin{itemize}
\item We propose the first structured 2D Gaussian representation for high-resolution images by introducing seed-based 2D neural Gaussians and a context-guided entropy coding scheme, enabling effective elimination of spatial redundancy and significant model size reduction.
\item We develop a multi-scale fitting strategy that enables coarse-to-fine optimization, substantially reducing optimization time while improving the reconstruction quality.
\item Extensive experiments on megapixel-scale datasets show that SGI achieves up to 7.5× compression over non-quantized 2D Gaussian baseline methods and 1.6× over quantized ones, while providing 1.6× to 6.5× faster optimization and maintaining or improving image fidelity.
\end{itemize}
\section{Related work}
\label{related_work}

{\bf Implicit Neural Representation.}
By utilizing a neural network to fit continuous mappings from coordinates to signal values, INR~\cite{Sitzmann-NeurIPS20, Saragadam-CVPR23} has demonstrated its effectiveness in various types of signal representations, including 3D scenes~\cite{Mildenhall-ECCV20, Barron-ICCV21, Barron-ICCV23}, 3D data~\cite{Sitzmann-MetaSDF, Han-TVCG22, Tang-PVIS24}, 2D images~\cite{Emilien-arXiv21, Chen-CVPR21, Emilien-TMLR22}, and videos~\cite{Chen-NeurIPS21, Chen-CVPR23, Zhao-CVPR23}.
For image representation, vanilla INRs~\cite{Sitzmann-NeurIPS20} typically optimize an MLP to approximate the pixel-wise RGB values.
However, for high-resolution images, such INRs must repeatedly forward pass the MLP numerous times to reconstruct the full image, which is computationally expensive.
Recent grid-based INRs~\cite{Muller-TOG22, Chen-ECCV22, Hu-ICCV23} move most network parameters to a parameteric grid, allowing efficient spatial information encoding and sampling.
Nevertheless, as the grid resolution increases, the number of parameters in grid-based INRs grows rapidly, leading to substantial GPU memory consumption.
Although some previous works introduce hierarchical structures~\cite{Martel-TOG21, Saragadam-ECCV22} to achieve efficient encoding, they still suffer from the slow feedforward process of the MLP network. To further improve efficiency, recent work has shifted from neural fields to point-based representations.

\noindent{\bf Differentiable Point-based Representation.}
Differentiable point-based representations~\cite{Kerbl-ToG23, Tang-TVCG25, Yang-CVPR24, Huang-SIGGRAPH24, Fan-NIPS24, Zhang-ECCV24, Zhang-SIGGRAPH25, Tang-AAAI25, Huang-AAAI25} have recently been widely studied for their ability to efficiently and flexibly capture intricate structural details.
Since Kerbl et al.~\cite{Kerbl-ToG23} demonstrated the capability of 3D Gaussian representation in scene reconstruction, point-based representation has emerged as a trending topic.
Extensive efforts have been made to extend the original 3DGS algorithm for image representation.
GaussianImage~\cite{Zhang-ECCV24} adapts Gaussian points to image space with fewer attribute parameters and optimizes weighted color attributes to approximate the alpha blending results, thereby reducing computation complexity.
After quantization-aware optimization, the compact 2D Gaussian representation can achieve high-fidelity image representation with remarkable decoding speed.
Following GaussianImage, other works have explored the applications of 2D Gaussian representation to realize arbitrary-scale image super-resolution~\cite{Peng-arXiv25, Hu-AAAI25, Chen-CVIU}, realistic image editing~\cite{Waczy-arXiv24}, and video representation~\cite{Smolak-arXiv24, Liu-arXiv25, Bond-arXiv25}.
Image-GS~\cite{Zhang-SIGGRAPH25} proposes a content-adaptive image representation for better primitive allocation. 
In the context of large image representation, LIG~\cite{Zhu-AAAI25} introduces a Level-of-Gaussian hierarchy to improve high-frequency fitting by sequentially refining residuals with additional Gaussians.

\begin{figure*}[tb]
\centering
\includegraphics[width=\linewidth]{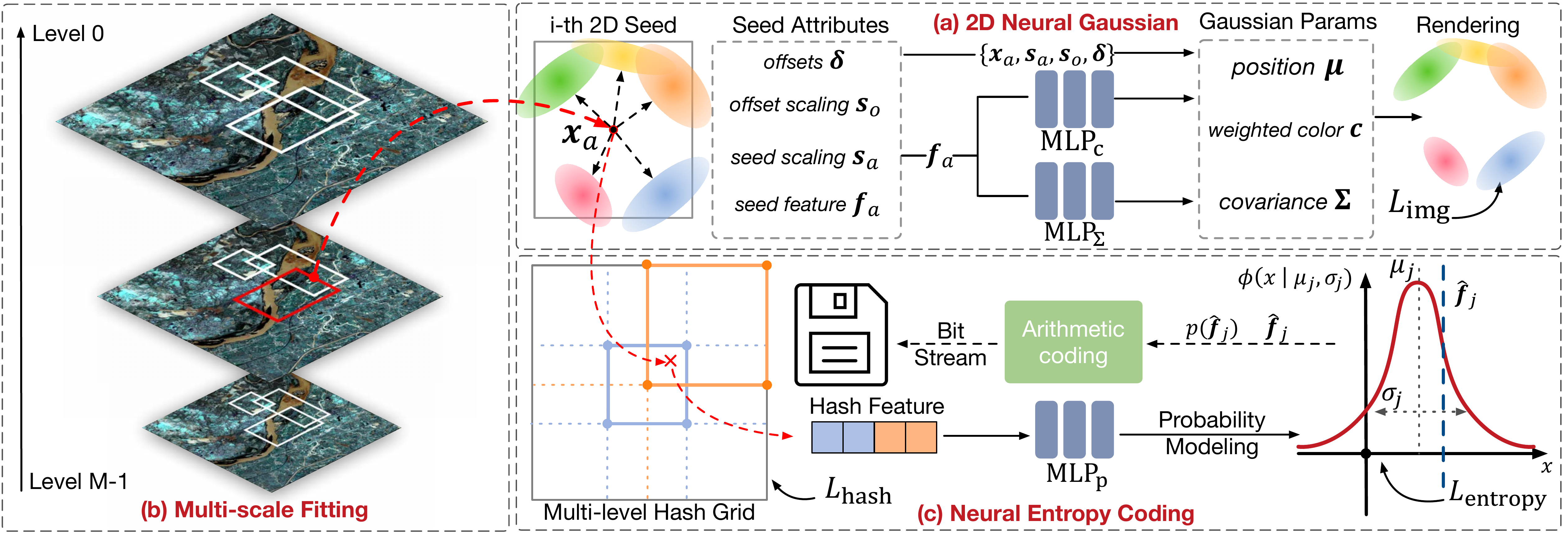}

\caption{
{\bf The overall pipeline of SGI.} We first introduce {\bf (a) seed-based 2D neural Gaussians}, where each seed predicts a group of 2D Gaussian primitives via two shared MLPs for decoding color and covariance. To accelerate optimization, we adopt a {\bf (b) multi-scale fitting} strategy that progressively refines the representation from coarse to fine using a Gaussian pyramid. Finally, we leverage {\bf (c) neural entropy coding} to further compress the explicit seed attributes for compact representation.
}
\vspace{-.15in}

\label{fig:workflow}
\end{figure*}

While achieving promising performance, these methods have limitations in exploiting spatial redundancy. 
Recent anchor-based frameworks~\cite{Lu-CVPR24, Cho-arXiv24, Zhang-arXiv25} have demonstrated remarkable efficiency gains in 3D scene representations to reduce this redundancy.
However, we empirically found that directly applying these 3D anchor representations to 2D image modeling does not yield comparable compression benefits.
In contrast, we design 2D seed points to provide structural regularity for our \emph{seed-based 2D neural Gaussians}, thereby establishing a compact, organized representation and enabling advanced neural entropy coding.
By integrating a context model~\cite{Chen-ECCV24, Wang-NeurIPS24, Chen-CVPR24}, our SGI encodes seed parameters based on their information entropy, leading to a significantly more compact image representation compared to other 2D Gaussian-based approaches. 
Unlike the Level-of-Gaussian strategy in LIG~\cite{Zhu-AAAI25}, which uses a subset of Gaussians solely for residual fitting, our \emph{multi-scale fitting} strategy employs the full representation in a hierarchical manner, ensuring that optimization acceleration and representation compression do not come at the cost of reconstruction fidelity. Compared to 3DGS-based methods such as Scaffold-GS~\cite{Lu-CVPR24} and HAC~\cite{Chen-ECCV24}, SGI achieves significantly faster optimization and lower GPU memory usage, owing to its use of  {\em seed-based 2D neural Gaussians} and context model-driven {\em entropy coding}. 
In addition, the proposed \emph{multi-scale fitting} strategy mitigates the extended optimization time typically due to the large number of parameters and the overhead of entropy estimation~\cite{Chen-ECCV24, Wang-NeurIPS24, Chen-CVPR24}.
Further comparisons are provided in Section~\ref{sec:comparison}.

\section{Methodology}
Figure~\ref{fig:workflow} shows the pipeline of SGI.
Our objective is to encode the high-resolution image with a set of seed-based 2D neural Gaussians (Section~\ref{sec:neural_gaussian}).
The structural regularity introduced by the seed-based representation allows us to extract a compact representation using neural entropy coding (Section~\ref{sec:entropy_coding}).
These seed attributes are optimized using a multi-scale strategy to improve convergence speed (Section~\ref{sec:multi_scale_fitting}). 
Finally, we describe how the seed parameters are encoded and decoded using entropy coding (Section~\ref{sec:bistream}).

\subsection{Seed-based 2D Neural Gaussians}
\label{sec:neural_gaussian}

While 2D Gaussian representations have shown strong potential for image representation~\cite{Zhang-ECCV24,Zhu-AAAI25}, existing methods treat each Gaussian independently, thereby overlooking spatial locality among Gaussian primitives and leading to significant parameter redundancy.
To address this limitation, we try to introduce \emph{seed-based 2D neural Gaussians} into the 2D image representation domain for the first time, drawing inspiration from recent progress in {\em anchor-centered} methods for 3D reconstruction~\cite{Lu-CVPR24, Cho-arXiv24, Zhang-arXiv25}. This approach aims to exploit the spatial locality by grouping Gaussians under a set of seeds and predicting their attributes through lightweight MLPs, rather than storing them directly. We introduce the detailed implementation of \emph{seed-based 2D neural Gaussians} below.

Given a predefined number of seeds $N$, we initialize them to uniformly cover the image. Each seed located at position $\boldsymbol{x_a} \in \mathbb{R}^2$ is associated with a set of attributes:
\begin{equation}
\mathcal{A} = \left\{ \boldsymbol{f_a} \in \mathbb{R}^D, \; \boldsymbol{s_o} \in \mathbb{R}^2, \; \boldsymbol{s_a} \in \mathbb{R}^2, \; \boldsymbol{\delta} \in \mathbb{R}^{K \times 2} \right\},
\end{equation}
where $\boldsymbol{f_a}$ is the seed feature, $\boldsymbol{\delta}$ contains learned offsets for $K$ associated Gaussians, $\boldsymbol{s_o}$ and $\boldsymbol{s_a}$ are per-seed scaling factors for offsets and scalings of the associated Gaussians.
The positions of these Gaussians are computed as:
\begin{equation}
\left\{ \boldsymbol{\mu}^{(k)} \right\}_{k=0}^{K-1} = \boldsymbol{x_a} + \left\{ \boldsymbol{\delta}^{(k)} \right\}_{k=0}^{K-1} \cdot \boldsymbol{s_o}.
\end{equation}
We then decode the opacity-weighted color coefficients $\mathbf{c'}  \in \mathbb{R}^3$ and covariance matrix $\boldsymbol{\Sigma} \in \mathbb{R}^{2 \times 2}$ of each Gaussian from $\boldsymbol{f_a}$ using two MLPs: 
$\text{MLP}_c$ outputs the colors, and $\text{MLP}_\Sigma$ predicts base scales $\boldsymbol{s_\text{base}}$ and rotation angles $\theta$.
The final scale $\boldsymbol{s} \in \mathbb{R}^2$ of each Gaussian used for rendering is computed by $\boldsymbol{s} = \boldsymbol{s_\text{base}} \cdot \boldsymbol{s_a}$.
The covariance matrix $\boldsymbol{\Sigma}$ is obtained by constructing a positive semidefinite matrix via:
\begin{equation}
\boldsymbol{\Sigma} = \mathbf{R} \mathbf{S} \mathbf{S}^\top\mathbf{R}^\top,
\end{equation}
\begin{equation}
\mathbf{R}(\theta) =
\begin{bmatrix}
\cos(\theta) & -\sin(\theta) \\
\sin(\theta) & \cos(\theta)
\end{bmatrix}, \quad
\mathbf{S} = 
\begin{bmatrix}
s_1 & 0 \\
0 & s_2
\end{bmatrix}.
\end{equation}

Finally, the rendered pixel color is calculated by using all Gaussians contribute to the pixel via an accumulated summation~\cite{Zhang-ECCV24}:
\begin{equation}
\boldsymbol{C} = \sum_{i \in I} \mathbf{c'}_i G_i(\mathbf{x}),
\label{eq:blend}
\end{equation}
where $I$ is the number of Gaussians contributing to the pixel, $\mathbf{c'}_i$ is an opacity-weighted color vector, and $G_i(\mathbf{x})$ is the spatial density of the $i$-th Gaussian at pixel $\mathbf{x}$, \mbox{defined as:} 
\begin{equation}
G(\mathbf{x}) = \exp\left(-\frac{1}{2}(\mathbf{x} - \boldsymbol{\mu})^\top \boldsymbol{\Sigma}^{-1}(\mathbf{x} - \boldsymbol{\mu})\right).
\label{gs_density}
\end{equation}
This formulation enables fully differentiable rasterization. 
By adopting and extending differentiable rasterization techniques from prior work~\cite{Kerbl-ToG23, Zhang-ECCV24}, our \emph{seed-based 2D neural Gaussians} achieves fast rendering and low memory consumption. 
Overall, the proposed \emph{seed-based 2D neural Gaussians} models an image using $N$ seed attributes $\{\mathcal{A}^{i}\}_0^{i=N-1}$ and two lightweight MLPs $\text{MLP}_c$ and $\text{MLP}_\Sigma$, which together predict 2D Gaussian primitives each defined by only three attributes (position, covariance, and weighted color), requiring just eight parameters per Gaussian.

\subsection{Entropy Coding with Context Model}
\label{sec:entropy_coding}
Theoretically, converting unstructured 2D Gaussians into seed-level attributes and shared MLPs should reduce the overall model size, similar to how Scaffold-GS~\cite{Lu-CVPR24} improves the storage efficiency of 3DGS~\cite{Kerbl-ToG23}.
However, this adaptation is far less straightforward in the 2D setting.
As shown in Table~\ref{table:GF2-comparison} and Table~\ref{table:ablation_lambda}, directly applying seed-based 2D neural Gaussians yields only about a 3\% reduction compared to pure 2D Gaussian representations (e.g., LIG~\cite{Zhu-AAAI25}).
This limited gain arises because 2D Gaussian formulations already eliminate several parameters, such as opacity, that dominate storage in 3D counterparts.
Nevertheless, the structural regularity introduced by seed-based neural Gaussians provides an organized representation that enables further compression by modeling and constraining the distribution of seed attributes, allowing fewer bits to be used during entropy coding.
Specifically, given quantized codes $\hat{y}$ and a probability model $p_{\hat{y}}(\hat{y})$, entropy coding techniques, such as arithmetic coding~\cite{Rissanen-TIT81}, can losslessly compress them. Thus, to compress the seed attributes via entropy coding, we need quantization to discretize the data and probability modeling to estimate symbol likelihoods.

\noindent{\bf Quantization.} 
Let $\boldsymbol{f}^{(i)}_j$ denote the $j$-th component of $i$-th seed's attributes $\mathcal{A}^{(i)}$.
Following~\cite{Chen-ECCV24,Pierre-ICLR21}, quantization is implemented using noise injection during training and rounding during testing:
\begin{align}
    \hat{\boldsymbol{f}}^{(i)}_j &= \boldsymbol{f}^{(i)}_j + \mathcal{U}\left(-\tfrac{1}{2}, \tfrac{1}{2}\right) \cdot q_j^{(i)},  \quad \text{for training} \\
             &= \operatorname{Round}(\boldsymbol{f}^{(i)}_j / q_j^{(i)}) \cdot q_j^{(i)}. \quad \text{for testing} \label{eq:quantization}
\end{align}
The quantization step size $q_j^{(i)}$ is computed as:
\begin{align}
    q_j^{(i)} &= Q_j \times \left(1 + \tanh(r_j^{(i)})\right),
\end{align}
where $Q_j$ is the quantization step size, $r_j^{(i)}$ is a refinement factor predicted by the context model in Eq.~\eqref{context_model_eq}.

\noindent{\bf Probability Modeling.}
Inspired by recent 3DGS work~\cite{Chen-ECCV24, Wang-NeurIPS24, Chen-CVPR24}, we employ a context model $\text{MLP}_p$ to estimate the distributions of seed attributes. Rather than directly conditioning on $\boldsymbol{f}^{(i)} \in \mathcal{A}^{(i)}$, we introduce a learnable binary hash grid $\mathcal{H}$~\cite{Shin-NIPS23} to capture the inherent spatial consistencies of the unorganized seeds for the context model:
\begin{align}
    \left\{ \mu_j^{(i)}, \sigma_j^{(i)}, r_j^{(i)} \right\}_{j=0}^{3} &= \text{MLP}_p(\mathcal{H}(\boldsymbol{x}_a^{(i)})), \label{context_model_eq} \\
    p(\hat{\boldsymbol{f}}_j^{(i)}) &= \int_{\hat{\boldsymbol{f}}_j^{(i)} - \frac{q_j}{2}}^{\hat{\boldsymbol{f}}_j^{(i)} + \frac{q_j}{2}} 
    \phi(x \mid \mu_j^{(i)}, \sigma_j^{(i)}) \, dx,
    \label{prob}
\end{align}
where $\phi$ is the probability density function of the Gaussian distribution, $j \in \{0,1,2,3\}$, $\hat{\boldsymbol{f}}_j^{(i)}$ is the quantized seed attribute for $\boldsymbol{f}_j^{(i)}$,
and $\mu_j^{(i)}$ and $\sigma_j^{(i)}$ are the predicted mean and standard deviation for each attribute dimension.
The context model and hash grid are trained by minimizing the entropy loss:
\begin{equation}
    L_{\text{entropy}} = \sum_{i=0}^{N-1}\left[ -\text{log}_2 \left[ \prod_{\boldsymbol{f}^{(i)} \in \{\mathcal{A}^{(i)}  \}} p(\hat{\boldsymbol{f}}^{(i)}) \right] \right].
    \label{eq:entropy}
\end{equation}
The hash grid is binarized to $\{-1, +1\}$ and optimized via straight-through estimation (STE)~\cite{Shin-NIPS23}, and its bit consumption is bounded by:
\begin{equation}
    L_{\text{hash}} = -n_1 \log_2(\frac{n_1}{n_1+n_0}) - n_0 \log_2(\frac{n_0}{n_1+n_0}),
    \label{eq:hash}
\end{equation}
where $n_1$ and $n_0$ are total counts of $+1$ and $-1$ in the hash grid, respectively.

\noindent{\bf Overall Loss.} The total optimization loss combines the rendering fidelity objective and bits consumption regularization:
\begin{equation}
    L = L_{\text{img}} + \frac{\lambda}{N \cdot d_\mathcal{A}}(L_{\text{entropy}} + L_{\text{hash}}),
    \label{eq:overall_loss}
\end{equation}
where $L_{\text{img}}$ is the $L_1$ loss between rendered and target images, $\lambda$ is a hyperparameter balancing the rate and fidelity, and $d_\mathcal{A}=D+4+2K$ is the total number of attribute dimensions for each seed.

\begin{table*}[!ht]
\vspace{-0.1in}
    \centering
    \caption{\textbf{Quantitative results on FGF2 and ICB.} Optimization time is in minutes, and model size is in megabytes (MB). To fit INR-based methods (SIREN and I-NGP) on these large-scale datasets under limited GPU memory, we adopt a patch-based training strategy that loads randomly sampled coordinate and pixel batches from CPU to GPU on the fly. 
    Our SGI is evaluated in two settings: \textbf{low-rate} (3.5M Gaussians) and \textbf{high-rate} (10M Gaussians), both demonstrating strong trade-offs between fidelity, compactness, and optimization efficiency. \sethlcolor{blue!20} \hl{Darker blue} and \sethlcolor{red!20} \hl{darker red} highlights indicate the best-performing method within the \sethlcolor{blue!8} \hl{low-rate} and \sethlcolor{red!8} \hl{high-rate} groups, respectively, in each metric. \emph{Reported numbers are averaged per image across the dataset.}}
    \small 
    \resizebox{\textwidth}{!}{
    \begin{tabular}{lcccS[table-format=3.2]S[table-format=3.2]cccS[table-format=3.2]S[table-format=3.2]}
        \toprule
    \multirow{2}{*}{Method} & \multicolumn{5}{c}{FGF2} &\multicolumn{5}{c}{ICB}\\
        \cmidrule(lr){2-6} \cmidrule(lr){7-11}
        
        & PSNR$\uparrow$ &SSIM$\uparrow$ &LPIPS$\downarrow$ &\multicolumn{1}{c}{Opt. Time$\downarrow$} &\multicolumn{1}{c}{Size$\downarrow$} & PSNR$\uparrow$ &SSIM$\uparrow$ &LPIPS$\downarrow$ &\multicolumn{1}{c}{Opt. Time$\downarrow$} &\multicolumn{1}{c}{Size$\downarrow$} 
      \\
        \midrule
        \rowcolor{blue!8} SIREN~\cite{Sitzmann-NeurIPS20} (NeurIPS'20)
        &22.05 &0.8020 &0.4831 &649.71 &\cellcolor{blue!20}{15.79}
        &27.62 &0.8600 &0.3817 &363.34 &15.79
        \\
        \rowcolor{blue!8} I-NGP~\cite{Muller-TOG22} (TOG'22)
        &28.55 &0.9592 &0.1043 &72.32 &21.07
        &33.09 &0.9532 &0.1176 &48.11 &21.07
        \\
        \rowcolor{blue!8} HAC~\cite{Chen-ECCV24} (ECCV'24)
        &25.15&0.9226&0.1767&261.69&16.78
        &34.47&0.9801&0.0688&270.57&13.52
        \\
        \rowcolor{blue!8} GaussianImage~\cite{Zhang-ECCV24} (ECCV'24) 
        &27.30 &0.9457 &0.1342 &322.17 &23.37
        &31.09 &0.9330 &0.1462 &282.61 &23.37
        \\
        \rowcolor{blue!8} Our SGI (low-rate)
        &\cellcolor{blue!20}{31.24} &\cellcolor{blue!20}{0.9863} &\cellcolor{blue!20}{0.0731} &\cellcolor{blue!20}{48.43} &16.33
        &\cellcolor{blue!20}{35.27} &\cellcolor{blue!20}{0.9853} &\cellcolor{blue!20}{0.0575} &\cellcolor{blue!20}{44.75} &\cellcolor{blue!20}{12.30}
        \\
        \midrule
        \rowcolor{red!8} 3DGS~\cite{Kerbl-ToG23} (TOG'23)
        &34.93 &0.9950 &0.0246 &642.85 &787.73
        &37.52 &0.9932 &0.0248 &515.99 &787.73
        \\
        \rowcolor{red!8} Scaffold-GS~\cite{Lu-CVPR24} (CVPR'24)
        &28.25 &0.9578 &0.0973 &248.83 &112.61
        &35.76 &0.9853 &0.0509 &162.11 &105.81
        \\
        
        \rowcolor{red!8} LIG~\cite{Zhu-AAAI25} (AAAI'25)
        &32.10 &0.9879 &0.0568 &\cellcolor{red!20}{87.56} &106.81
        &36.40 &0.9888 &0.0180 &\cellcolor{red!20}{68.73} &106.81
        \\
        \rowcolor{red!8} Our SGI (high-rate) 
        &\cellcolor{red!20}{36.27} &\cellcolor{red!20}{0.9961} &\cellcolor{red!20}{0.0162} &97.75 &\cellcolor{red!20}{41.74}
        &\cellcolor{red!20}{39.09} &\cellcolor{red!20}{0.9949} &\cellcolor{red!20}{0.0122} &86.11 &\cellcolor{red!20}{32.15}
        \\
        \bottomrule
    \end{tabular}
    }
    \label{table:GF2-comparison}
    \vspace{-0.05in}
\end{table*}

\subsection{Multi-scale Fitting}
\label{sec:multi_scale_fitting}

Training on large, high-resolution images can be computationally intensive, even for Gaussian-based models. Besides, the quantization-aware training and probability modeling for the entropy coding introduce extra computation overhead for training.
To accelerate optimization, we propose a \emph{multi-scale fitting} strategy that gradually refines the representation by using the solution from a coarser scale as a warm start for the next finer scale.
Given a target image $I$, we construct a Gaussian pyramid~\cite{Adelson-RCA84} with $M$ levels, producing a sequence of downsampled images $\{I_0 = I, I_1, \dots, I_{M-1}\}$ from fine to coarse, each downsampled by a factor of two.
At each level $l$, we maintain a set of seed positions and attributes $\mathbb{A}^{(l)} = \{\boldsymbol{x_a}^{(i,l)}, \mathcal{A}^{(i,l)} \}_{0}^{i=N-1}$ and parameters $\theta^{(l)}$ of $\text{MLP}_c$ and $\text{MLP}_{\Sigma}$. Starting from the coarsest level $l = M-1$, we optimize:
\begin{equation}
\mathbb{A}^{(l)}_*, \theta^{(l)}_* = \arg\min_{\mathbb{A}^{(l)}, \theta^{(l)}} \; L_{\text{img}}(I_l, \hat{I}_l(\mathcal{G}^{(l)}(\mathbb{A}^{(l)}, \theta^{(l)}))),
\end{equation}
where $\mathcal{G}^{(l)}(\mathbb{A}^{(l)}, \theta^{(l)})$ are the Gaussians inferred from seed parameters and MLPs, $\hat{I}_l$ is the rendered image at level $l$.
The optimized parameters are transferred to the next finer level with adaptation:
\begin{equation}
\mathbb{A}^{(l-1)} \leftarrow \phi_{\text{init}}(\mathbb{A}^{(l)}_*), \quad \theta^{(l-1)} \leftarrow \theta^{(l)}_*,
\label{eq:update_multi_scale}
\end{equation}
where $\phi_{\text{init}}(\cdot)$ adapts seed attributes for the higher-resolution image. In particular, seed positions and scales are scaled by a factor of two to account for resolution doubling.
This hierarchical fitting process continues iteratively until the finest level $l = 0$ is reached.

\subsection{Bitstream Generation and Decoding}
\label{sec:bistream}
The encoding and decoding process involves the seed positions and attributes $\{\boldsymbol{x}_a^{(i)}, \mathcal{A}^{(i)}\}_{i=0}^{N-1}$, as well as the binary hash grid $\mathcal{H}$. The seed positions are compressed using geometric point cloud compression (GPCC)~\cite{Chen-GetMobile23}, while the hash grid is encoded using arithmetic coding.
The seed attributes $\{\mathcal{A}^{(i)}\}_{i=0}^{N-1}$ are also entropy-coded via arithmetic coding, with probabilities for each component estimated by the context model defined in Eq.~\eqref{context_model_eq} and Eq.~\eqref{prob}.
In the end, we store the encoded seed parameters $\{\boldsymbol{x}_a^{(i)}, \mathcal{A}^{(i)}\}_{i=0}^{N-1}$, the binary hash grid $\mathcal{H}$, two MLPs ($\text{MLP}_c$ and $\text{MLP}_{\Sigma}$) used for decoding neural Gaussians, and the context model $\text{MLP}_p$.

At decoding time, the seed positions and hash grid are first decoded. The context model (with hash-based spatial features) is then used to reconstruct the probability distribution for each attribute component, which guides the arithmetic decoder in recovering the quantized seed attributes. These are subsequently passed through MLPs to reconstruct the full set of Gaussian parameters for rendering.

\section{Experiments}
\label{Experiments}

\begin{figure*}[tb]
 \begin{center}
 $\begin{array}{c@{\hspace{0.025in}}c@{\hspace{0.025in}}c@{\hspace{0.025in}}c@{\hspace{0.025in}}c@{\hspace{0.025in}}c@{\hspace{0.025in}}c}
 \includegraphics[width=\imgwidth]{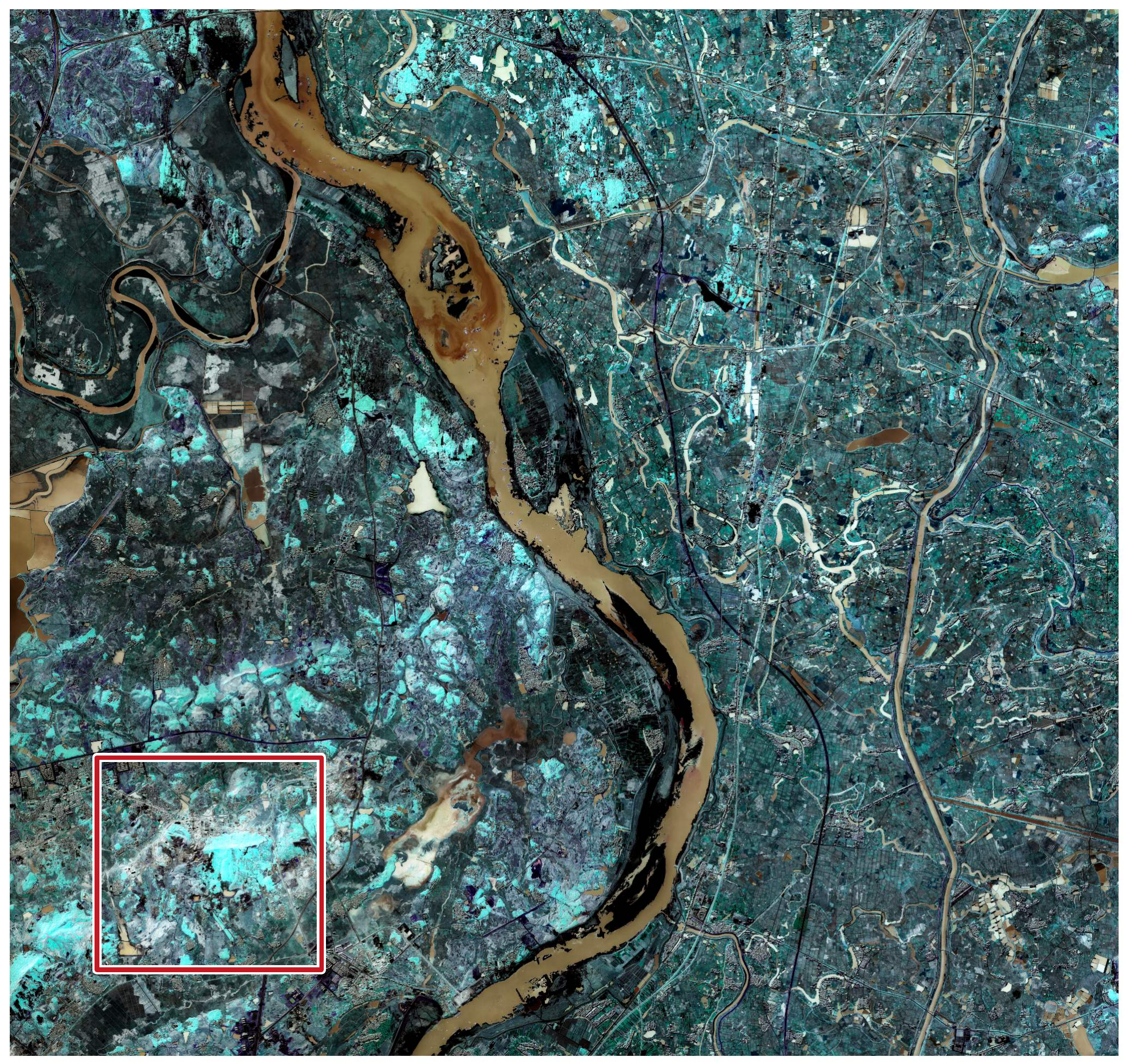}&
 \includegraphics[width=\imgwidth]{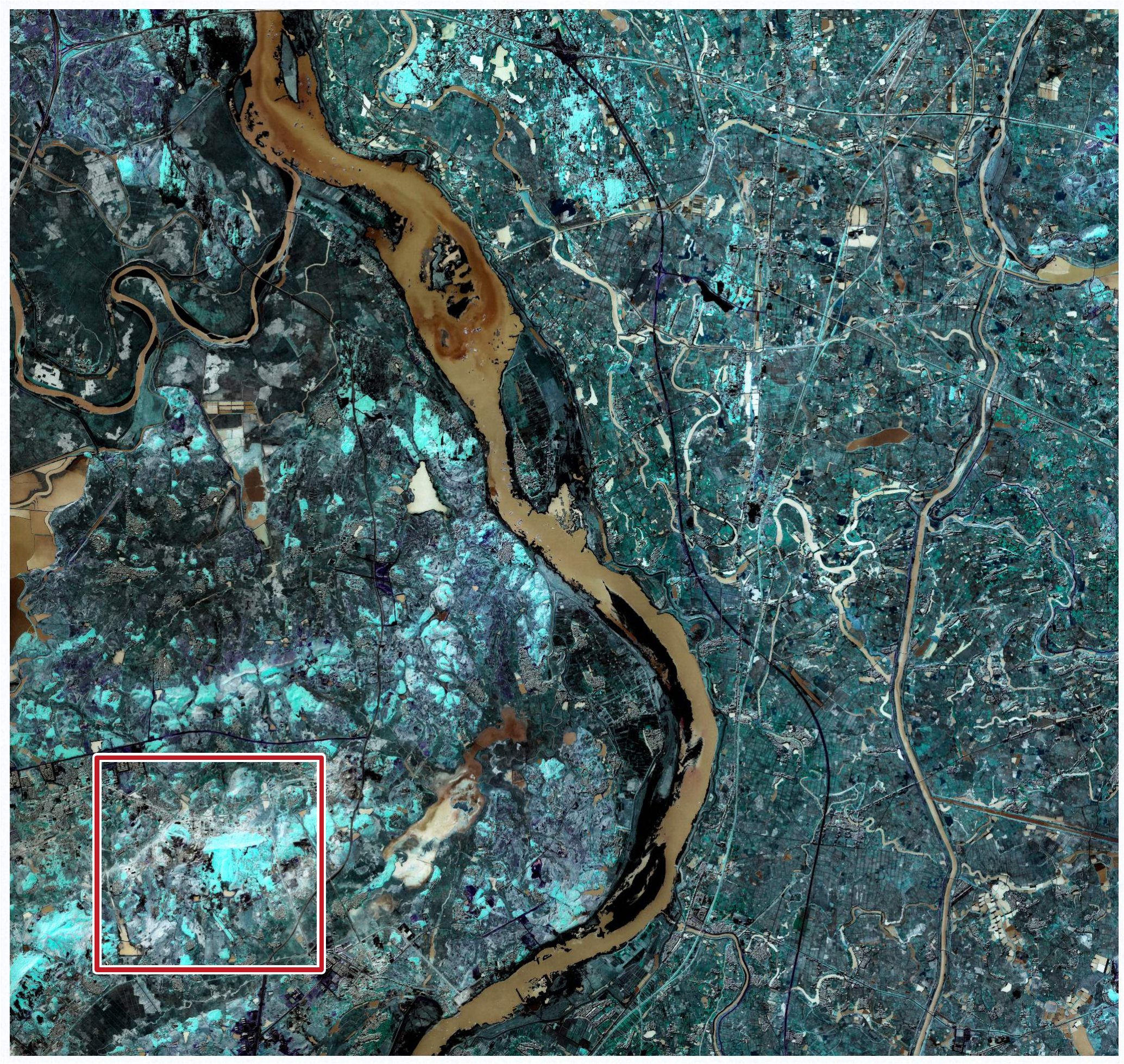}&
 \includegraphics[width=\imgwidth]{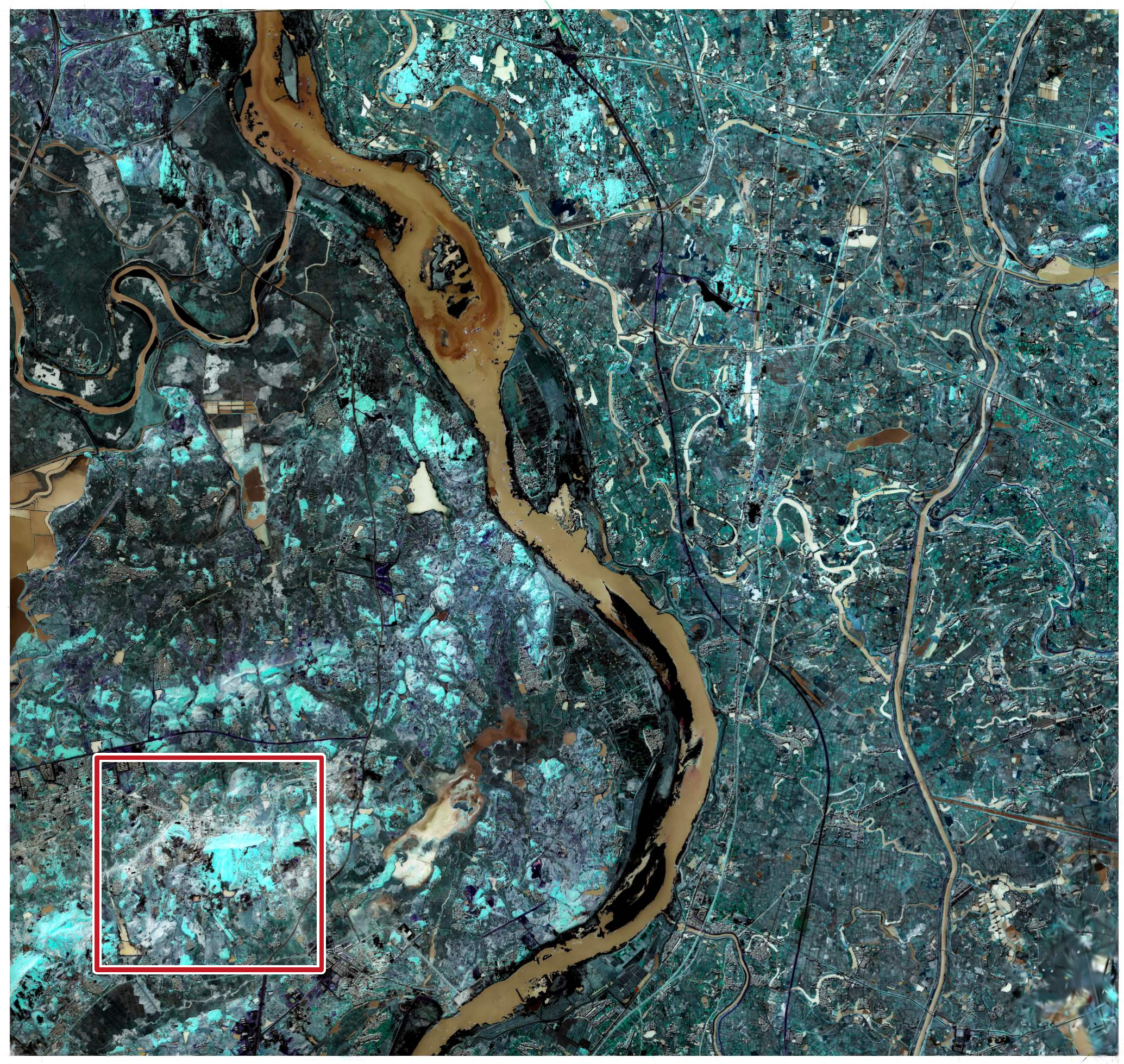}&
  \includegraphics[width=\imgwidth]{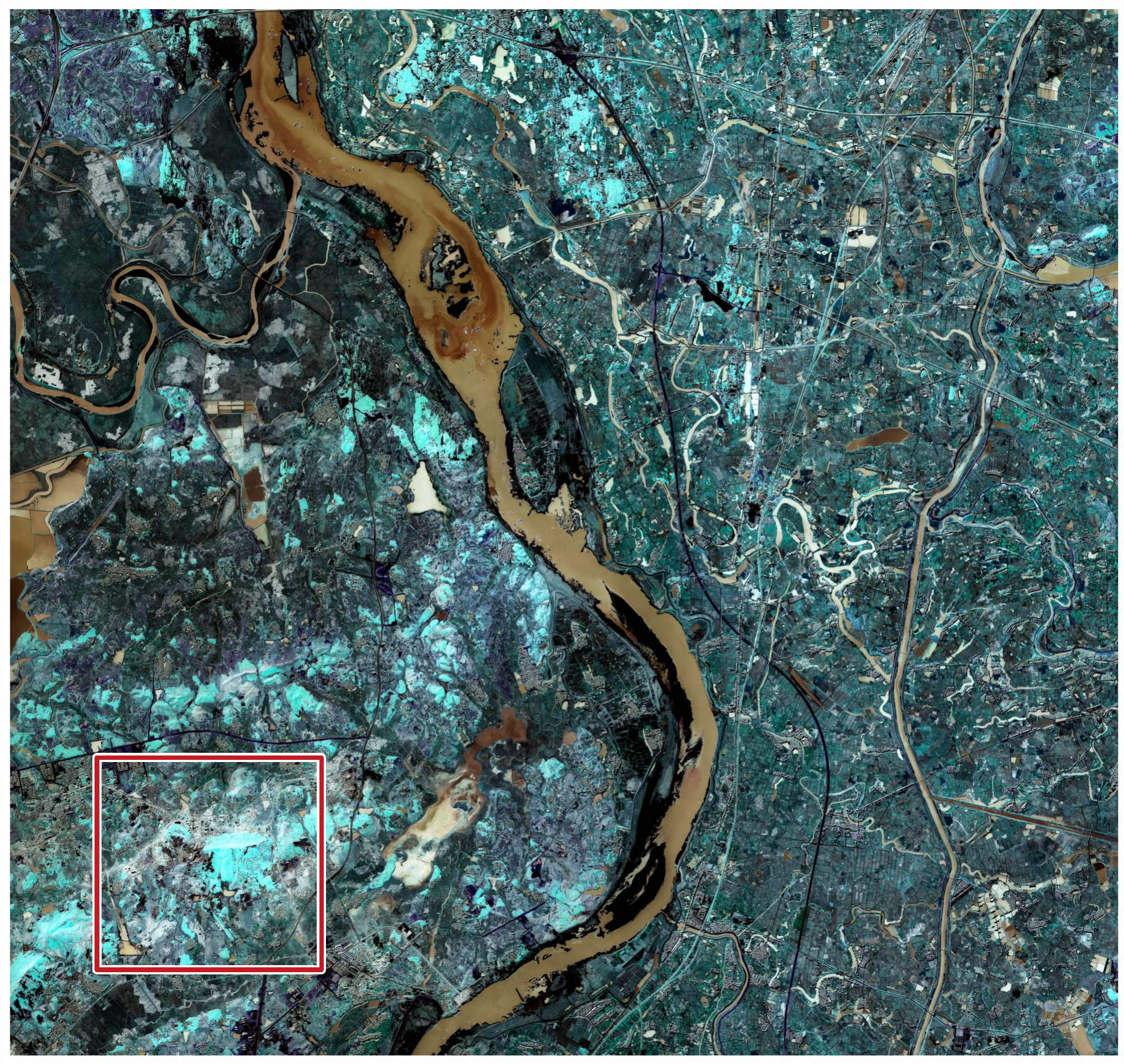}&
 \includegraphics[width=\imgwidth]{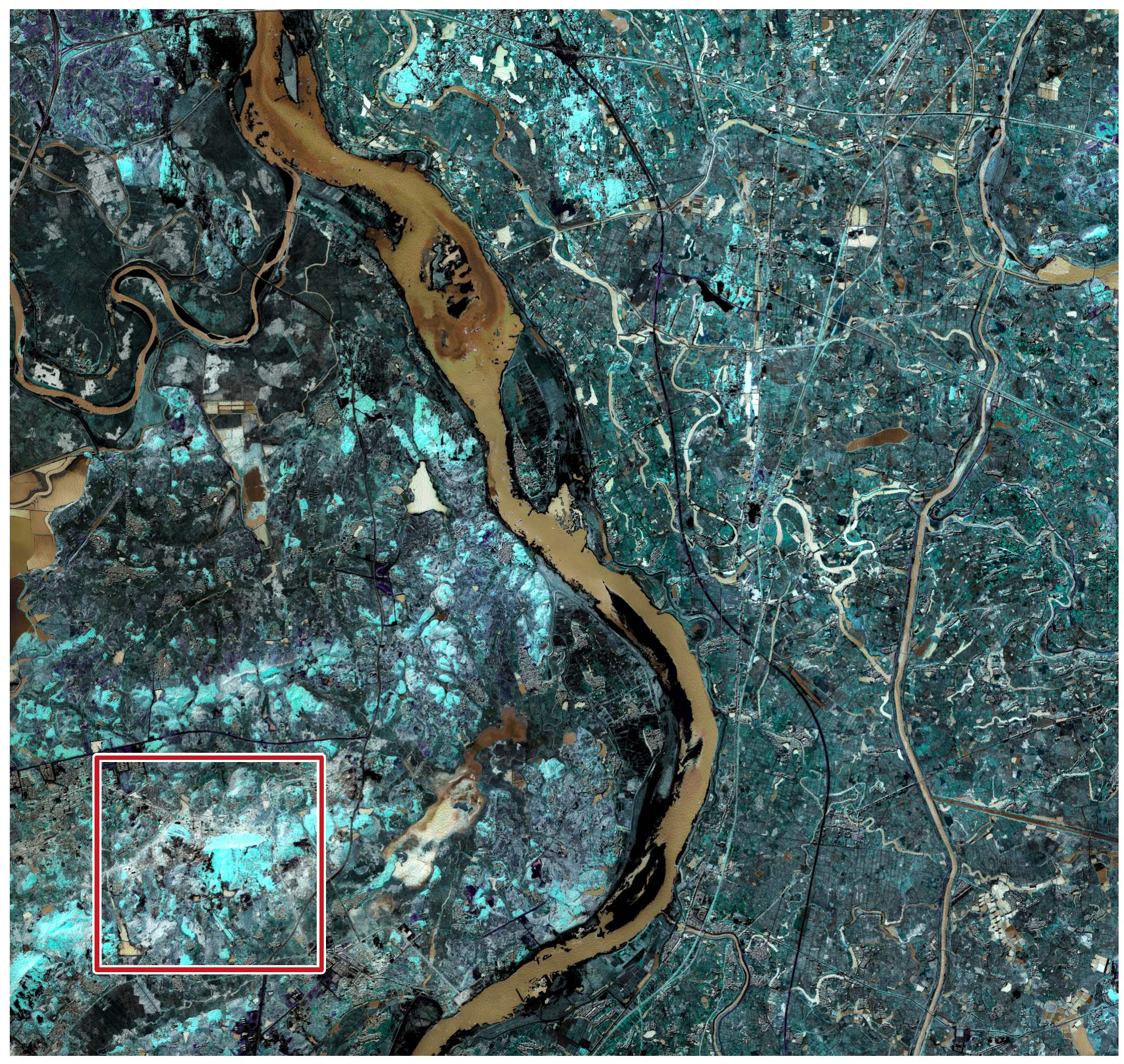}&
 \includegraphics[width=\imgwidth]{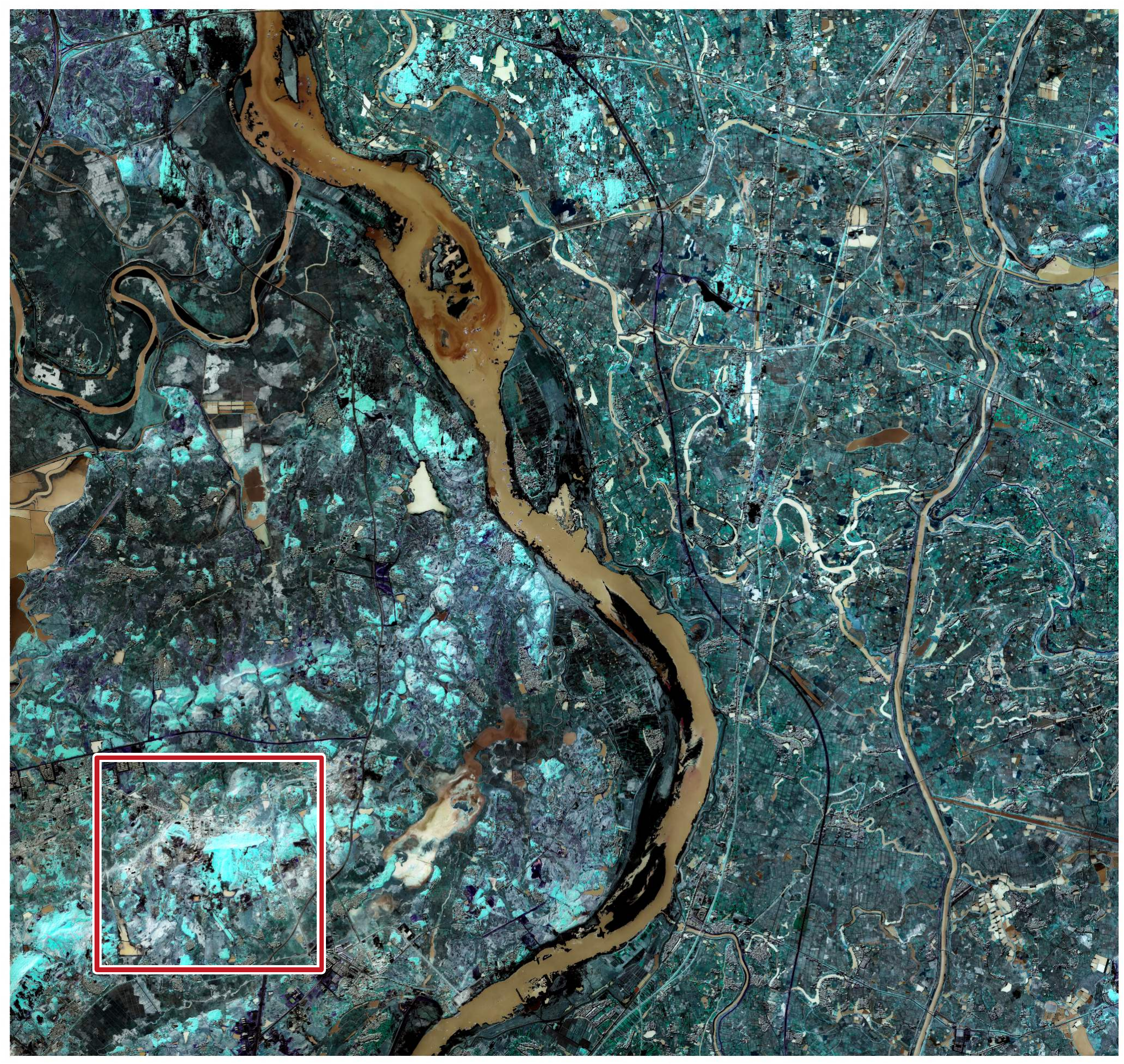}&
 \includegraphics[width=\imgwidth]{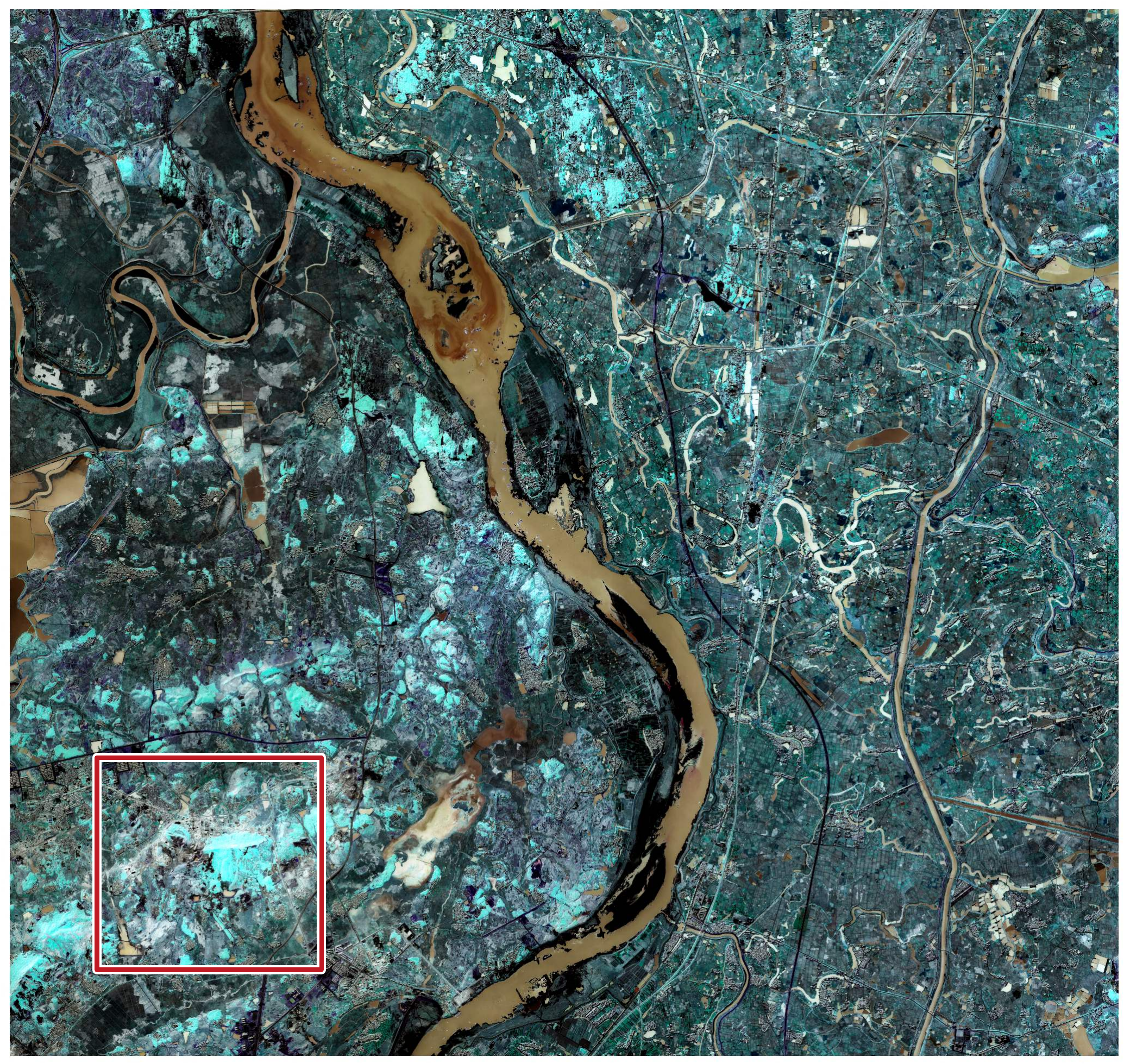}\\
 
 \includegraphics[width=\imgwidth]{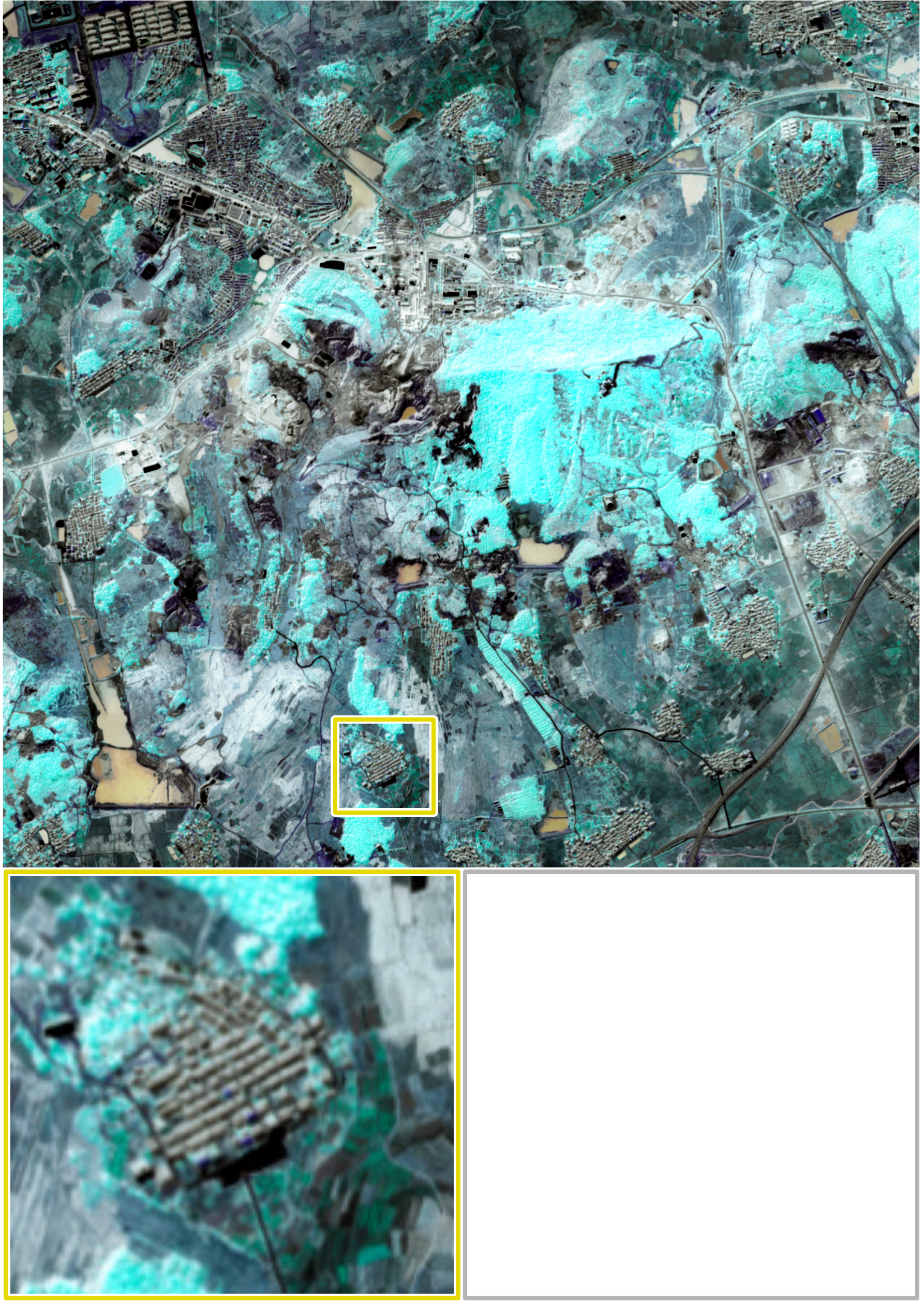}&
 \includegraphics[width=\imgwidth]{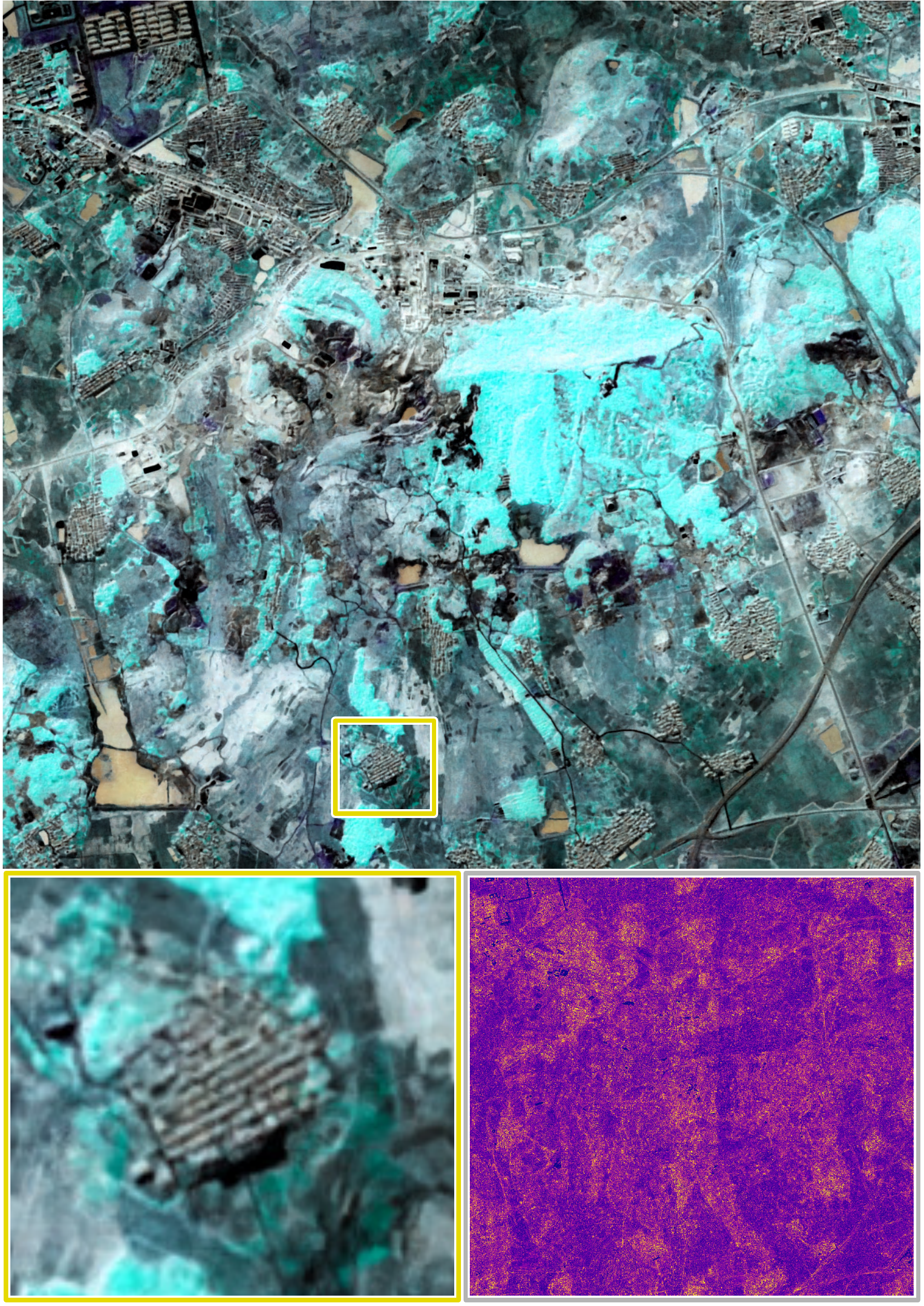}&
 \includegraphics[width=\imgwidth]{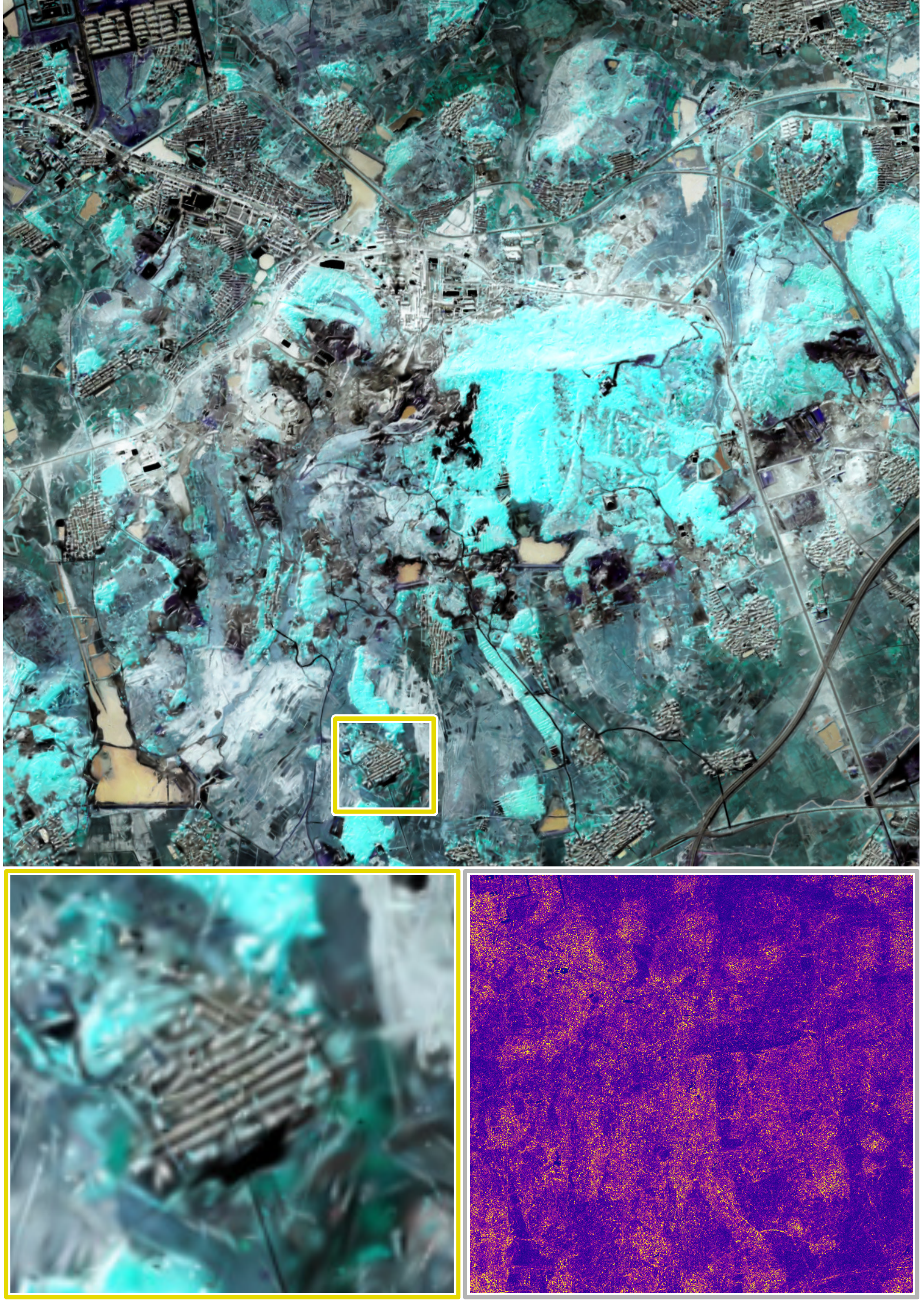}&
 \includegraphics[width=\imgwidth]{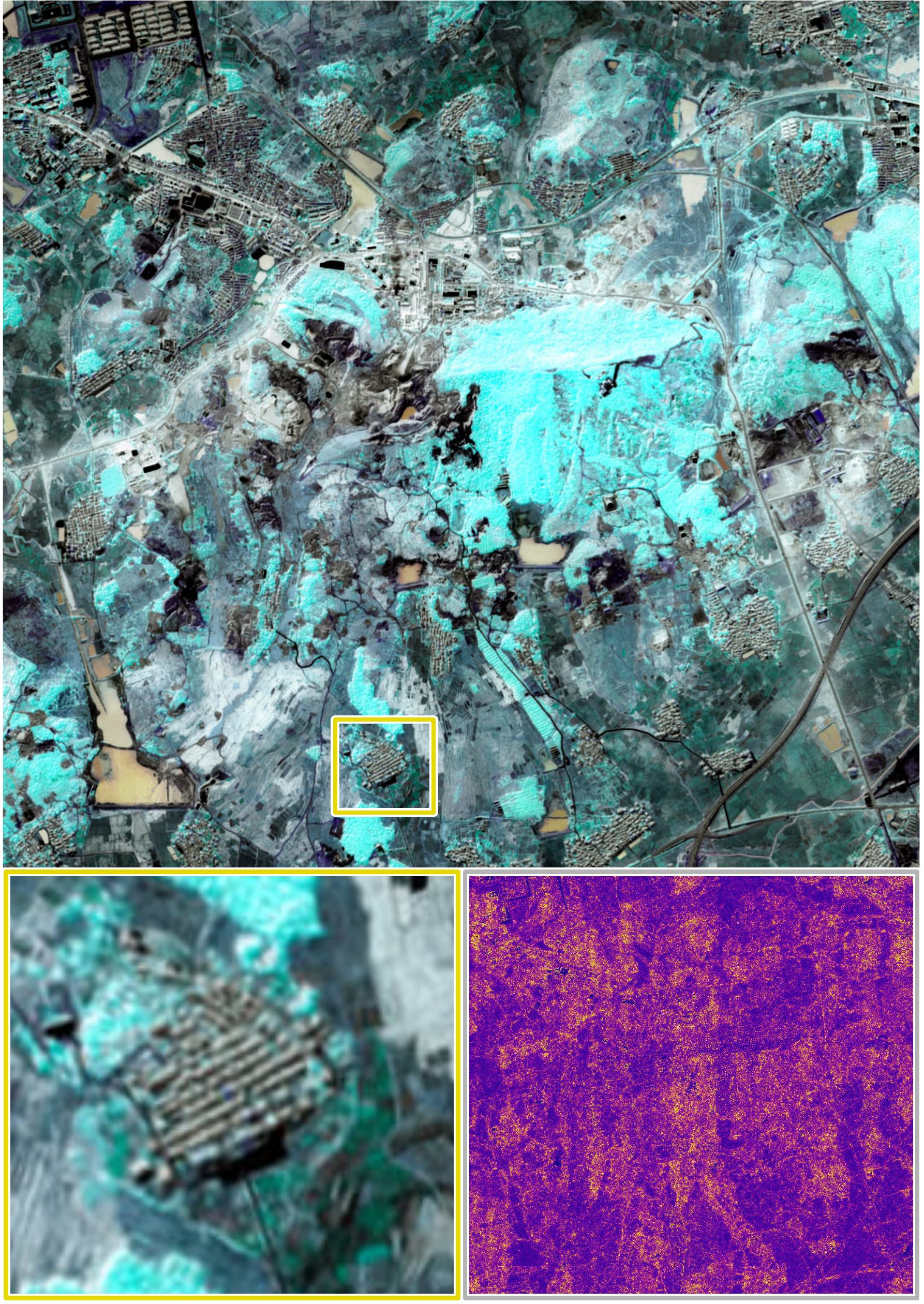}&
 \includegraphics[width=\imgwidth]{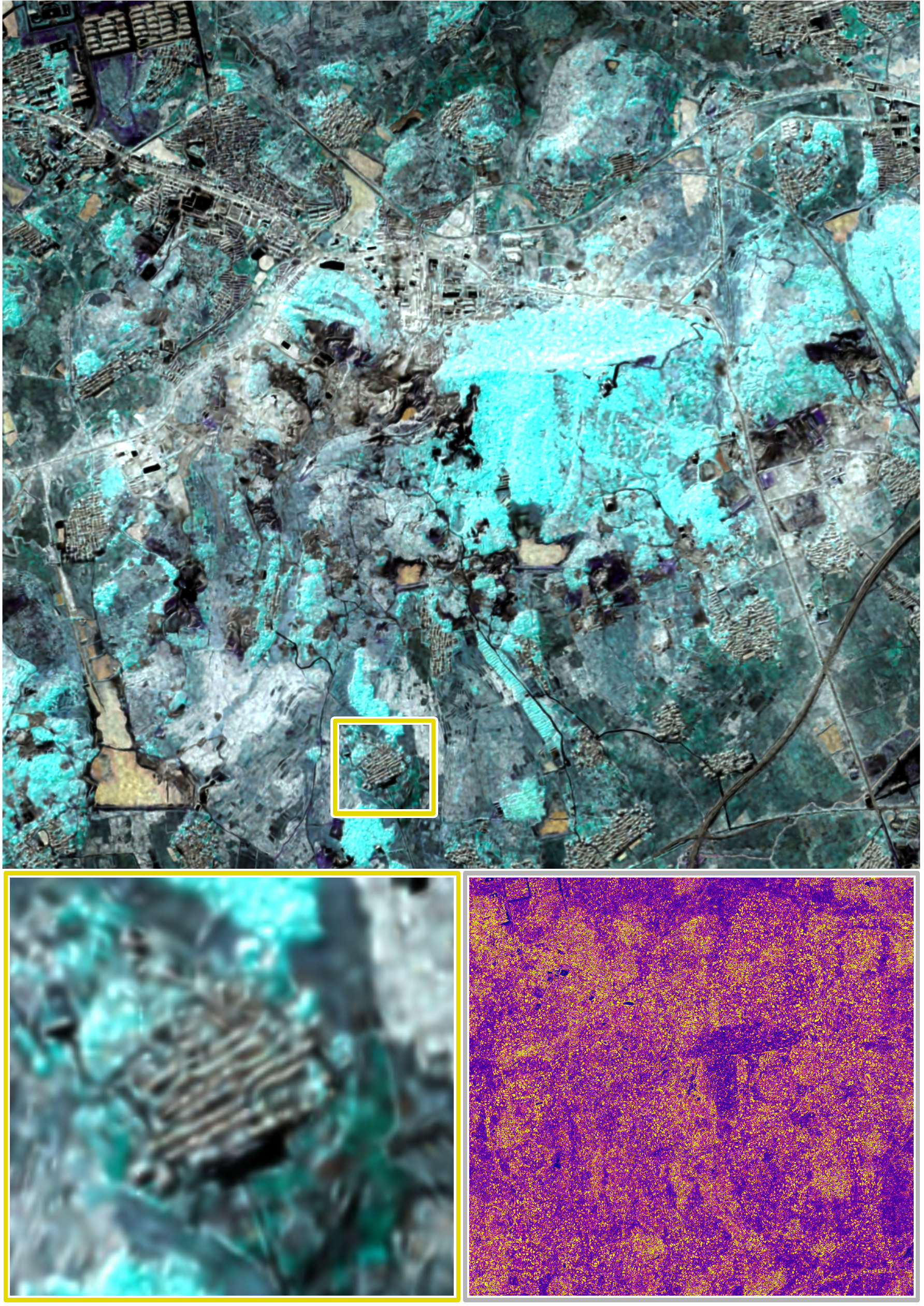}&
 \includegraphics[width=\imgwidth]{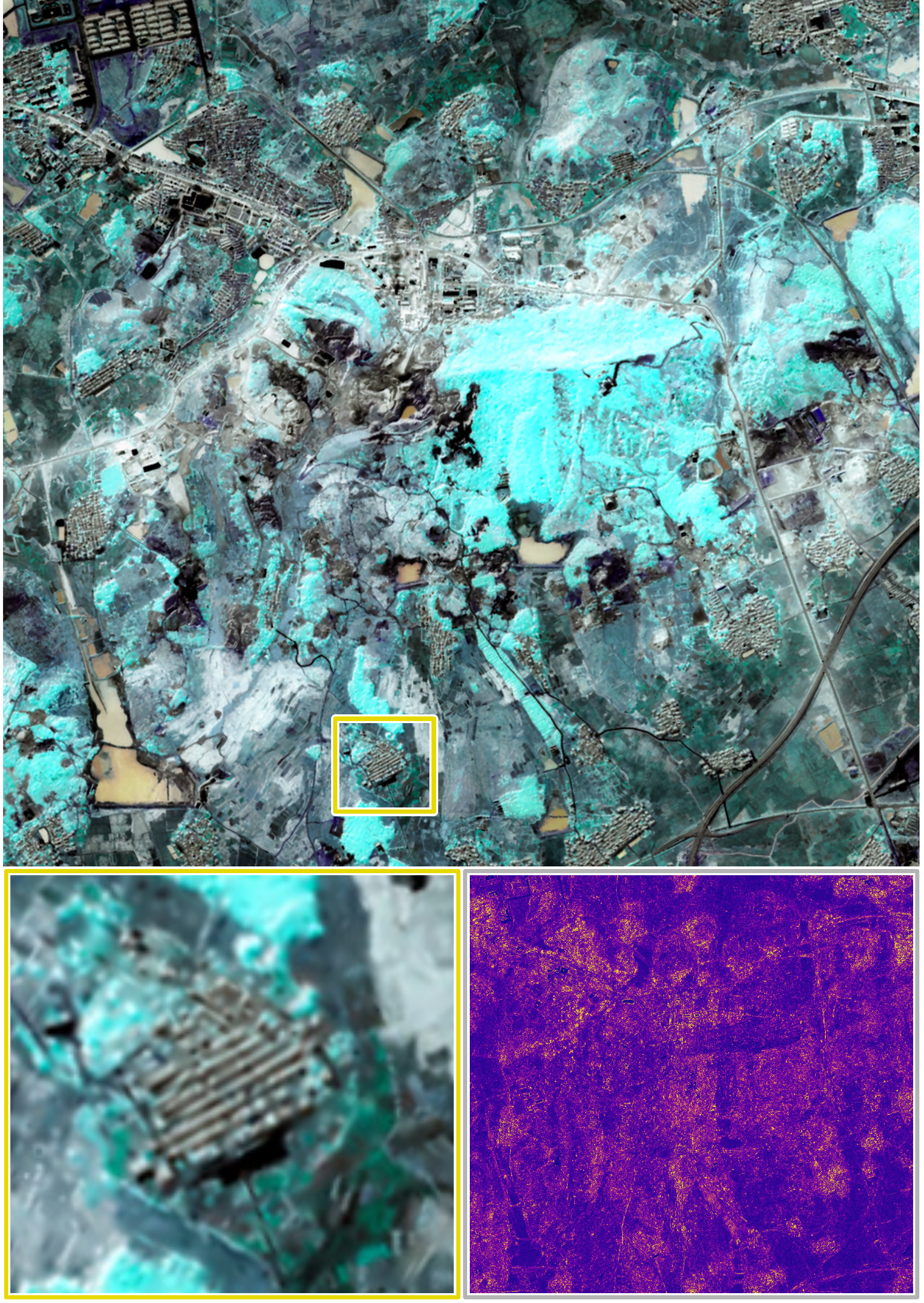}&
 \includegraphics[width=\imgwidth]{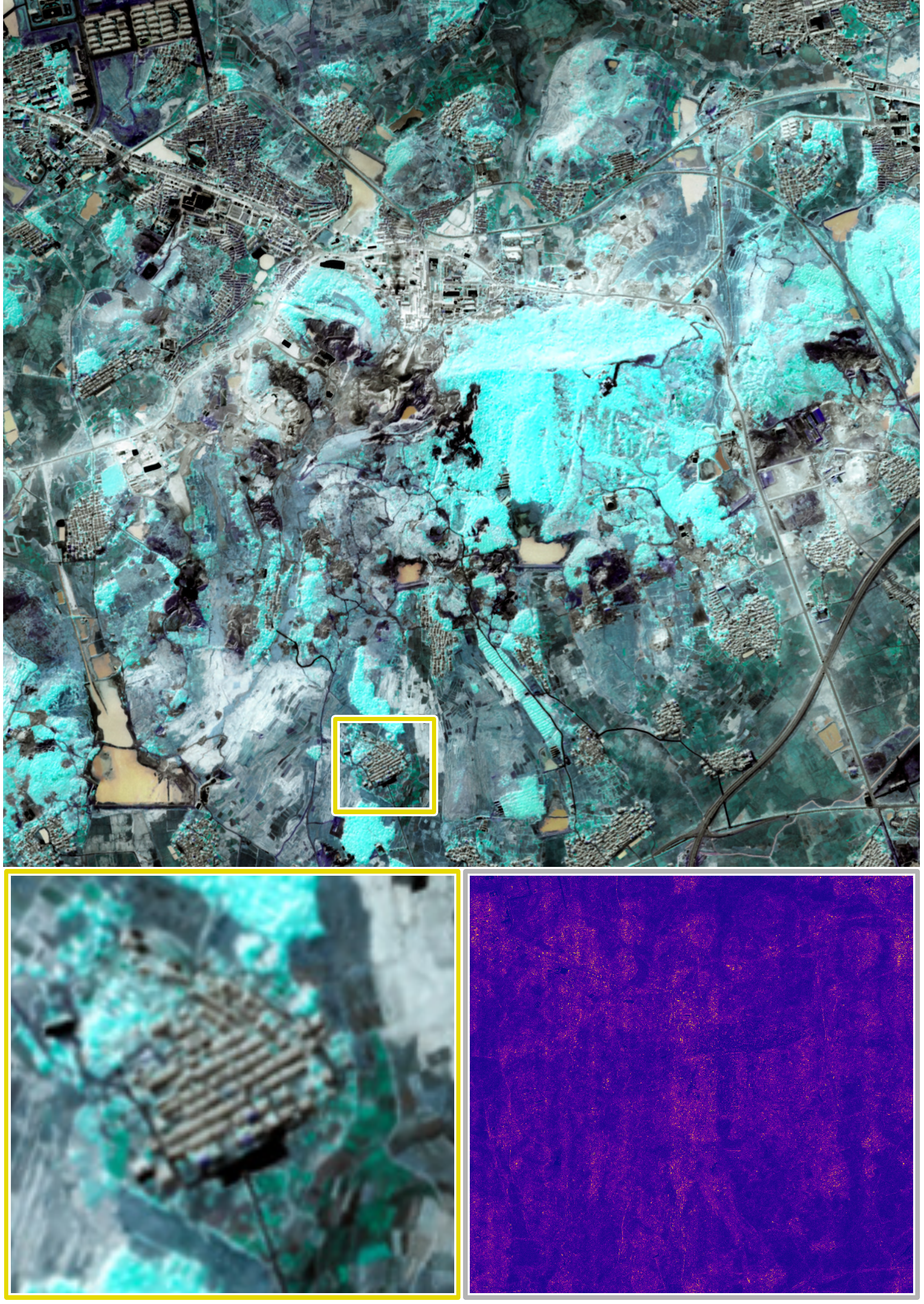}\\

\mbox{\footnotesize } &
\mbox{\smallfont
 29.08dB/21.07MB}&
\mbox{\smallfont 29.30dB/112.61MB}  &
\mbox{\smallfont 32.81dB/106.81MB} &
\mbox{\smallfont 27.56dB/23.37MB}  &
\mbox{\smallfont 31.52dB/16.00MB} &
\mbox{\smallfont 36.76dB/40.55MB}\\

\includegraphics[width=\imgwidth]{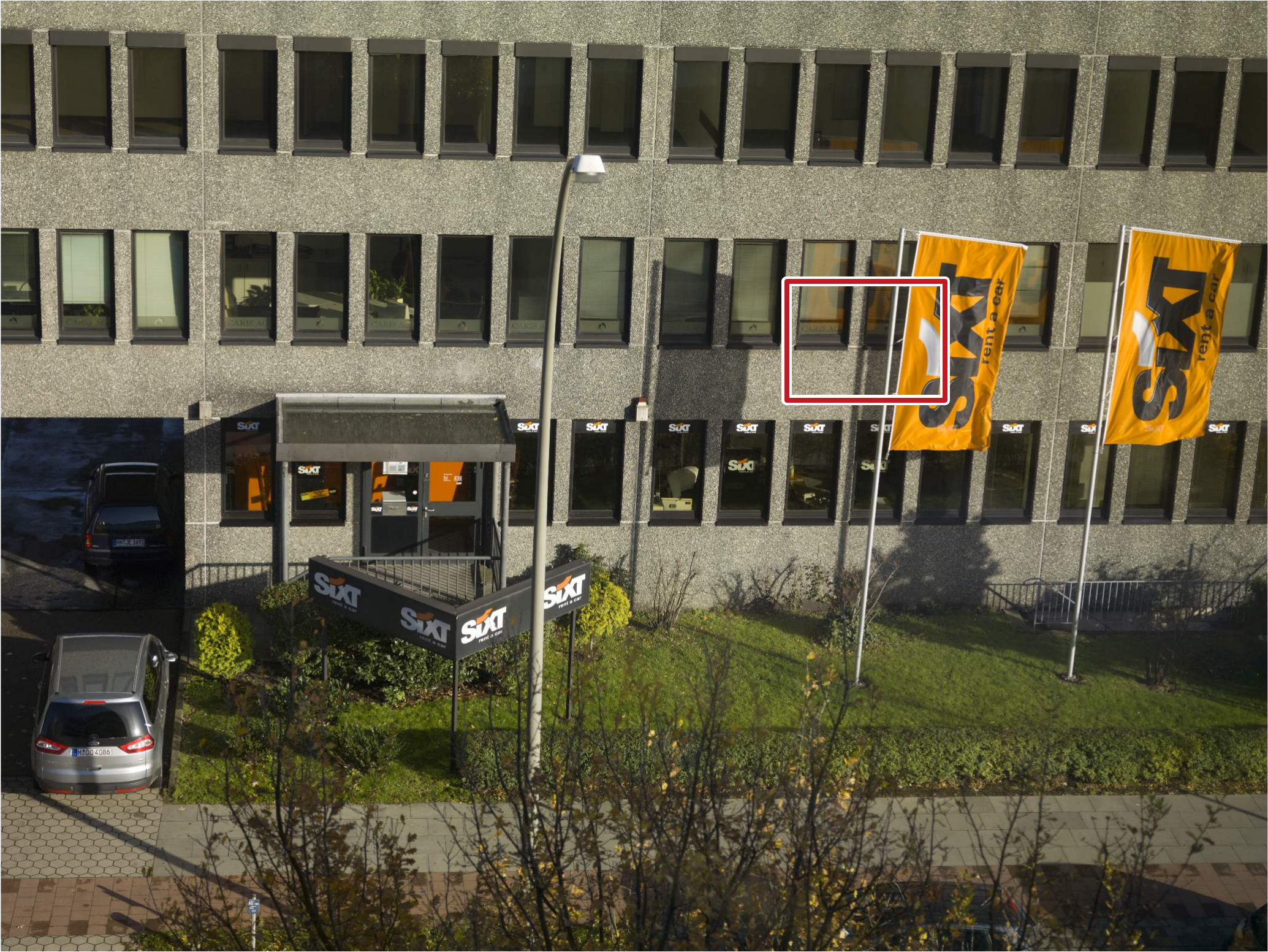}&
\includegraphics[width=\imgwidth]{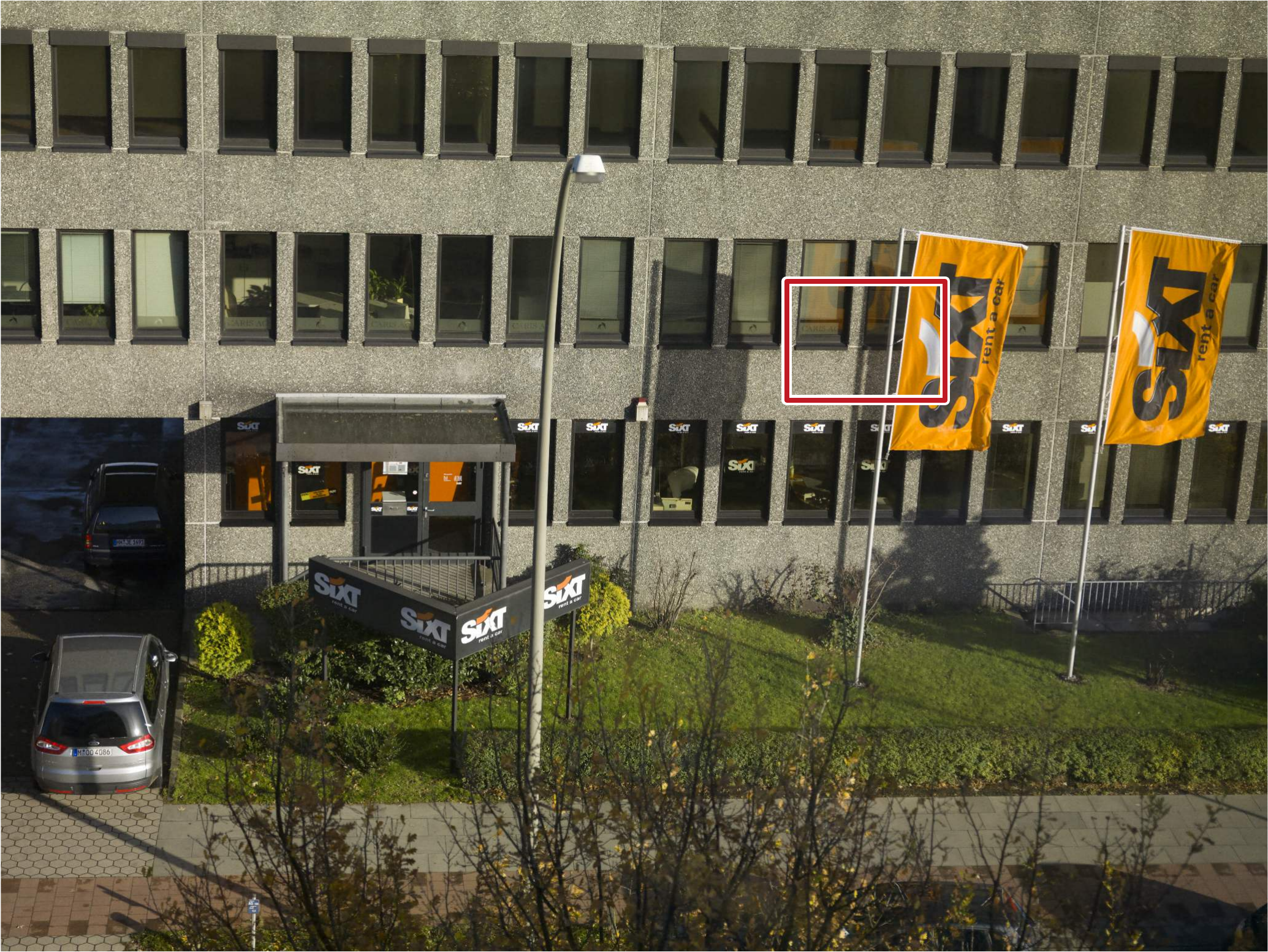}&
 \includegraphics[width=\imgwidth]{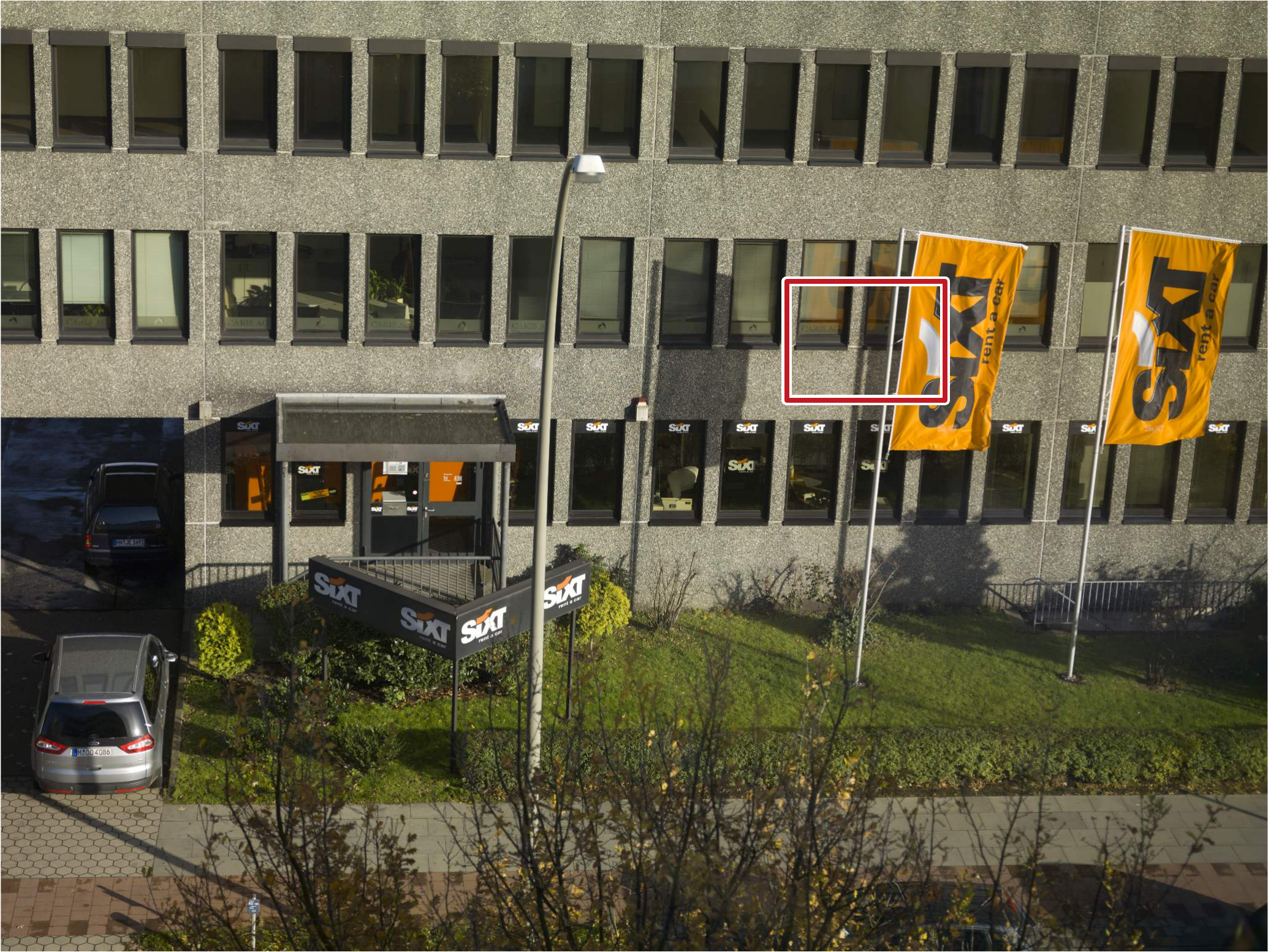}&
  \includegraphics[width=\imgwidth]{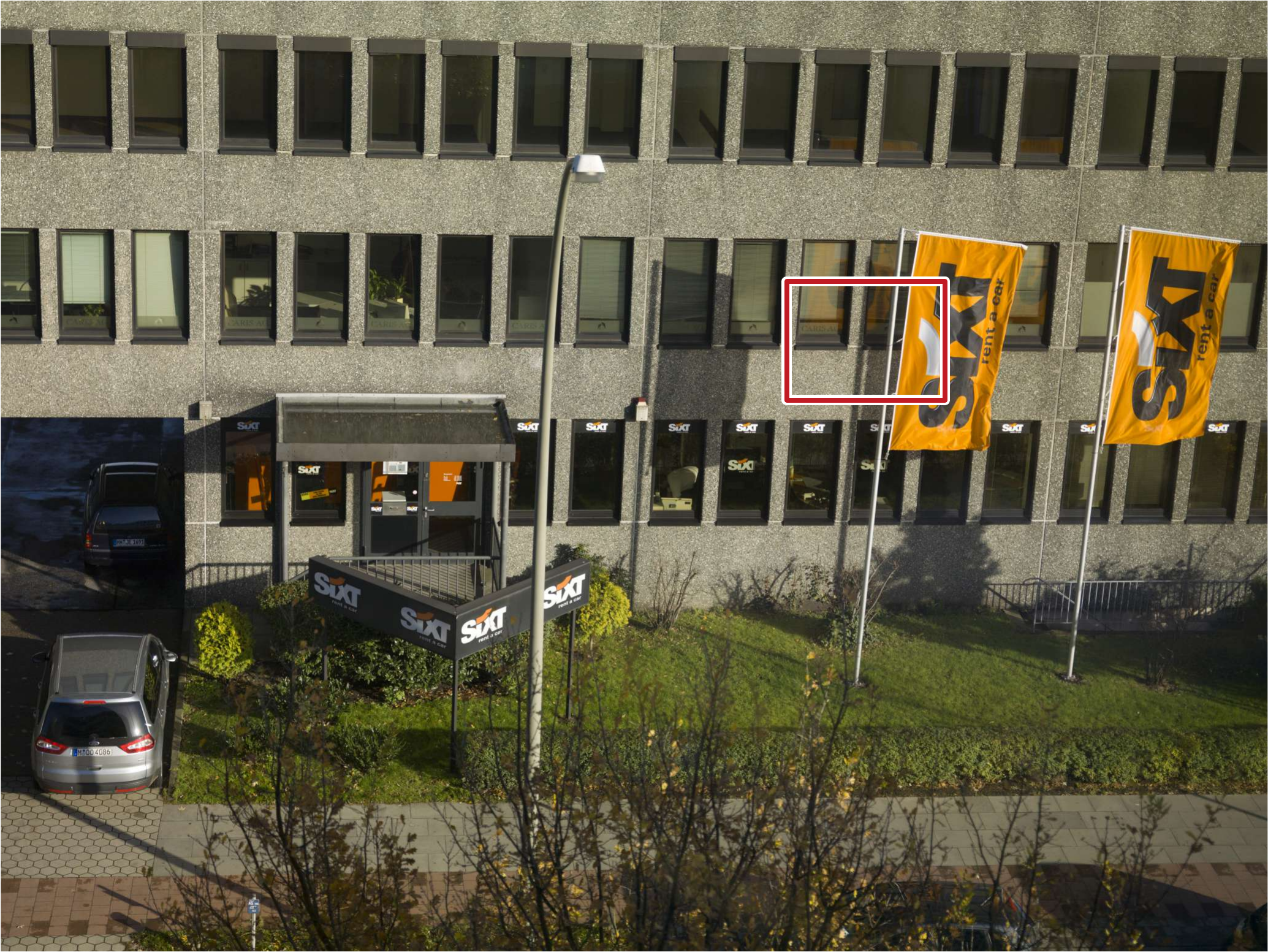}&
 \includegraphics[width=\imgwidth]{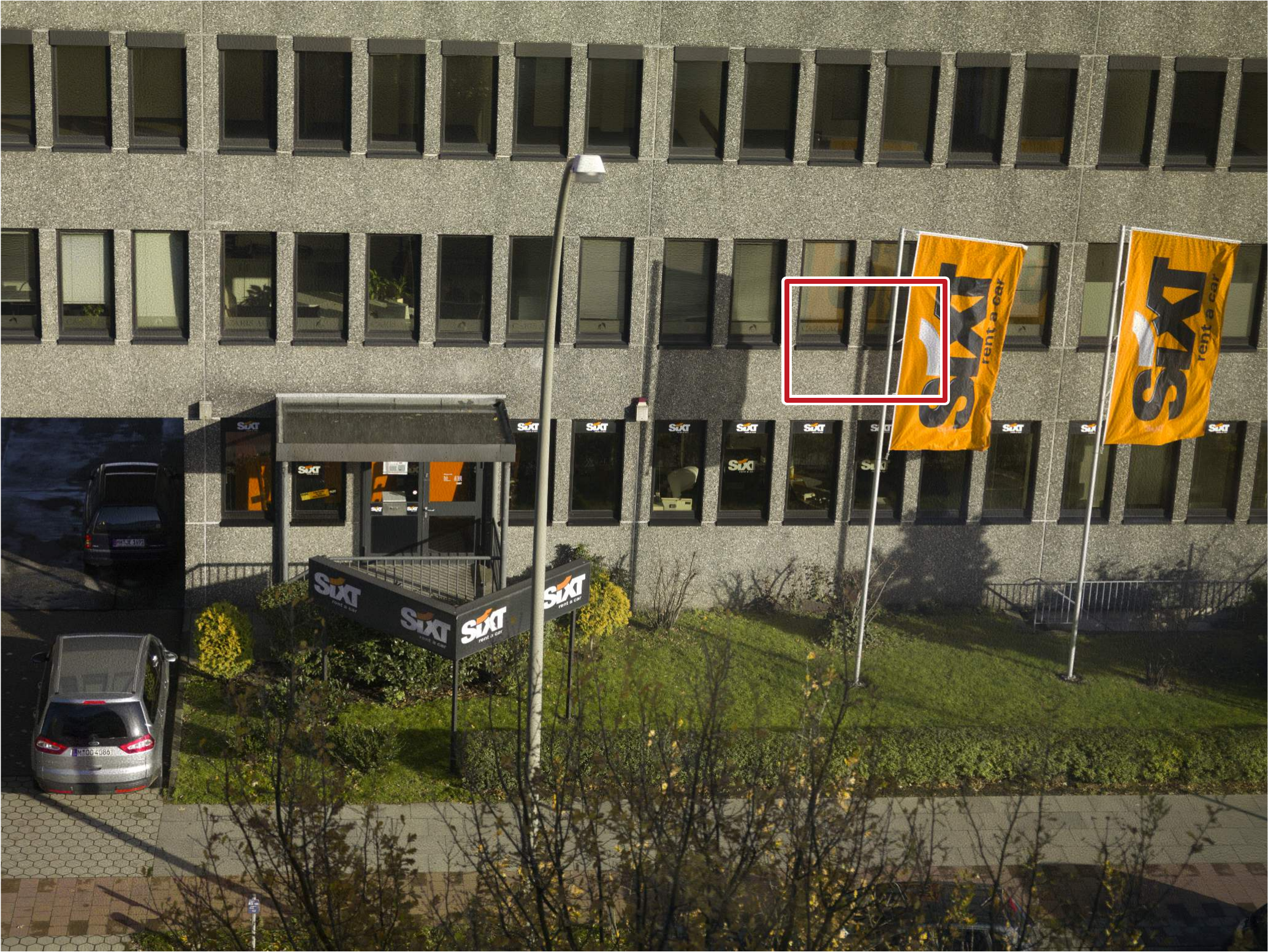}&
 \includegraphics[width=\imgwidth]{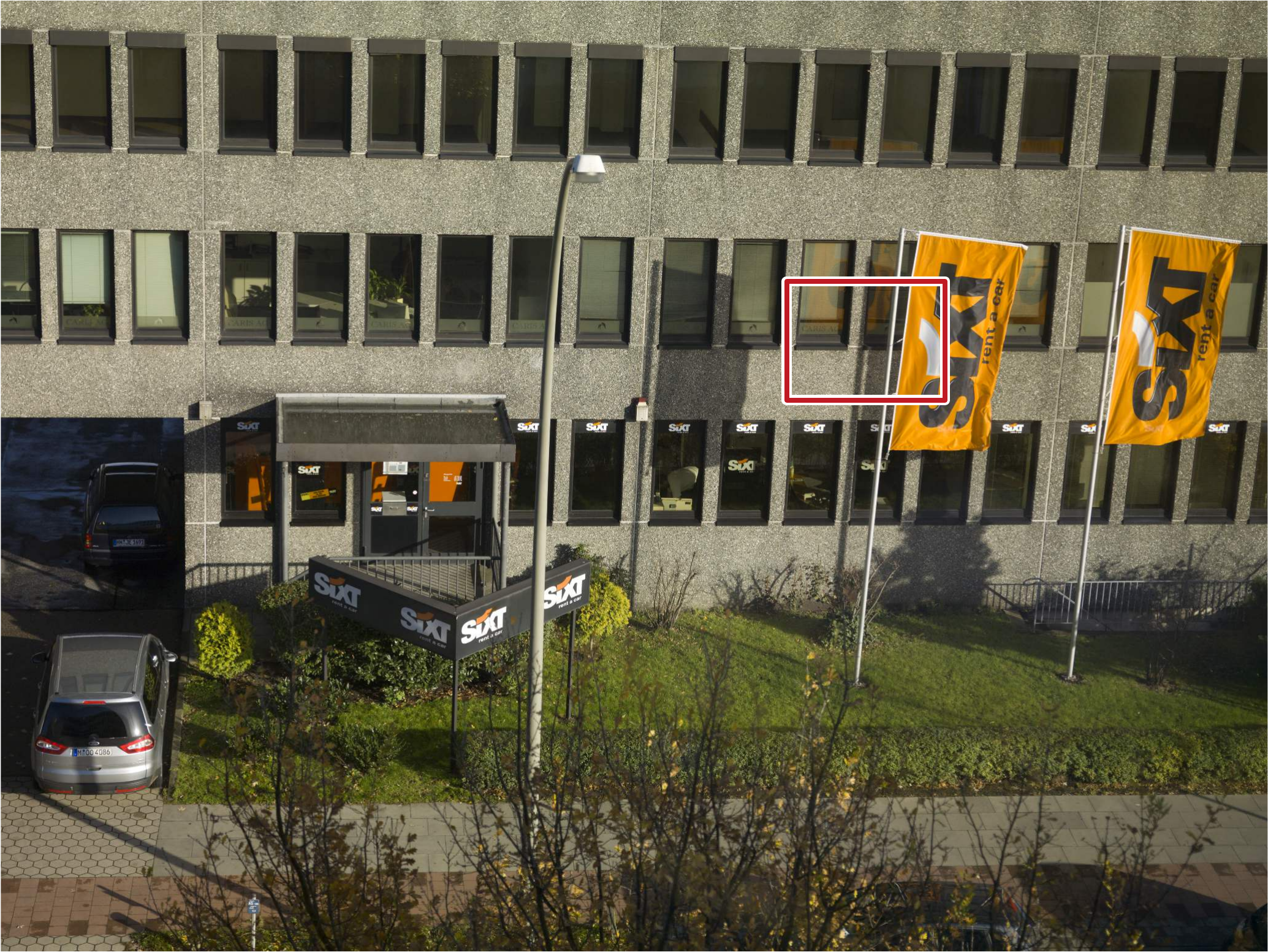}&
 \includegraphics[width=\imgwidth]{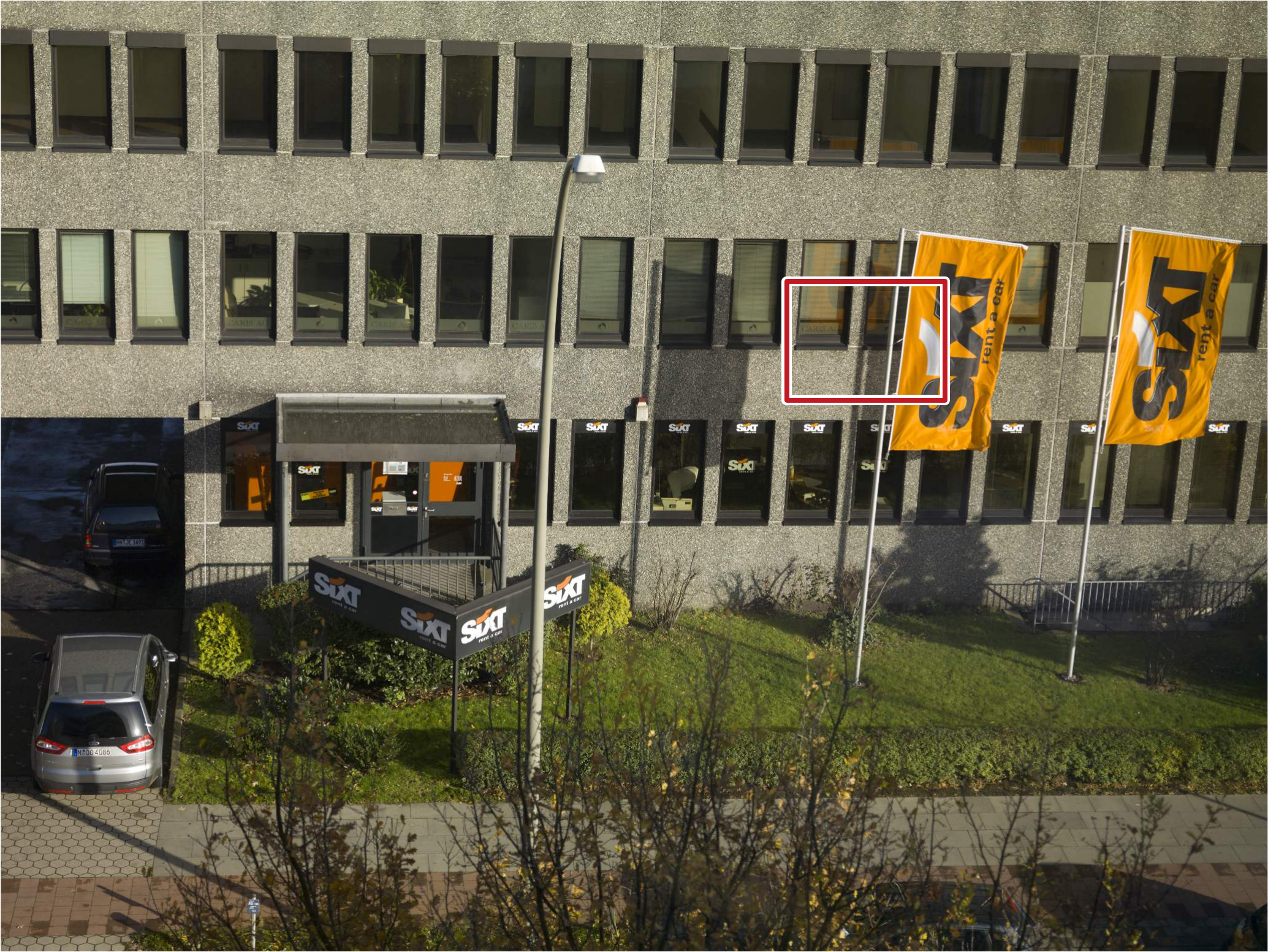}\\
 \includegraphics[width=\imgwidth]{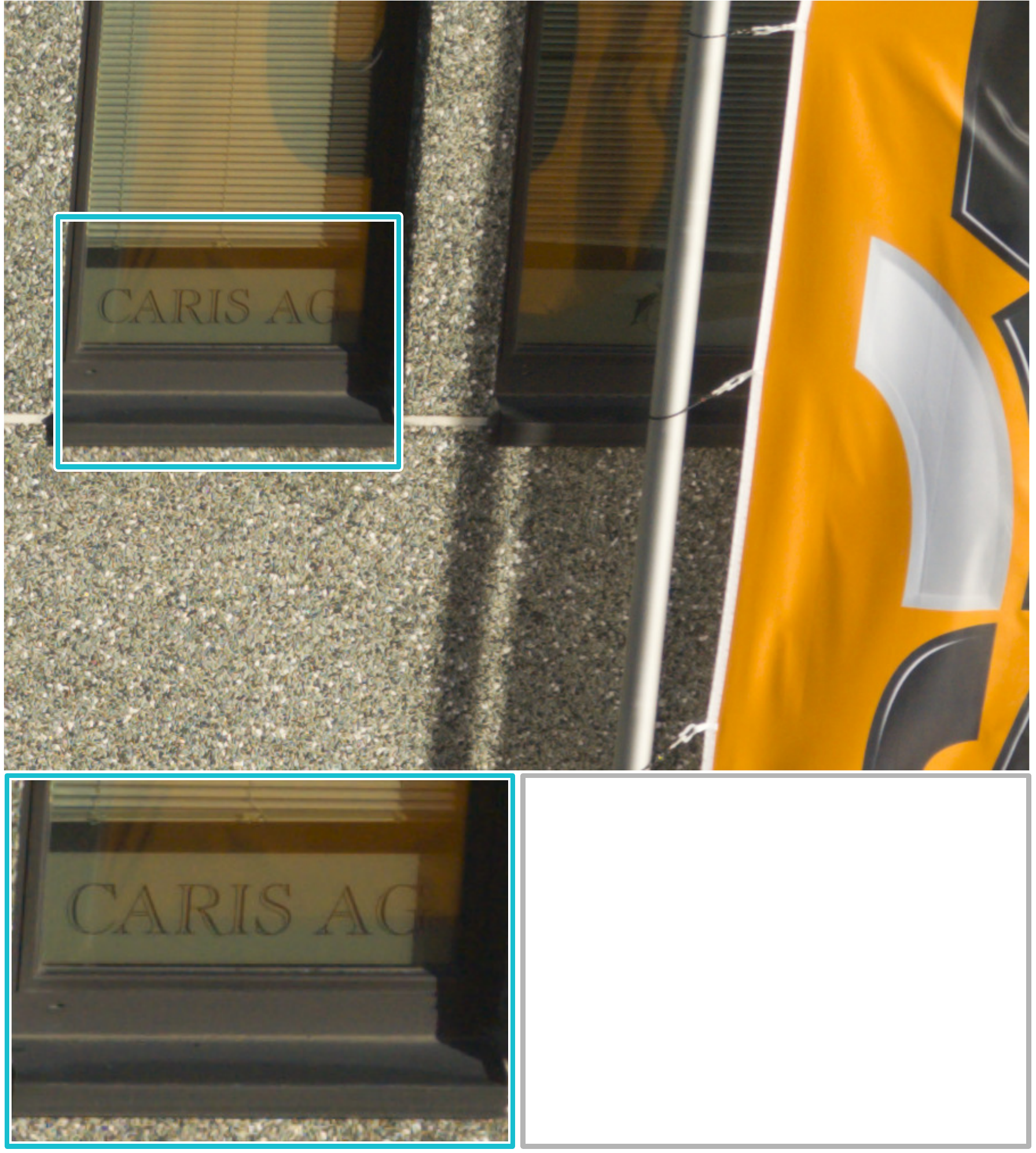}&
\includegraphics[width=\imgwidth]{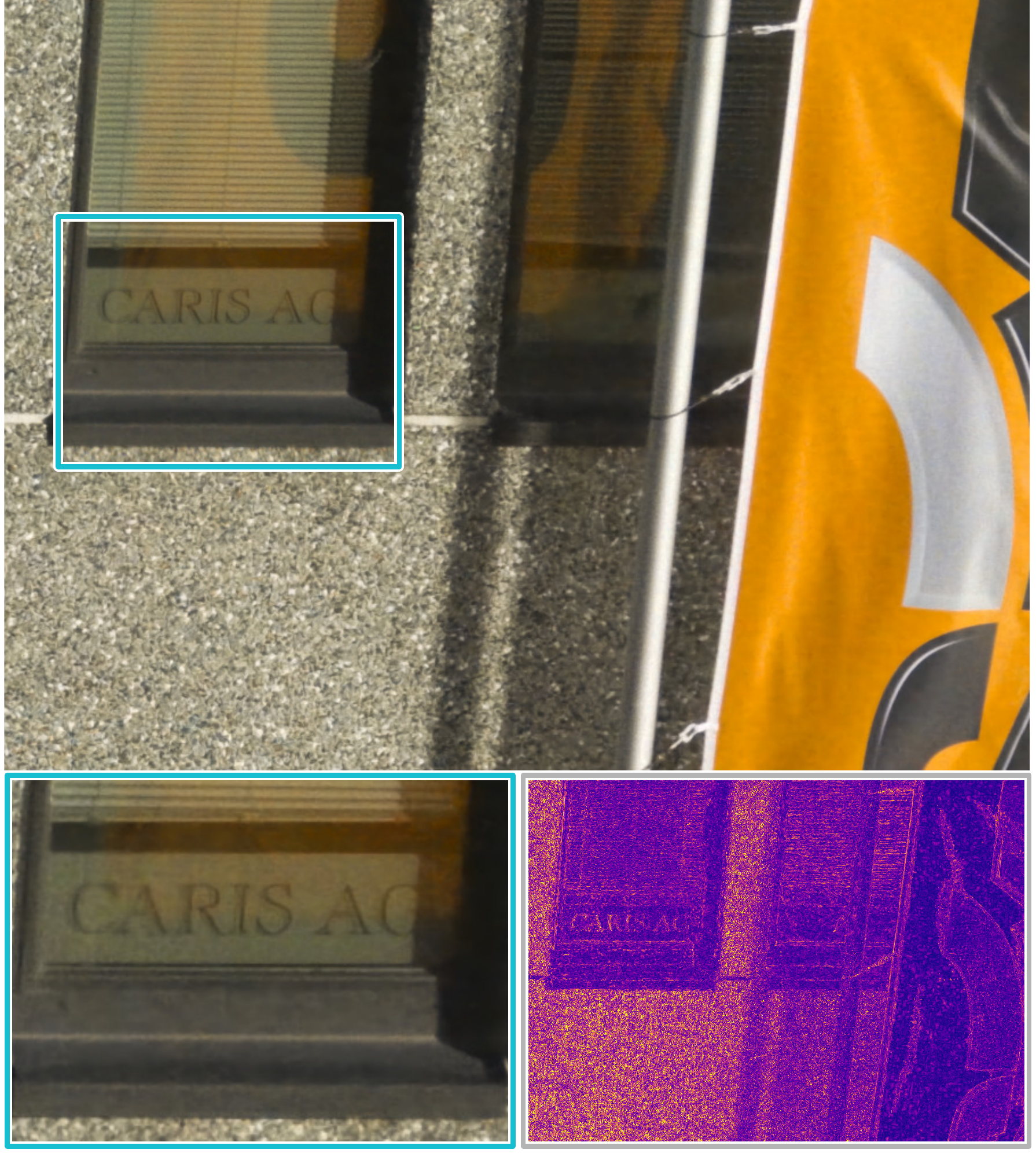}&
 \includegraphics[width=\imgwidth]{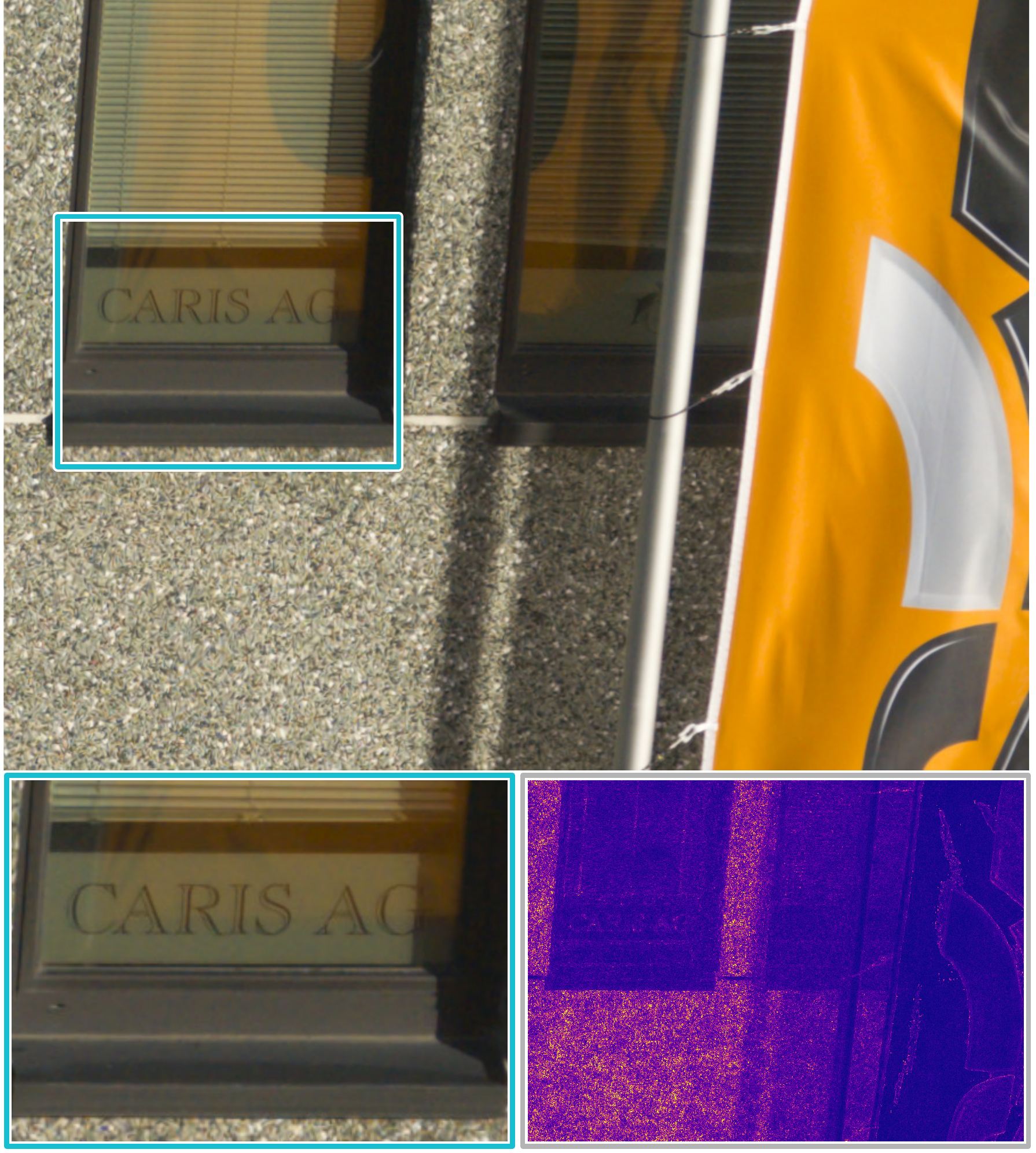}&
  \includegraphics[width=\imgwidth]{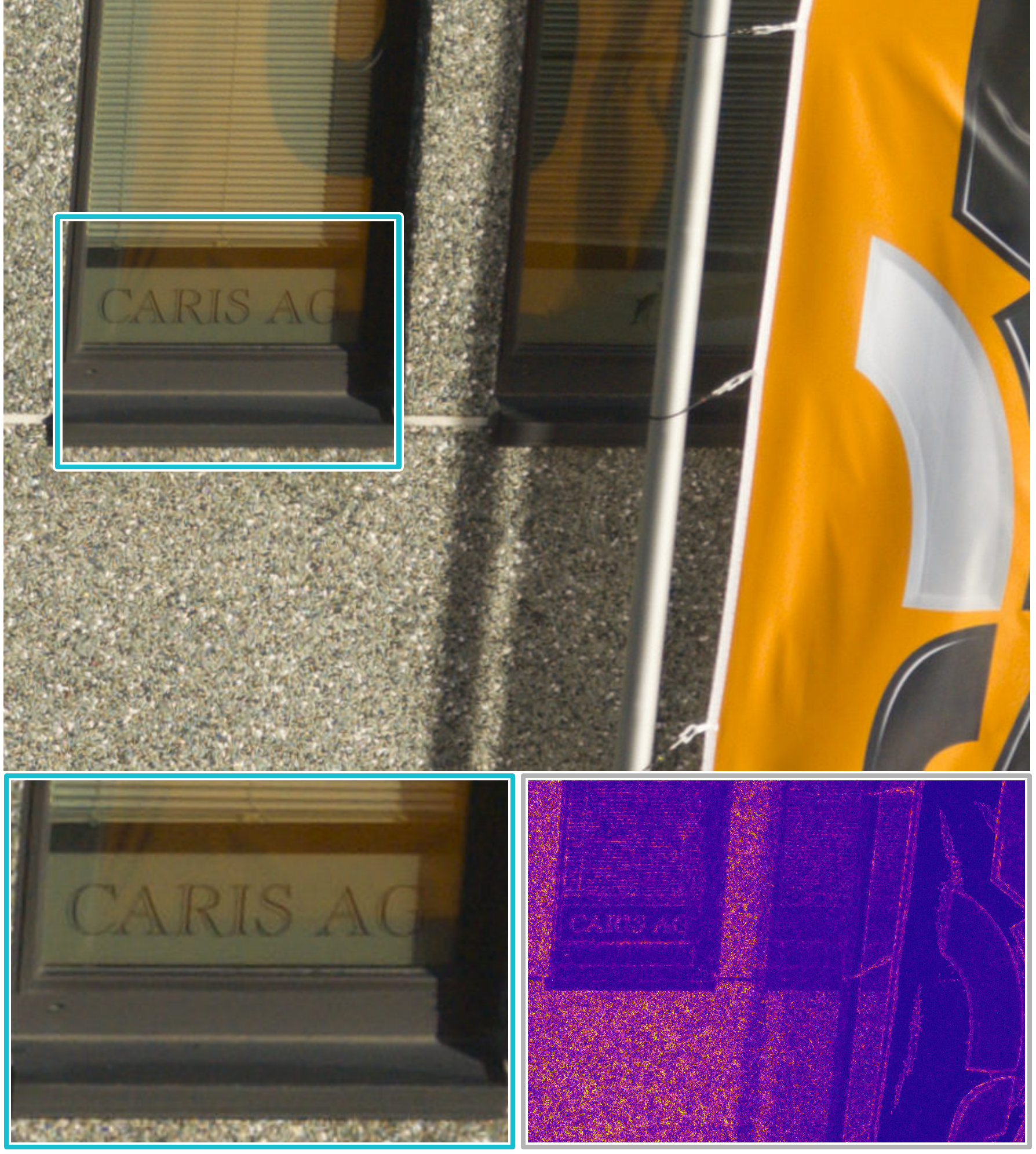}&
 \includegraphics[width=\imgwidth]{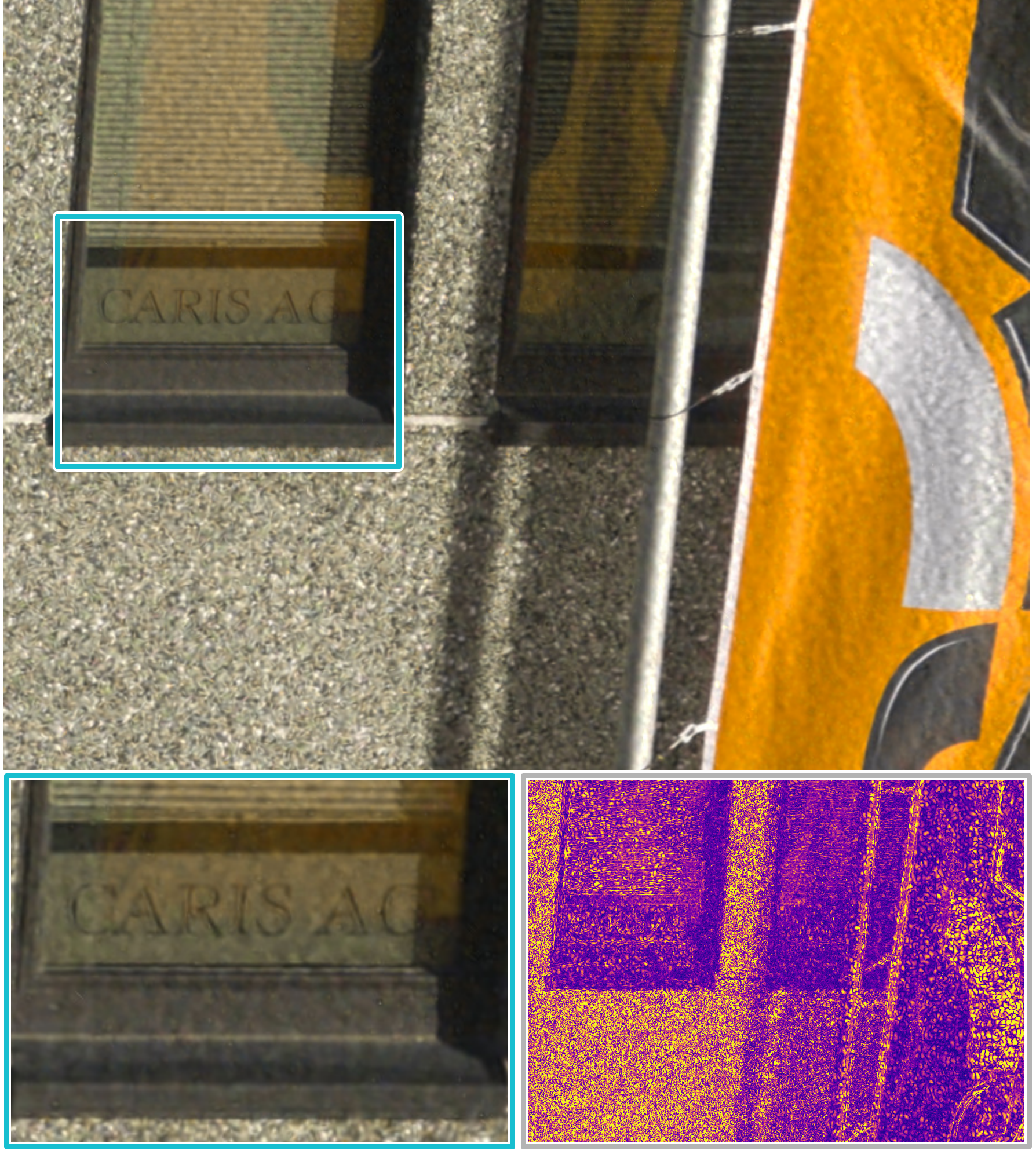}&
 \includegraphics[width=\imgwidth]{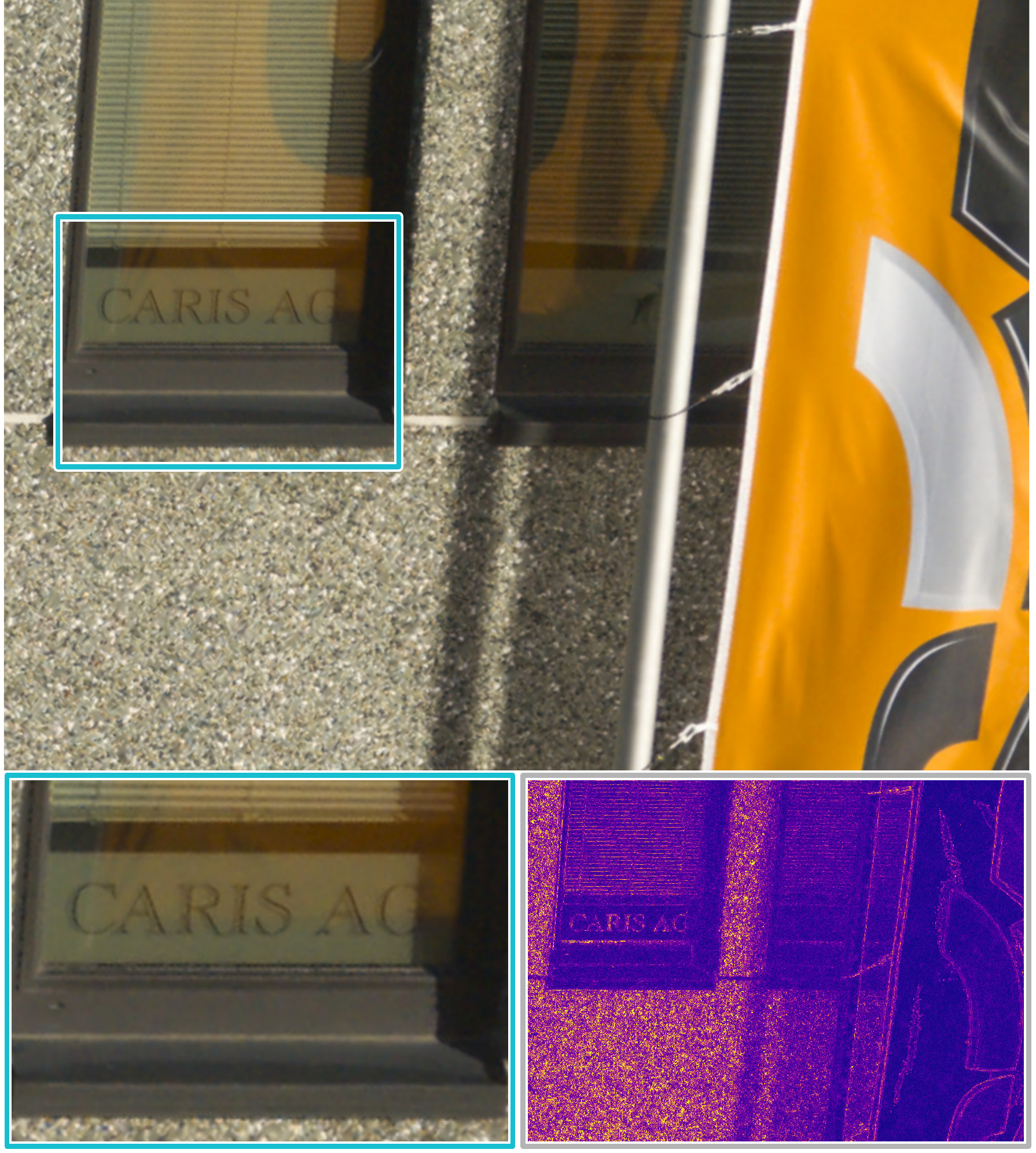}&
 \includegraphics[width=\imgwidth]{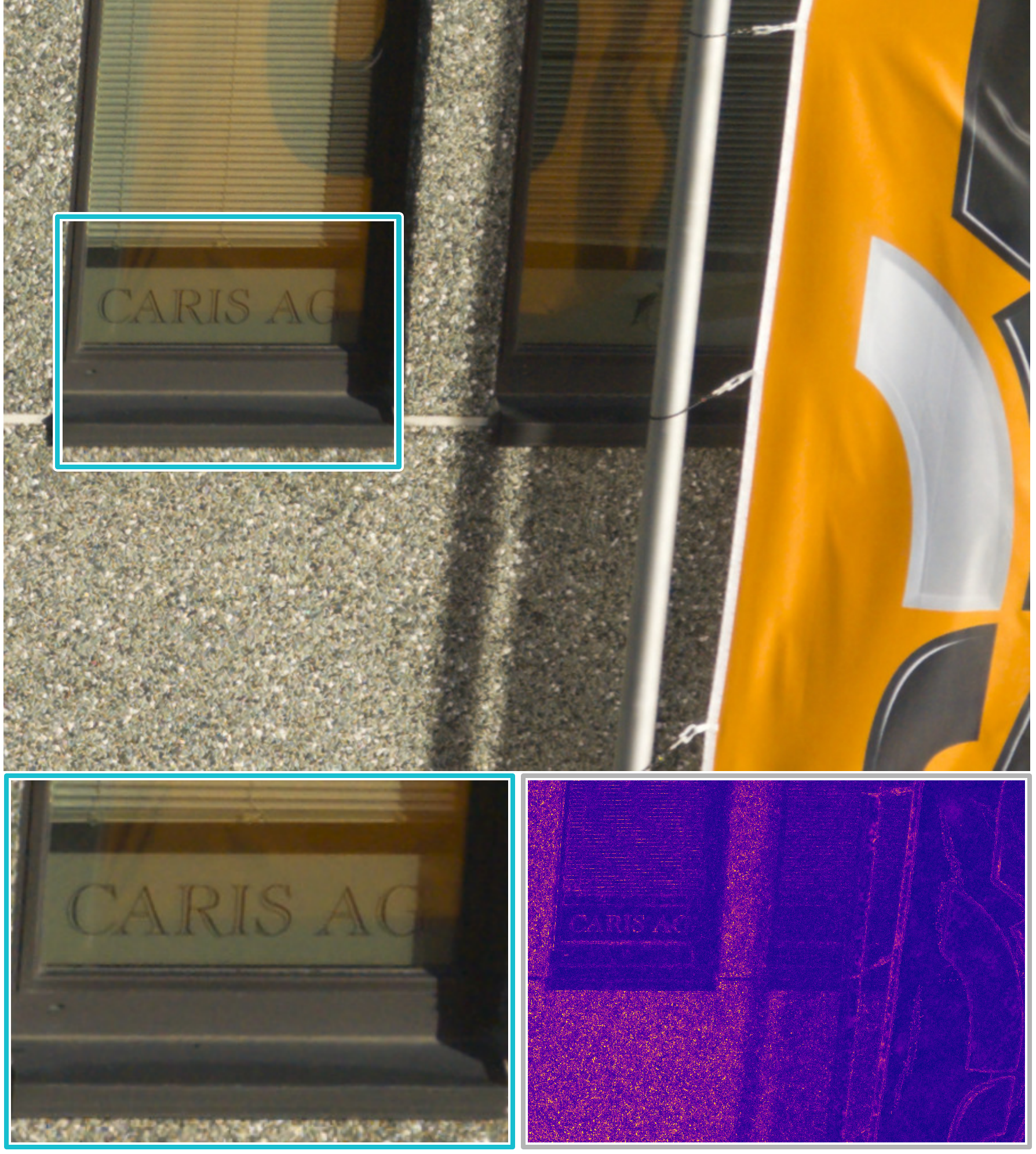}\\

\mbox{\smallfont } & 
\mbox{\smallfont 34.05dB/21.07MB}&
\mbox{\smallfont 37.06dB/105.81MB} &
\mbox{\smallfont 37.34dB/106.81MB} &
\mbox{\smallfont 20.80dB/23.37MB}&
\mbox{\smallfont 35.77dB/11.76MB}&
\mbox{\smallfont 39.57dB/31.01MB}\\
\mbox{\smallfont (a): GT} &
\mbox{\smallfont (b): I-NGP}&
\mbox{\smallfont (c): Scaffold-GS} &
\mbox{\smallfont (d): LIG} &
\mbox{\smallfont (e): GaussianImage} &
\mbox{\smallfont (f): SGI (low-rate)}&
\mbox{\smallfont (g): SGI (high-rate) }\\

\end{array}$
\end{center}
\vspace{-.15in} 
\caption{{\bf Visual comparisons on the FGF2 and ICB (w/ zoom-in cases and error maps).}
Qualitative results on representative examples from FGF2 (top) and ICB (bottom), comparing SGI with I-NGP~\cite{Muller-TOG22}, Scaffold-GS~\cite{Lu-CVPR24}, LIG~\cite{Zhu-AAAI25}, and GaussianImage~\cite{Zhang-ECCV24}.
Zoom-in regions highlight perceptual differences.
In the third row of each block, we visualize the per-pixel reconstruction error using heatmaps, where warmer colors (e.g., yellow) indicate higher deviation from the ground truth.
PSNR (dB) and storage size (MB) for each method are shown below the visualizations.
}
\label{fig:GF2-comprasion}
\vspace{-.15in}
\end{figure*}

\subsection{Experiment Setup}

{\bf Datasets, Metrics, and Baselines.}
We evaluate our method on three real-world datasets with high-resolution images spanning diverse domains, including natural, satellite, and biomedical images.
The first contains four satellite images, each with approximately 51 megapixels (MP), from the Full-resolution Gaofen-2 (FGF2) dataset~\cite{Li-IAC15}.
The second dataset includes two natural images from the Image Compression Benchmark (ICB)~\cite{image-dataset}, 
with resolutions of 27.7 MP and 39.1 MP,  and the biomedical dataset (STimage~\cite{Chen-NIPS24}) consists of three histopathology images averaging 76 MP.
We compare our method with a wide range of existing image representation techniques, including both INR and Gaussian-based methods.  
Specifically, we include SIREN~\cite{Sitzmann-NeurIPS20} and Instant-NGP (I-NGP)~\cite{Muller-TOG22} as representative INR baselines, and compare against 3D Gaussian methods (3DGS~\cite{Kerbl-ToG23}, Scaffold-GS~\cite{Lu-CVPR24}, and HAC~\cite{Chen-ECCV24}) as well as 2D Gaussian approaches (LIG~\cite{Zhu-AAAI25}, GaussianImage~\cite{Zhang-ECCV24}).
To assess reconstruction quality, we report peak signal-to-noise ratio (PSNR), structural similarity index (SSIM), and learned perceptual image patch similarity (LPIPS).  
We also measure storage cost in MB and optimization time in minutes.

\begin{table}[!htb]
    \centering
    \caption{\textbf{Quantitative results on STimage.} Optimization time is in minutes, and model size is in megabytes (MB). To fit INR-based methods (SIREN and I-NGP) on these large-scale datasets under limited GPU memory, we adopt a patch-based training strategy that loads randomly sampled coordinate and pixel batches from CPU to GPU on the fly. 
    Our SGI is evaluated in two settings: \textbf{low-rate} (3.5M Gaussians) and \textbf{high-rate} (10M Gaussians), both demonstrating strong trade-offs between fidelity, compactness, and optimization efficiency. \sethlcolor{blue!20} \hl{Darker blue} and \sethlcolor{red!20} \hl{darker red} highlights indicate the best-performing method within the \sethlcolor{blue!8} \hl{low-rate} and \sethlcolor{red!8} \hl{high-rate} groups, respectively, in each metric. \emph{Reported numbers are averaged per image across the dataset.}}
    \vspace{-0.01in}
    \resizebox{\linewidth}{!}{
    {
    \begin{tabular}{lcccS[table-format=3.2]S[table-format=3.2]}
        \toprule
        
       Method & PSNR$\uparrow$ &SSIM$\uparrow$ &LPIPS$\downarrow$ &\multicolumn{1}{c}{Opt. Time$\downarrow$} &\multicolumn{1}{c}{Size$\downarrow$}
      \\
        \midrule
        \rowcolor{blue!8} SIREN~\cite{Sitzmann-NeurIPS20} (NeurIPS'20)
        &28.09 &0.8843 &0.3654 &983.48 &15.79
        \\
        \rowcolor{blue!8} I-NGP~\cite{Muller-TOG22} (TOG'22)
        &\cellcolor{blue!20}{35.57} &\cellcolor{blue!20}{0.9763} &\cellcolor{blue!20}{0.0518} &129.96 &21.07
        \\
        \rowcolor{blue!8} GaussianImage~\cite{Zhang-ECCV24} (ECCV'24) 
        &25.75 &0.6404 &0.4006 &407.70 &23.37
        \\
        \rowcolor{blue!8} Our SGI (low-rate)
        &33.96 &0.9743 &0.1196 &\cellcolor{blue!20}{103.43} &\cellcolor{blue!20}{10.05}
        \\
        \midrule
        \rowcolor{red!8} 3DGS~\cite{Kerbl-ToG23} (TOG'23)
        &38.51 &\cellcolor{red!20}{0.9957} &\cellcolor{red!20}{0.0192} &723.43 &787.73
        \\
        
        \rowcolor{red!8} LIG~\cite{Zhu-AAAI25} (AAAI'25)
        &34.33 &0.9714 &0.1386 &\cellcolor{red!20}{106.82} &106.81
        \\   
        \rowcolor{red!8} Our SGI (high-rate) 
        &\cellcolor{red!20}{38.72} &0.9935 &0.0208 &136.26 &\cellcolor{red!20}{22.03}
        \\
        \bottomrule
    \end{tabular}
    }
    }
    \label{table:ST-comparison}
\end{table}

\begin{table*}[!t]
\vspace{-0.1in}
    \centering
    \caption{\textbf{Ablation studies on \# of Gaussians $K$ in each seed.}}
    \vspace{-0.1in}
    \resizebox{\linewidth}{!}{
    \begin{tabular}{ccccccccccc}
        \toprule
        \multirow{2}{*}{$K$} & \multicolumn{5}{c}{FGF2} &\multicolumn{5}{c}{ICB}\\
        \cmidrule(lr){2-6} \cmidrule(lr){7-11}
        & PSNR$\uparrow$ &SSIM$\uparrow$ &LPIPS$\downarrow$ &Opt. Time (Min)$\downarrow$ &Size (MB)$\downarrow$ 
        & PSNR$\uparrow$ &SSIM$\uparrow$ &LPIPS$\downarrow$ &Opt. Time (Min)$\downarrow$ &Size (MB)$\downarrow$
        \\
        \midrule
        5 
        &31.29 &0.9861 &0.0698 &44.27 &18.48
        &35.03 &0.9841 &0.0626 &44.70 &13.64 
        \\
        10 
        &31.24 &0.9863 &0.0731 &48.43 &16.33
        &35.27 &0.9853 &0.0575 &44.75 &12.30
        \\
        15 
        &30.61 &0.9836 &0.0828 &55.33 &15.32
        &34.88 &0.9842 &0.0599 &50.60 &11.48
        \\
        20 
        &30.62 &0.9834 &0.0809 &56.02 &14.83
        &34.57 &0.9824 &0.0668 &53.37 &10.87
        \\
        \bottomrule
    \end{tabular}
    }
    \label{table:ablation_k}
\end{table*}

\noindent{\bf Implementation Details.}
\label{sec:implementation_detail}
Our SGI implementation builds on the 2D Gaussian CUDA kernels~\cite{Zhang-ECCV24,Zhu-AAAI25} and the open source library \texttt{gsplat}~\cite{Ye-JMLR25}.
All our models are trained and tested on NVIDIA A10 GPUs (24 GB memory). 
All Gaussian-based methods, including our SGI under the low-rate setting, use 3.5 million Gaussians.  
We also report results for a high-rate version of SGI that uses 10 million Gaussians.
Due to the substantial memory consumption of INR-based methods, we adopt random sampling of image coordinates and signal values during training, with a batch size of 32{,}000.  
SGI is optimized using Adam for 15,000 steps, with loss weight $\lambda = 0.001$ and pyramid level $M=3$.
Other hyperparameters follow the official implementations of the respective baselines unless otherwise specified. \textit{More implementation details are in the Appendix.}

\subsection{Comparison with Baselines}
\label{sec:comparison}
As shown in Table~\ref{table:GF2-comparison}, when evaluated on FGF2 and ICB, SGI significant reduces storage size compared to both INR and Gaussian-based baselines, while maintaining or surpassing image fidelity in PSNR, SSIM, and LPIPS. In the low-rate setting (3.5M Gaussians), SGI achieves a strong trade-off between accuracy and compactness, outperforming the prior 2D Gaussians baseline GaussianImage by a large margin on both datasets.
In the high-rate configuration (10M Gaussians), SGI achieves the best overall fidelity across all metrics, outperforming the second-best 3DGS with much lower storage cost.
In terms of optimization efficiency, INR method SIREN suffer from slow convergence due to patch-wise optimization imposed by limited GPU memory, while 3DGS, Scaffold-GS and HAC incur significantly longer optimization times due to the complexity of 3D Gaussian rendering. Compared to the next-fastest I-NGP, our SGI achieves both higher fidelity and lower storage cost. This optimization efficiency is particularly important, as image representation requires optimization a separate model for each image.
Figure~\ref{fig:GF2-comprasion} further demonstrates that SGI better preserves fine-grained textures and high-frequency details compared to existing baselines.

The results on biomedical dataset STimage are presented in Table~\ref{table:ST-comparison}. 
Scaffold-GS leverages 3D Gaussian primitives to represent images and optimize MLPs during training, resulting in out-of-memory (OOM) issues when facing the STimage dataset, which has a higher resolution than FGF2 dataset.
Therefore, we exclude Scaffold-GS~\cite{Lu-CVPR24} from this evaluation. 
In the low-rate setting, SGI achieves competitive or superior fidelity compared to other baselines while using the least optimization time and the smallest model size. In the high-rate setting, SGI obtains the highest PSNR across all methods, demonstrating that increased Gaussian capacity leads to further fidelity improvement while maintaining a favorable balance between efficiency and compactness. 
We show the qualitative comparisons on STimage in the Appendix due to page limit.

\begin{table}[t]
    \centering
    \caption{{\bf Ablation studies on $\lambda$ in $L$.}}
    \label{table:ablation_lambda}
    \vspace{-0.1in}
    \scalebox{0.75}{
    \begin{tabular}{lcccc} 
        \toprule
        \multirow{2}{*}{$\lambda$} & \multicolumn{2}{c}{FGF2} & \multicolumn{2}{c}{ICB} \\
        \cmidrule(lr){2-3} \cmidrule(lr){4-5}
        & PSNR$\uparrow$ & Size (MB)$\downarrow$ & PSNR$\uparrow$ & Size (MB)$\downarrow$ \\
        \midrule
        0       & 32.36 & 104.08 & 36.13 & 102.35 \\
        0.0005  & 31.30 & 18.93 & 35.48 & 14.90 \\
        0.001   & 31.24 & 16.33 & 35.27 & 12.30 \\
        0.003   & 30.21 & 11.93 & 33.51 & 8.34  \\
        \bottomrule
    \end{tabular}
    }
    \vspace{-0.1in}
\end{table}

\begin{table}[t]
    \centering
    \caption{{\bf Ablation studies on multi-scale fitting.}}
    \label{table:ablation_multi_scale_fitting}
    \vspace{-0.1in}
    \scalebox{0.75}{
    \begin{tabular}{ccccc}
        \toprule
        \multirow{2}{*}{$M$} & \multicolumn{2}{c}{FGF2} & \multicolumn{2}{c}{ICB} \\
        \cmidrule(lr){2-3} \cmidrule(lr){4-5}
        & PSNR$\uparrow$ & Opt. Time (Min)$\downarrow$ & PSNR$\uparrow$ & Opt. Time (Min)$\downarrow$ \\
        \midrule
        1 & 30.58 & 71.59 & 35.02 & 70.63 \\
        2 & 30.60 & 55.96 & 35.06 & 49.90 \\
        3 & 31.24 & 48.43 & 35.27 & 44.75 \\
        4 & 30.56 & 42.27 & 33.78 & 37.67 \\
        \bottomrule
    \end{tabular}
    }
    \vspace{-0.1in}
\end{table}

\begin{table}[bt]
    \centering
    \caption{\textbf{The storage size (MB) of each component} of our method evaluated on the FGF2 dataset.}
    \small 
    \vspace{-0.1in}
    \resizebox{\linewidth}{!}{
    {
    \begin{tabular}{cccccccc}
        \toprule
      Number of Gaussians  &$\boldsymbol{x_a}$ &$\boldsymbol{f_a}$ &$\{\boldsymbol{s_o},\boldsymbol{s_a}\}$ &$\boldsymbol{\delta}$ &$\mathcal{H}$ &MLPs &Total Size
      \\
        \midrule
        3.5e6 &0.9995 &7.8689 &1.5427 &5.7646 &0.0293 &0.1273 &16.3324 \\
        7.0e6 &1.8360 &15.5719 &2.4960 &11.0664 &0.0292 &0.1273 &31.1268\\
        1.0e7 &2.4442 &20.5720 &3.3549 &15.2157 &0.0298 &0.1273 &41.7439\\
        \bottomrule
    \end{tabular}
    }
    \label{table:component_size}
    \vspace{-0.25in}
    }
\end{table}

\subsection{Ablation Studies and Discussions}
\label{sec:ablation}

\begin{figure*}[t]
 \begin{center}
 $\begin{array}{c@{\hspace{0.025in}}c@{\hspace{0.025in}}c}
 \includegraphics[width=0.32\linewidth]{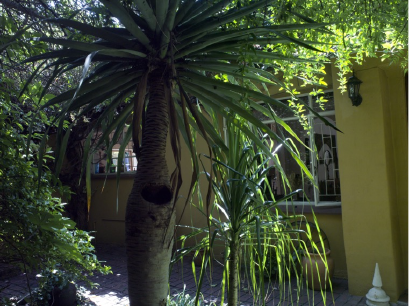}&
 \includegraphics[width=0.32\linewidth]{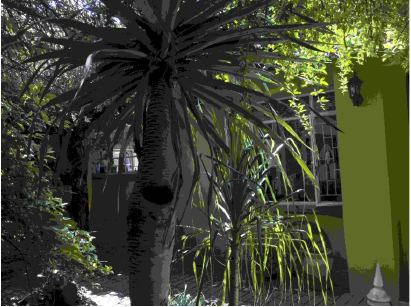}&
 \includegraphics[width=0.32\linewidth]{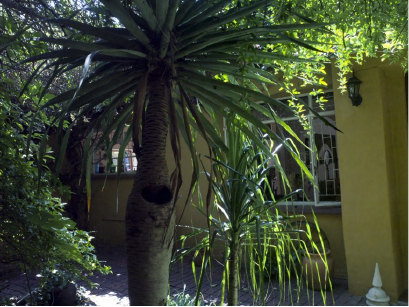} \\

\mbox{\smallfont } & 
\mbox{\smallfont Bpp=0.296, PSNR=22.77dB}&
\mbox{\smallfont Bpp=0.245, PSNR=26.09dB}\\
\mbox{\smallfont (a): GT} &
\mbox{\smallfont (b): JPEG}&
\mbox{\smallfont (c): SGI}\\

\end{array}$
\end{center}
\vspace{-.2in} 
\caption{{\bf Visual comparison with traditional image codec method JPEG on the ICB dataset at low Bpp.}
We report the bit per pixel (bpp) and PSNR (dB) for each method below the visualizations.}
\label{fig:low_bit_vis}
\vspace{-.15in}
\end{figure*}

\begin{figure}[th]
\centering
\includegraphics[width=0.95\linewidth]{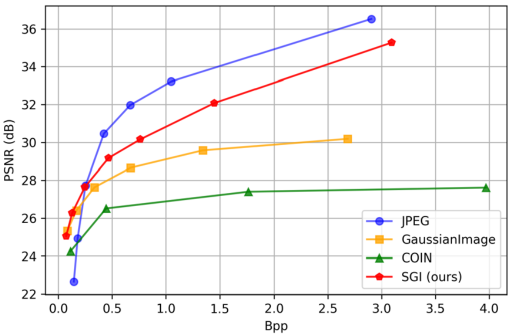}
\caption{
{\bf Rate-distortion curves} of our approach and different compression baselines on ICB dataset in PSNR. 
}
\vspace{-.2in}
\label{fig:rd_curve}
\end{figure}

\noindent {\bf Evaluation of the Number of Gaussians per Seed.}  
To assess the effectiveness of our \emph{seed-based neural Gaussians}, we evaluate SGI on FGF2 and ICB by varying the number of Gaussians $K$ assigned to each seed, as shown in Table~\ref{table:ablation_k}.  
To maintain a consistent total number of Gaussians, we adjust the number of seeds accordingly.  
The dimension $D$ of the seed feature $\boldsymbol{f_a}$ and the latent size of the MLPs are also adapted based on $K$.  
As observed in the table, larger $K$ values lead to smaller overall model sizes, as more Gaussian attributes are compactly encoded within each seed's feature.
However, when $K$ becomes too large, a slight drop in fidelity and an increase in optimization time are observed due to the larger MLP size and seed features.
We choose $K = 10$ in this paper as a trade-off between representation quality and model compactness.

\noindent{\bf Ablation on the Hyperparameter $\lambda$ in $L$.}
We evaluate the impact of the rate-distortion trade-off parameter $\lambda$ in the loss function $L$ in Table~\ref{table:ablation_lambda}. When $\lambda = 0$, the entropy coding is completely deprecated, leaving only the seed-based 2D neural Gaussians.
Notably, using seed-based 2D neural Gaussians alone yields less redundancy reduction than in 3D cases, highlighting the importance of our entropy modeling in achieving effective 2D Gaussian compression. 
As $\lambda$ increases, the storage size reduces due to improved entropy modeling, but overly large values may slightly degrade reconstruction quality. We empirically set $\lambda = 0.001$.

\noindent{\bf Ablation on Multi-scale Fitting.}
Table~\ref{table:ablation_multi_scale_fitting} shows the optimization efficiency and representation quality of SGI under different numbers of Gaussian pyramid levels.
We exclude the model size metric here as \emph{multi-scale fitting} theoretically will not affect the final model size. 
With the total number of optimization steps fixed, we report the actual optimization time.
The results show that our \emph{multi-scale fitting} significantly accelerates optimization while maintaining or even improving image fidelity.
When $M = 4$, we observe a slight PSNR drop, as the optimization budget is distributed across more levels, leaving fewer steps for the final high-resolution stage.
We thus use $M = 3$ in main experiments.

\noindent{\bf Component-wise Storage Analysis.}
Table~\ref{table:component_size} breaks down the storage cost of each component in SGI under different numbers of Gaussians.  
The hash grid $\mathcal{H}$ is stored in a compact binary format, while the MLPs are lightweight by design, contributing only a small number of parameters.
Together, they impose minimal storage overhead and scale well with increasing Gaussians, enabling efficient high-resolution image representation.

\noindent {\bf Comparisons with Image Compression Baselines.} Figure~\ref{fig:rd_curve} presents the rate-distortion (RD) curves of various codecs on the ICB dataset. Our method significantly outperforms the 2DGS-based method GaussianImage and the INR-based method COIN by a large margin. 
It achieves competitive performance across the entire bit-rate range and outperforms the conventional JPEG codec in the low bit-rate regime.
As illustrated in Figure~\ref{fig:low_bit_vis}, JPEG suffers from severe color shifts and visual artifacts at low bitrates, resulting in noticeably degraded PSNR. In contrast, our SGI method produces visually faithful reconstructions with superior PSNR while using fewer bits, demonstrating the potential of structured 2D Gaussian representations for next-generation image compression.

\section{Conclusion}
\label{sec:limitation}
In this work, we present SGI, an efficient and compact framework for high-resolution image representation using structured 2D Gaussians. 
By organizing unstructured Gaussian primitives under seeds and decoding their attributes through shared lightweight MLPs, our \emph{seed-based 2D neural Gaussians} establish a structured representation that enables further compression. Building on this structure, we introduce an \emph{entropy coding scheme} with a learnable context model guided by a binary hash grid, enabling effective probabilistic modeling of seed attributes. 
To mitigate the optimization overhead introduced by the large number of Gaussians and entropy estimation, we develop a \emph{multi-scale fitting strategy} that progressively refines the representation from coarse to fine, accelerating optimization and enhancing its stability. 
Experiments on three megapixel-scale datasets show that SGI achieves competitive or superior fidelity with significantly reduced model size and optimization time, highlighting its potential as an effective solution for efficient and compact large-scale image representation.

 \section*{Acknowledgment}
This work was supported in part by the U.S. National Science Foundation under grants IIS-2101696, OAC-2104158, and IIS-2401144. J. Chen was supported by the Bundesministerium für Forschung, Technologie und Raumfahrt, BMFTR under the funding reference 161L0272 and the “Ministerium für Kultur und Wissenschaft des Landes Nordrhein-Westfalen” and “Der Regierende Bürgermeister von Berlin, Senatskanzlei Wissenschaft und Forschung.”
{
    \small
    \bibliographystyle{ieeenat_fullname}
    \bibliography{main}
}

\clearpage
\appendix
\setcounter{figure}{0}
\setcounter{table}{0}
\renewcommand{\thefigure}{A\arabic{figure}}
\renewcommand{\thetable}{A\arabic{table}}
\setcounter{page}{1}
\maketitlesupplementary

\section{More Implementation Details}
The learning rates used for different components are listed in Table~\ref{table:lr}. 
We apply learning rate decay to all components except for the seed features and scaling parameters.
For optimization, we adopt the Adam optimizer with default momentum parameters $\beta_1 = 0.9$, $\beta_2 = 0.999$, and set the numerical stability term to $\epsilon = 1 \times 10^{-15}$. 
Each of the three MLPs (i.e., $\text{MLP}_c, \text{MLP}_{\Sigma}, \text{MLP}_p$) is composed of two fully connected (linear) layers with a ReLU activation function applied between them.

\begin{table}[ht]
\centering
\captionof{table}{Details of the learning rates in SGI.}
\begin{tabular}{lc}
    \toprule    
    Component & Learning rate \\
    \midrule
    Seed position $\boldsymbol{x_a}$ & 0 \\
    Offsets $\boldsymbol{\delta}$ & 0.01 \\
    Seed feature $\boldsymbol{f_a}$ & 0.0075 \\
    Scaling $\{\boldsymbol{s_o}, \boldsymbol{s_a}\}$ & 0.007 \\
    $\text{MLP}_c$ & 0.008 \\
    $\text{MLP}_{\Sigma}$ & 0.004 \\
    $\text{MLP}_p$ & 0.005 \\
    Hash grid $\mathcal{H}$ & 0.005 \\
    \bottomrule
\end{tabular}
\label{table:lr}
\end{table}

\begin{algorithm}[thbp]
\caption{Structured Gaussian Image (SGI)}
\label{alg:sgi}
\textbf{Input:} Image $I$ \\
\textbf{Hyperparameters:} Pyramid levels $M$, seed number $N$, Gaussians per seed $K$
\begin{algorithmic}[1]
\State Initialize seeds $\mathbb{A} = \{\boldsymbol{x}_a^{(i)}, \mathcal{A}^{(i)}\}_{i=0}^{N-1}$
\State Initialize learnable parameters $\theta = \{\theta_c, \theta_\Sigma, \theta_p, \theta_{\mathcal{H}}\}$ for MLPs and binary hash grid $\mathcal{H}$
\For{$l = M{-}1, \dots, 0$}
    \State Downsample $I$ to $I_l$
    \State Initialize or adapt $\mathbb{A}^{(l)}, \theta^{(l)}$ from level $l{+}1$
    \Comment{Eq.~\eqref{eq:update_multi_scale}}
    \Repeat
        \State Inject noise to $\mathbb{A}^{(l)}$ for quantization-aware training
        \Comment{Eq.~\eqref{eq:quantization}}
        \State Decode Gaussians: $\{\boldsymbol{\mu}^{(k)}, \boldsymbol{\Sigma}^{(k)},
        \boldsymbol{c}^{(k)}\}_{k=0}^{K-1}$
        \Comment{Section~\ref{sec:neural_gaussian}}
        \State Render image $\hat{I}_l$, compute $L_\text{img}(I_l, \hat{I}_l)$
        \State Estimate seed attributes probability via $\text{MLP}_p$ and $\mathcal{H}$ \Comment{Eqs.~\eqref{context_model_eq}, \eqref{prob}}
        \State Compute $L_{\text{entropy}}$ and $L_{\text{hash}}$
        \Comment{Eqs.~\eqref{eq:entropy}, ~\eqref{eq:hash}}
        \State Update $\mathbb{A}^{(l)}, \theta^{(l)}$ using total loss
        \Comment{Eq.~\eqref{eq:overall_loss}}
    \Until{convergence}
\EndFor
\State Compress seed positions via GPCC
\State Entropy-code seed attributes and hash grid via arithmetic coding
\State \Return seed positions, entropy-code attributes, and model parameters $\theta$
\end{algorithmic}
\end{algorithm}

\section{Pseudocode}
Algorithm~\ref{alg:sgi} presents the full training pipeline of SGI. The input image is first downsampled to form a multi-level Gaussian pyramid, enabling coarse-to-fine optimization over $M$ levels. At each level $l$, seed parameters and MLP weights are either initialized or adapted from the coarser level $l{+}1$.
During training, we inject quantization-aware noise into seed attributes and decode their corresponding Gaussian primitives using shared MLPs. 
The image reconstruction loss $L_{\text{img}}$ is computed with the SGI rendered images and ground truth images.
In parallel, a context model guided by a binary hash grid estimates the probability distribution of seed attributes for entropy coding.
After optimization, seed positions are compressed using GPCC, and seed attributes along with the hash grid are entropy-coded using arithmetic coding.

\section{Additional Experimental Results}

\begin{table}[hth]
    \centering
    \caption{\textbf{Unsupervised super-resolution performance on a DIV2K sample.} Quantitative results are measured in PSNR (dB).}
    \resizebox{0.8\columnwidth}{!}{
    \begin{tabular}{cccc}
        \toprule
        Method & $\times 2$ & $\times 4$ & $\times 8$
        \\
        \midrule
        Bilinear interpolation 
        &28.40 &27.24  &26.86
        \\
        SGI
        &\textbf{28.88} &\textbf{27.58} &\textbf{27.17}
        \\
        \bottomrule
    \end{tabular}
    }
    \label{table:continuous}
\end{table}

\vspace{-0.02in}
\begin{table}[!htb]
\vspace{-0.1in}
    \centering
    \caption{\textbf{Comparisons on DIV2K.} Optimization time is in minutes, and model size is in MB. \emph{Numbers are averaged per image.}}
    \vspace{-0.1in}
    \resizebox{\linewidth}{!}{
    {
    \begin{tabular}{lcccS[table-format=3.2]S[table-format=3.2]}
        \toprule
       Method & PSNR (dB)$\uparrow$ &SSIM$\uparrow$ &LPIPS$\downarrow$ &\multicolumn{1}{c}{Opt. Time$\downarrow$} &\multicolumn{1}{c}{Size$\downarrow$}
       \\
        \midrule
        SIREN (NeurIPS'20)
        &24.71 &0.6432 &0.5019 &28.37 &0.45
        \\
        GaussianImage (ECCV'24) 
        &35.22 &0.9327 &0.1469 &21.91 &3.07
        \\
        LIG (AAAI'25)
        &44.87 &0.9904 &0.0113 &9.97 &15.26
        \\
        \midrule
        Our SGI (low-rate) 
        &28.69 &0.8206 &0.3163 &3.31 &0.33
        \\
        Our SGI (med-rate) 
        &37.72 &0.9643 &0.0778 &6.80 &1.82
        \\
        Our SGI (high-rate) 
        &44.03 &0.9872 &0.0298 &15.64 &5.40
        \\
        \bottomrule
    \end{tabular}
    }
    }
    \label{table:div2k}
    \vspace{-0.2in}
\end{table}

\label{appendix:continuity}
\subsection{Continuous Attribute of SGI Image Representation} 
We investigate the continuous nature of the SGI-based image representation in Table~\ref{table:continuous}. Specifically, we downsample a 2K image from the DIV2K dataset~\cite{Ignatov-ECCV19} to $2\times$, $4\times$, and $8\times$ resolutions. Our SGI model is trained using 50,000 Gaussians only on the $8\times$ downsampled image. The learned low-resolution representation is then directly used to generate higher-resolution outputs, leveraging the continuous property of the SGI representation. SGI achieves approximately a 0.5~dB gain in PSNR over bilinear interpolation on the low-resolution image. These results not only serve as the foundation for our multi-scale fitting strategy, but also demonstrate that the 2D Gaussian representation used in this work enables arbitrary-scale super-resolution, similar to~\cite{Chen-CVIU, Hu-AAAI25}.

\subsection{Evaluations on the DIV2K Dataset}
While our main focus is on large, megapixel-scale images, where prior Gaussian- and INR-based representations face severe scalability and optimization challenges, the
results on DIV2K~\cite{Ignatov-ECCV19} in Table~\ref{table:div2k} show that SGI also achieves strong RD performance on a standard image compression benchmark.

\begin{figure*}[tb]
\vspace{-.15in}
 \begin{center}
 $\begin{array}{c@{\hspace{0.025in}}c@{\hspace{0.025in}}c@{\hspace{0.025in}}c@{\hspace{0.025in}}c@{\hspace{0.025in}}c}
 \includegraphics[width=0.155\linewidth]{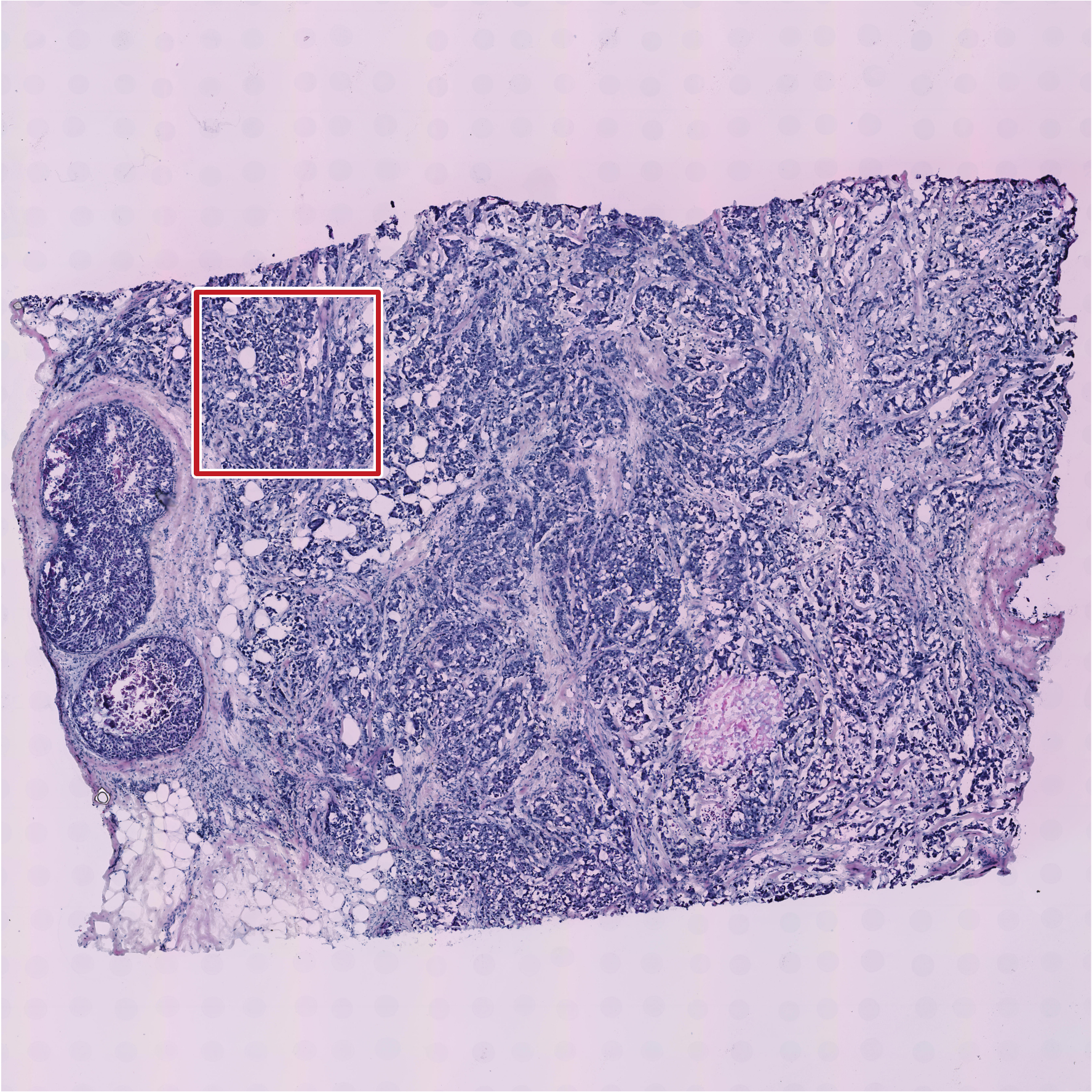}&
 \includegraphics[width=0.155\linewidth]{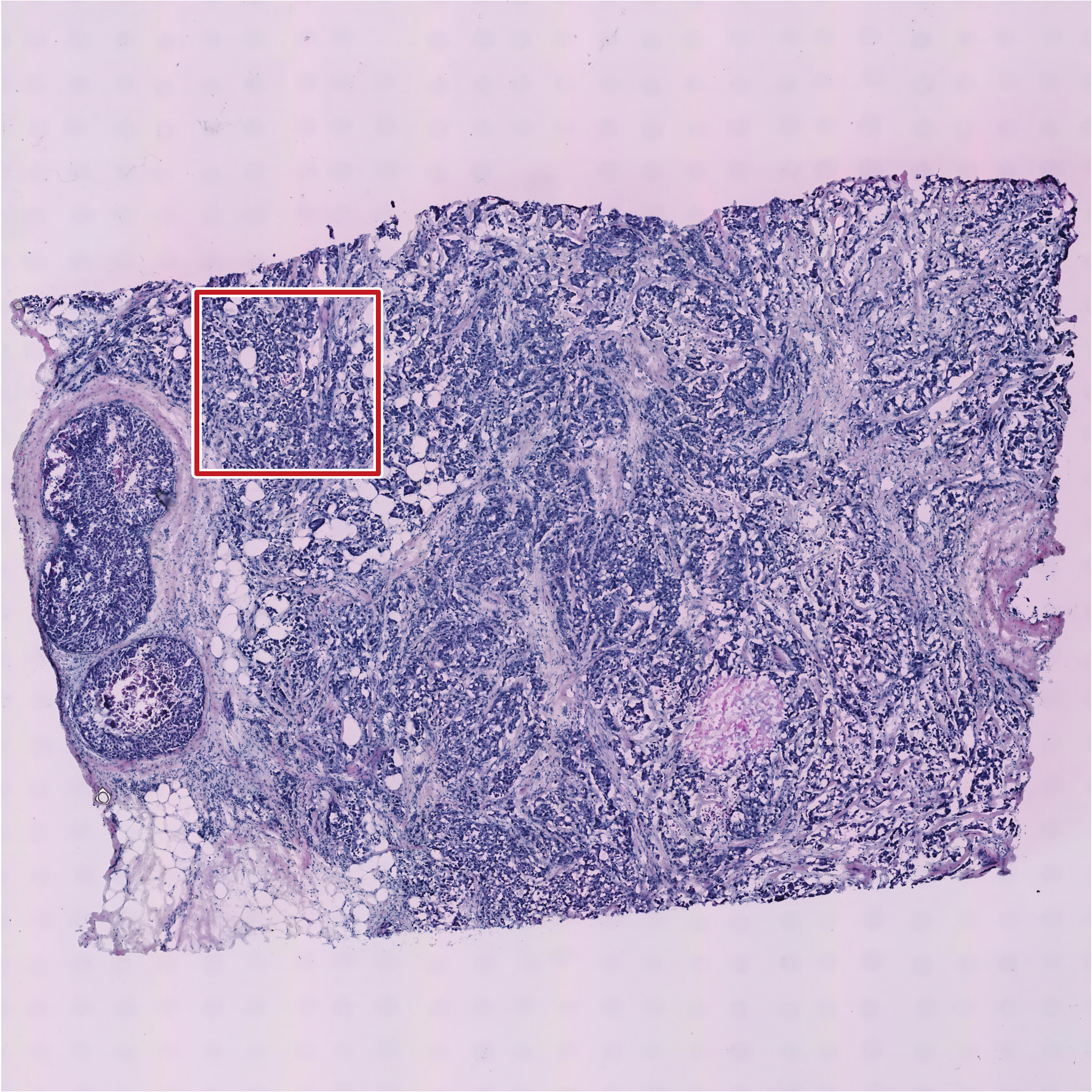}&
 \includegraphics[width=0.155\linewidth]{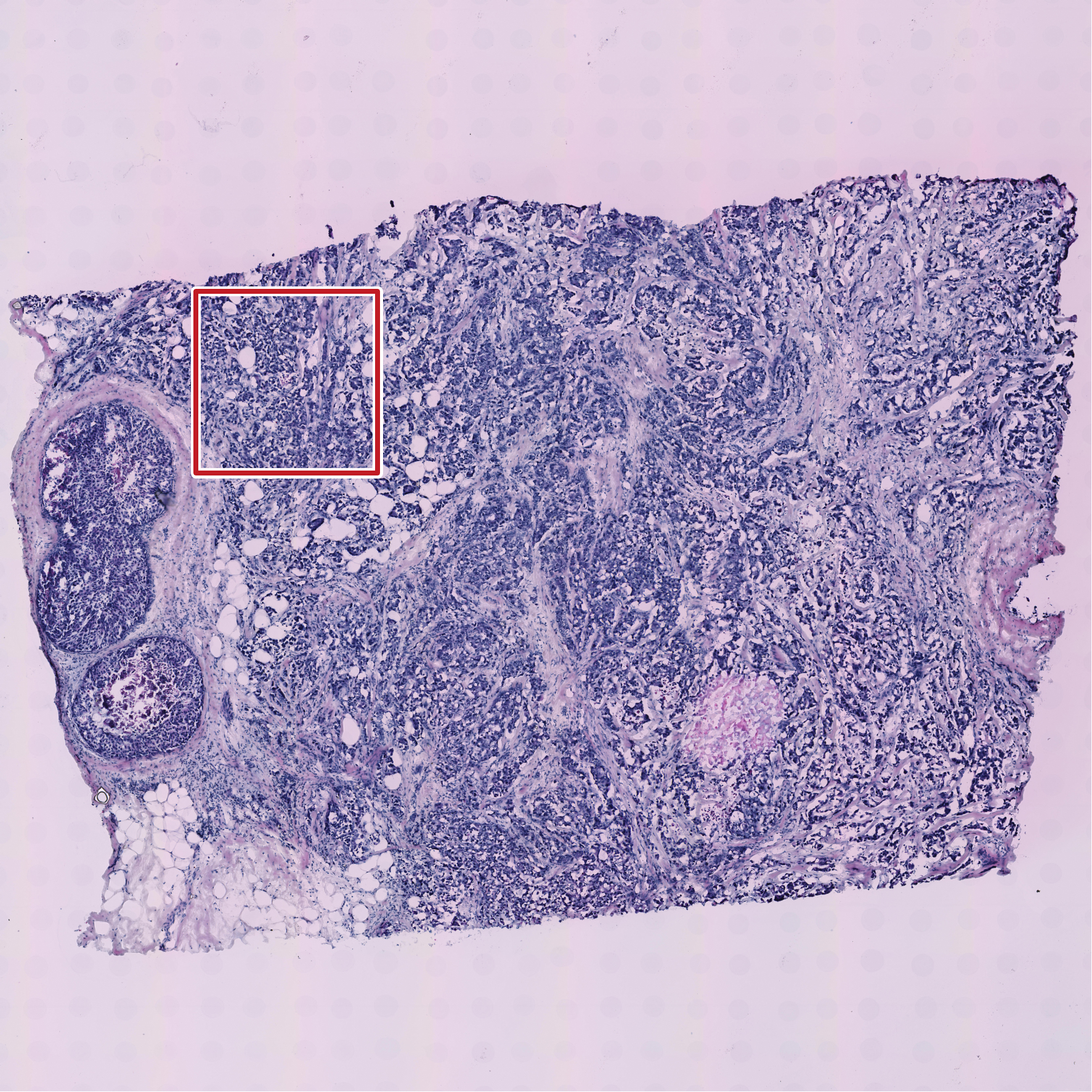}&
 \includegraphics[width=0.155\linewidth]{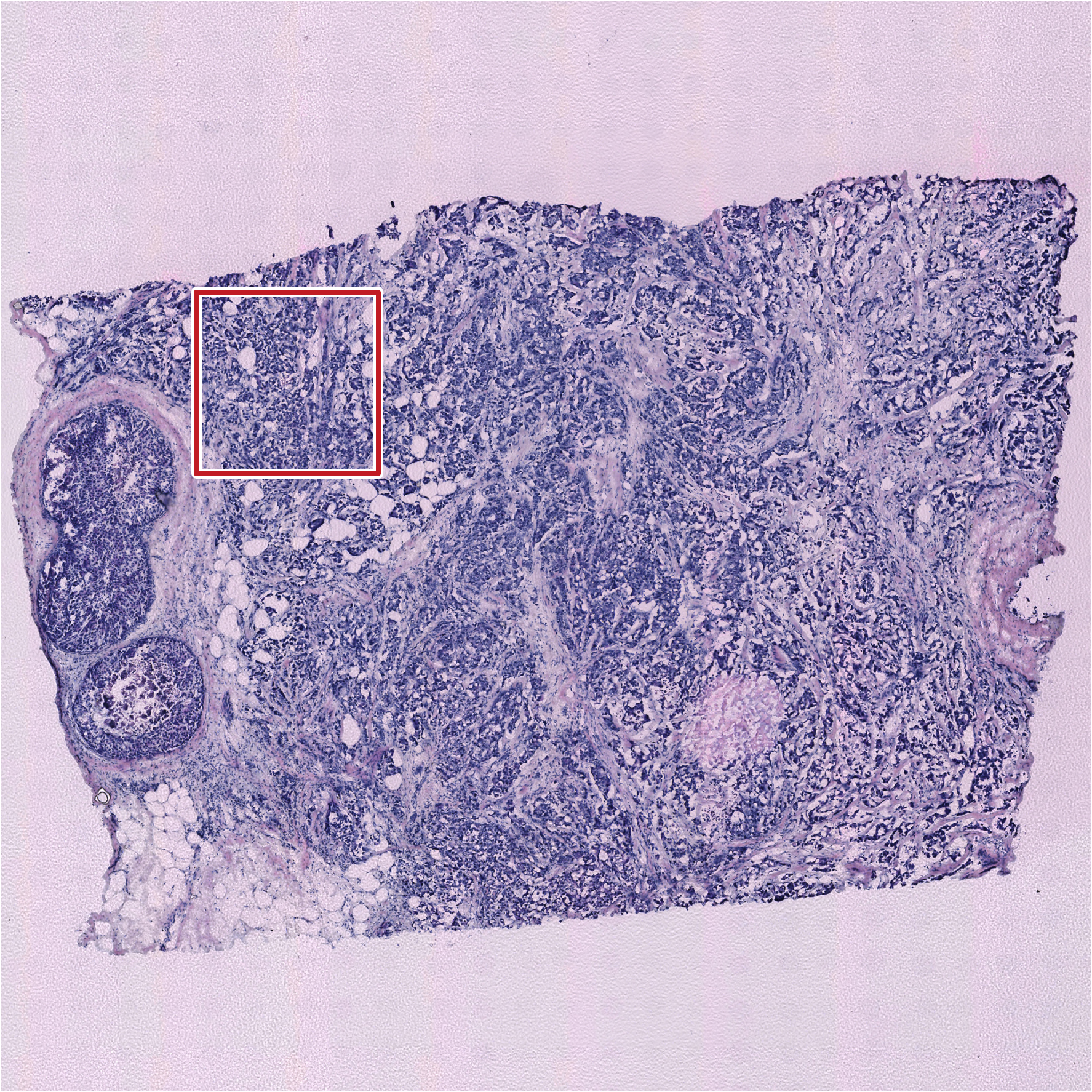}&
 \includegraphics[width=0.155\linewidth]{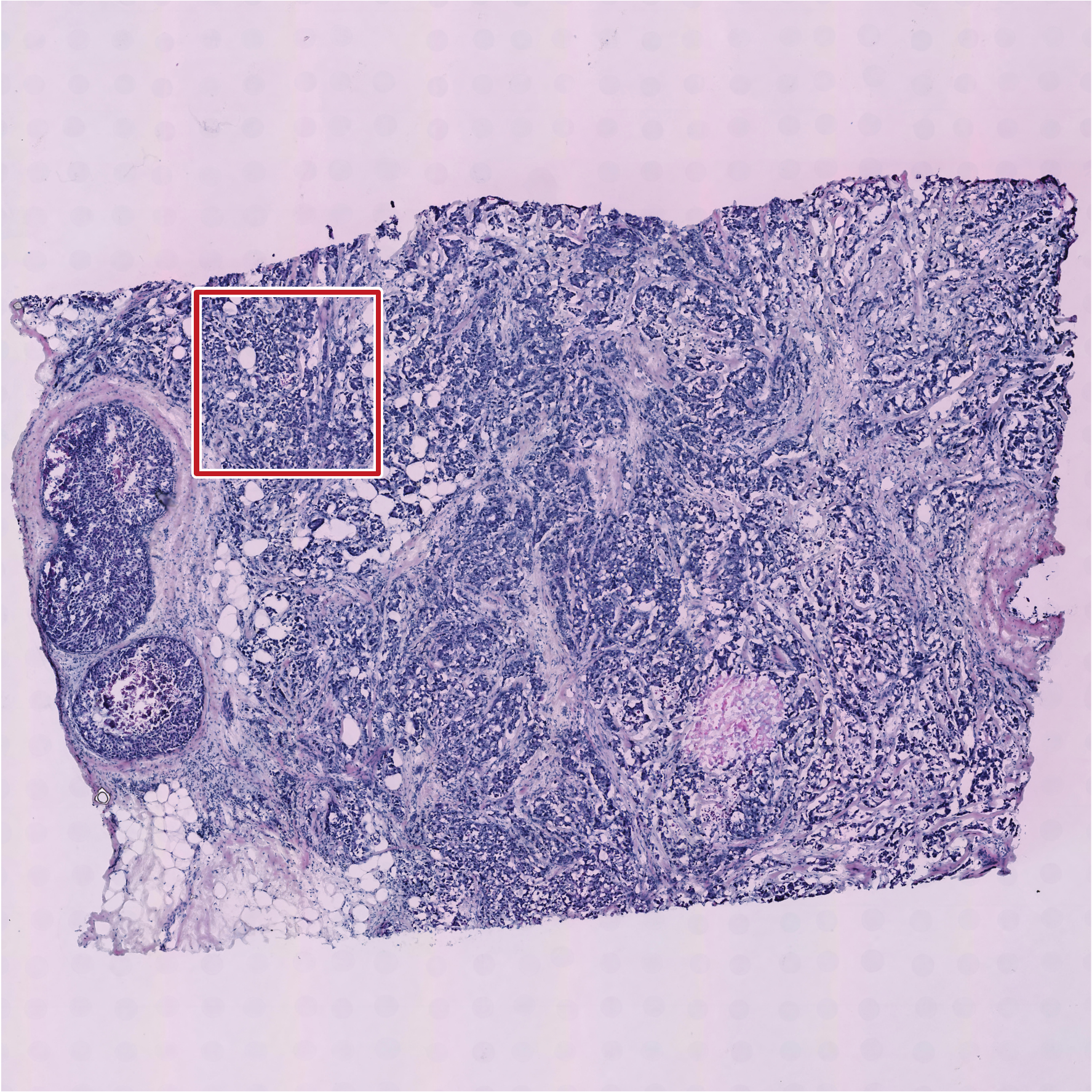}&
 \includegraphics[width=0.155\linewidth]{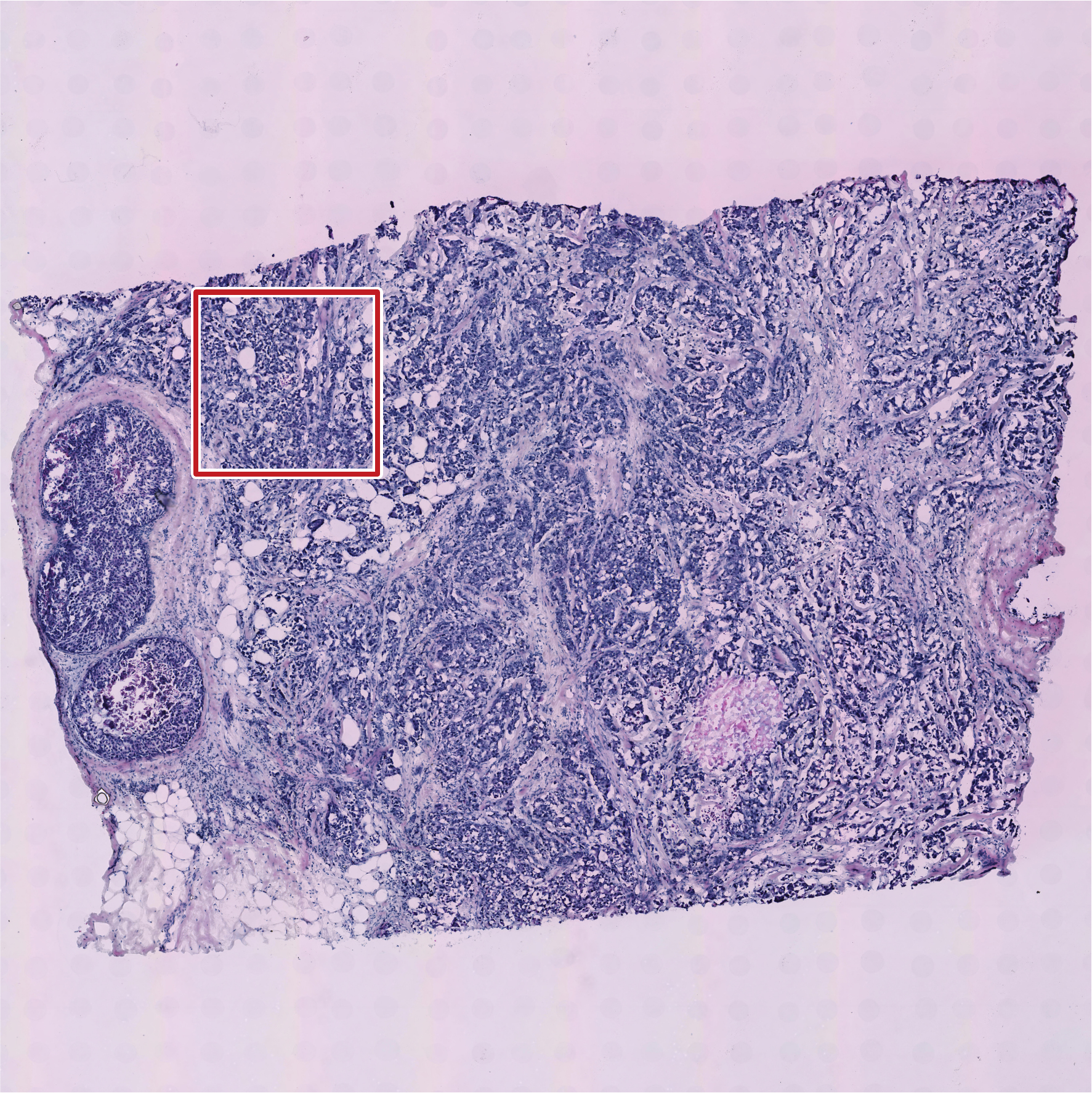}\\
 
 \includegraphics[width=0.155\linewidth]{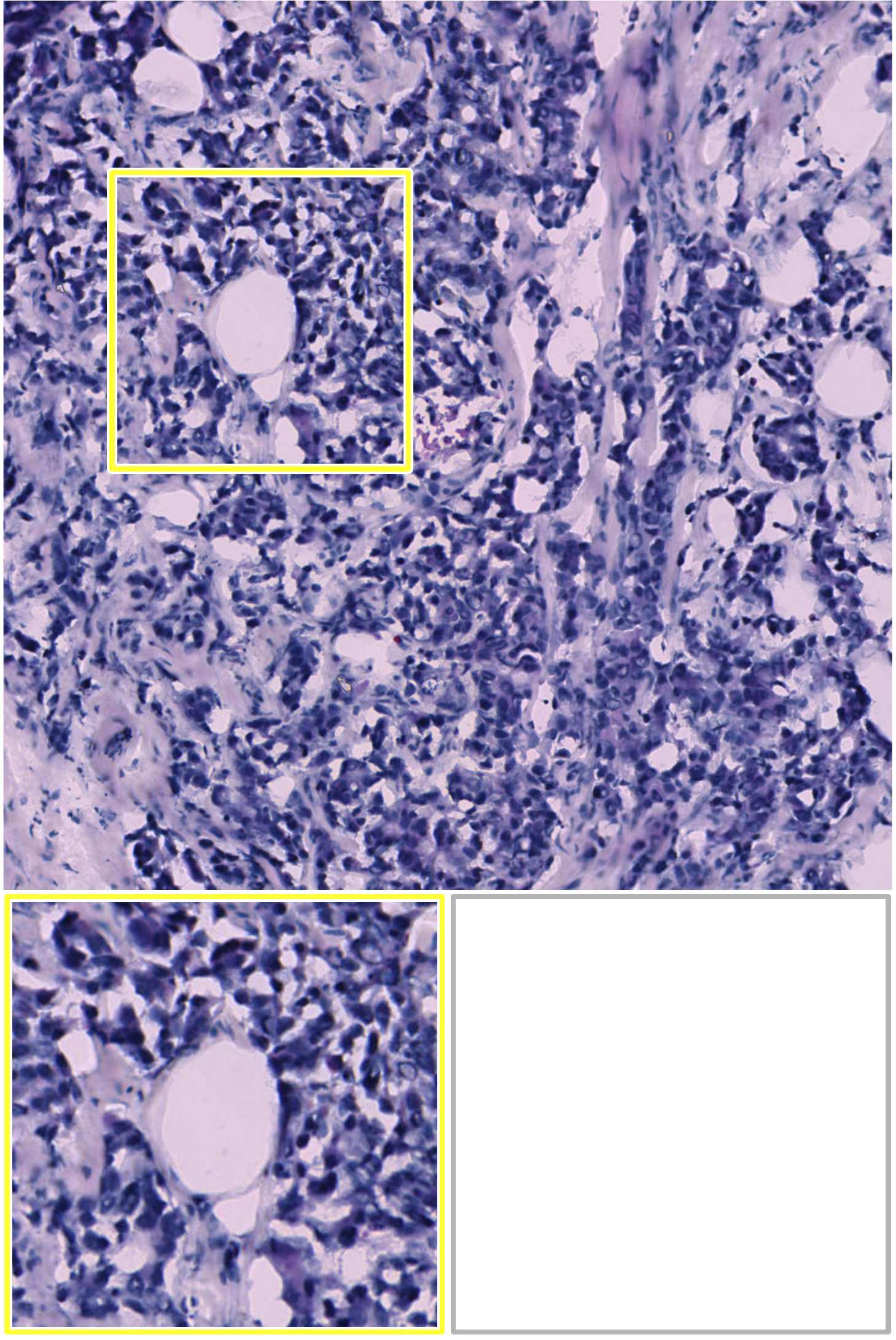}&
 \includegraphics[width=0.155\linewidth]{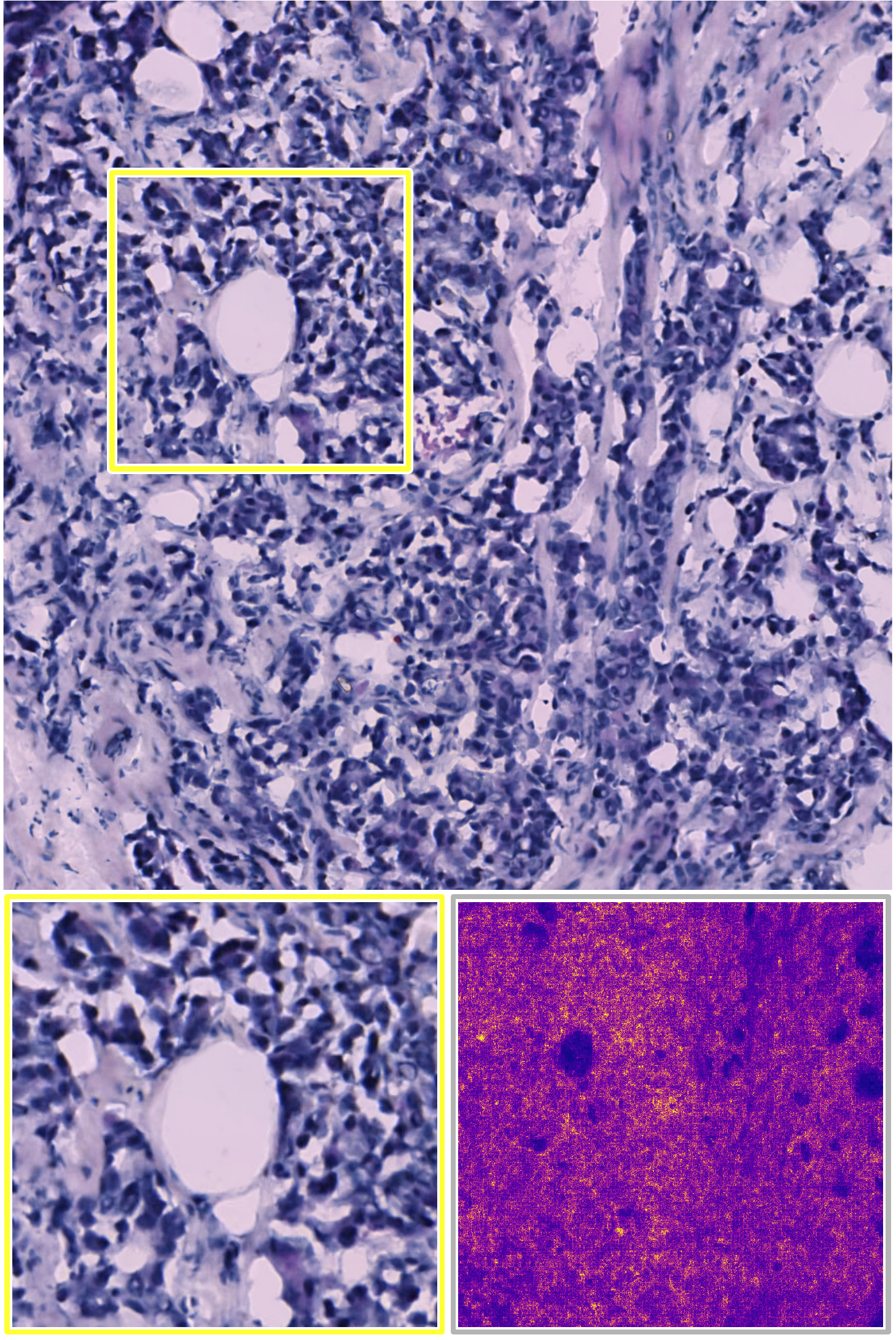}&
 \includegraphics[width=0.155\linewidth]{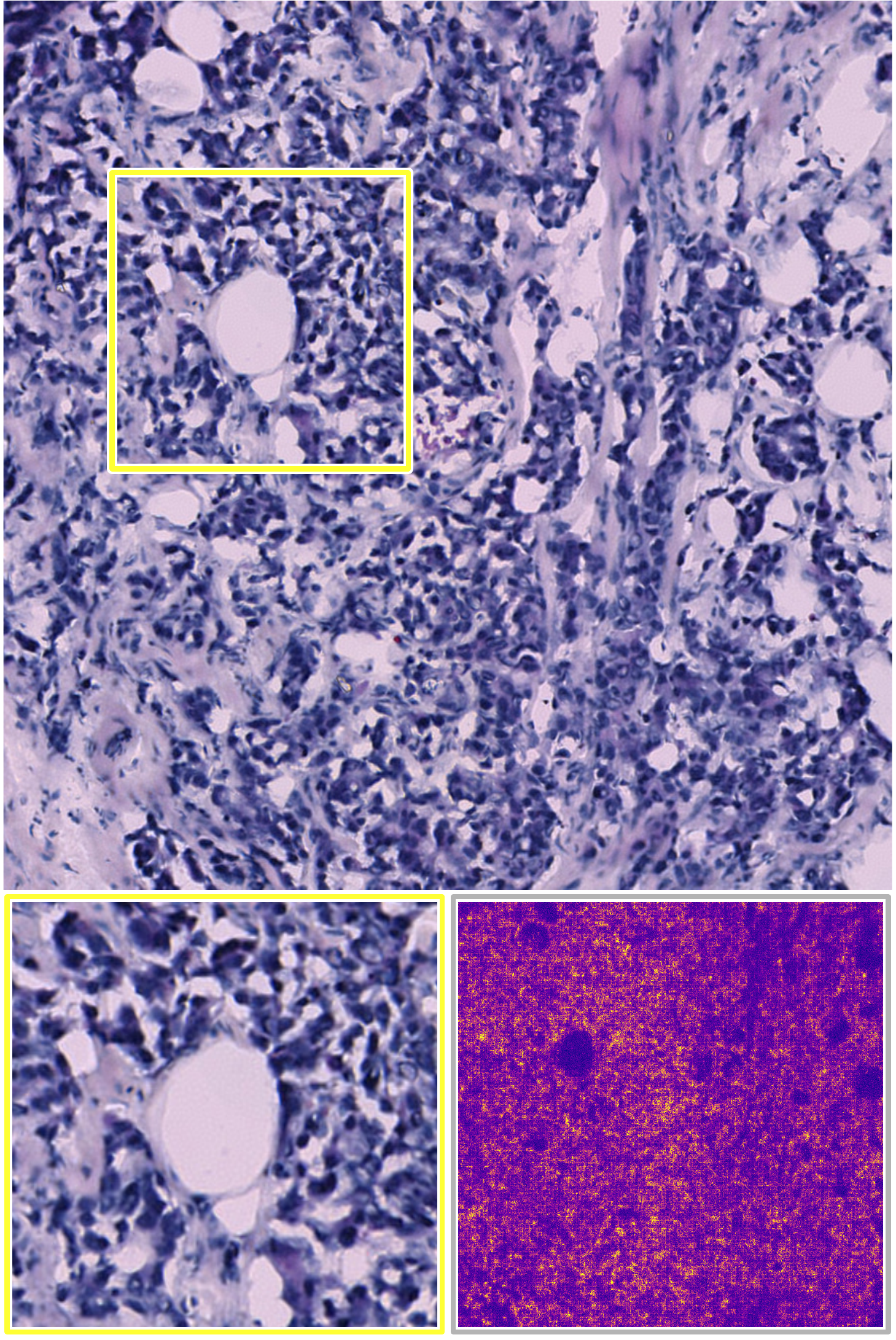}&
 \includegraphics[width=0.155\linewidth]{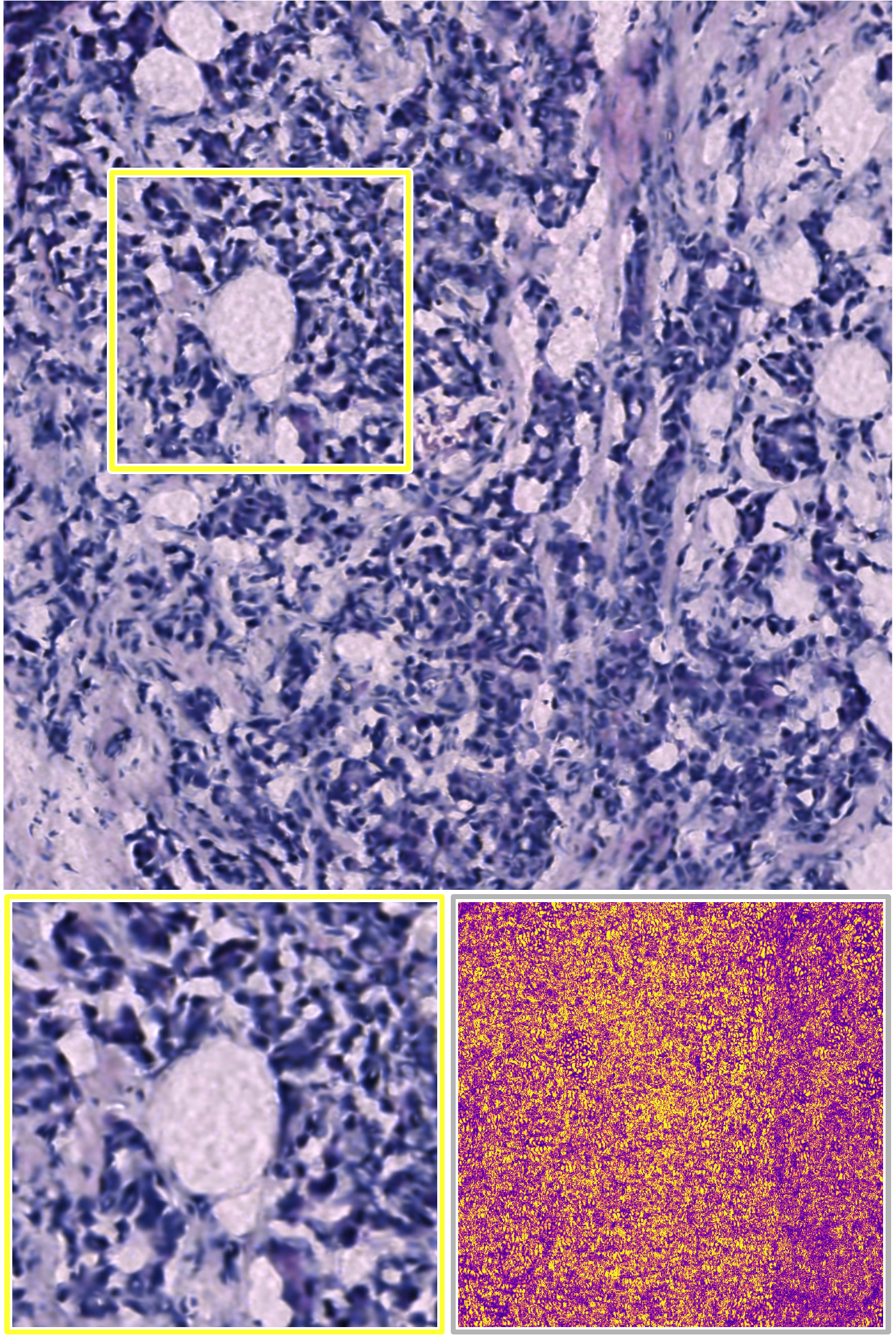}&
 \includegraphics[width=0.155\linewidth]{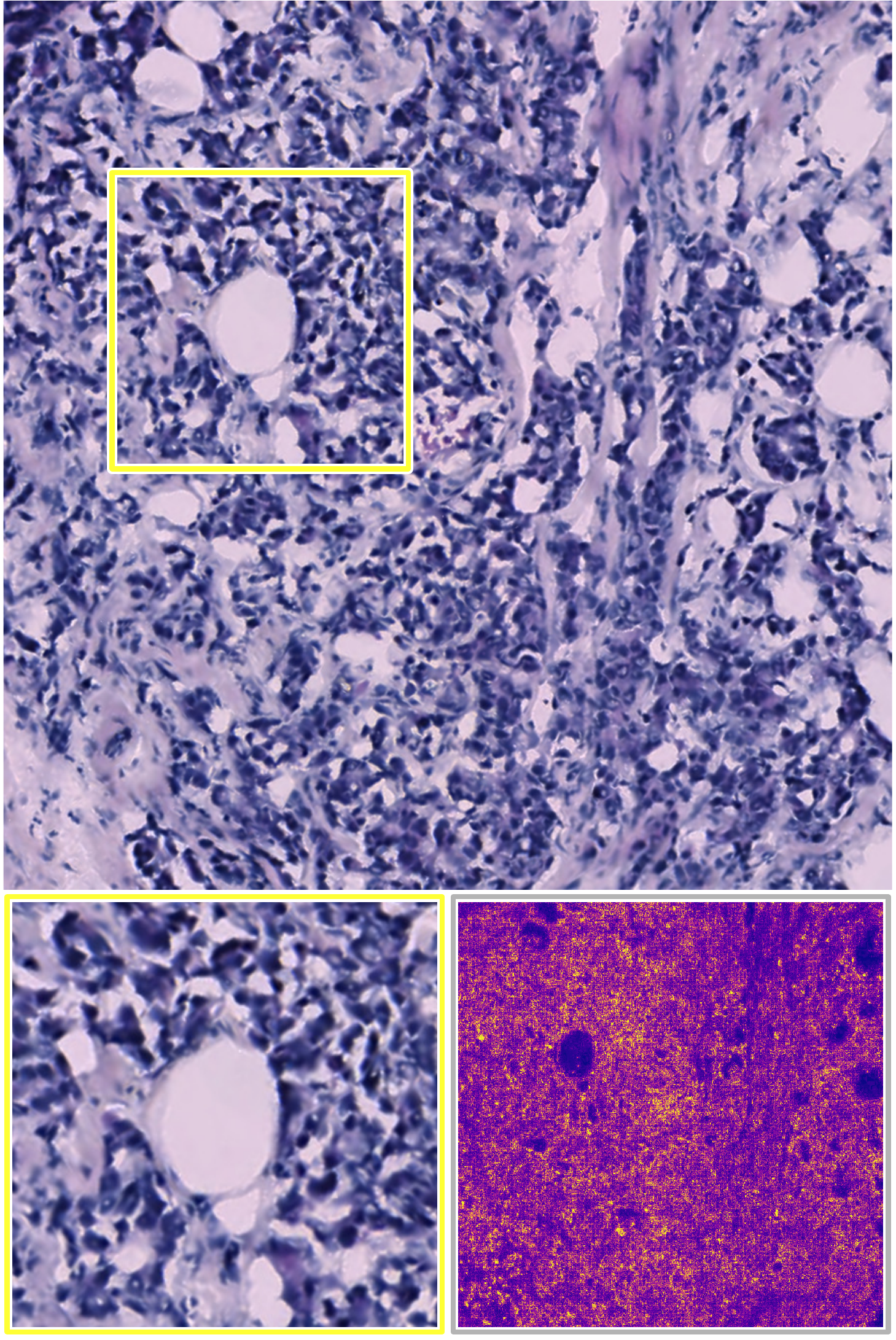}&
 \includegraphics[width=0.155\linewidth]{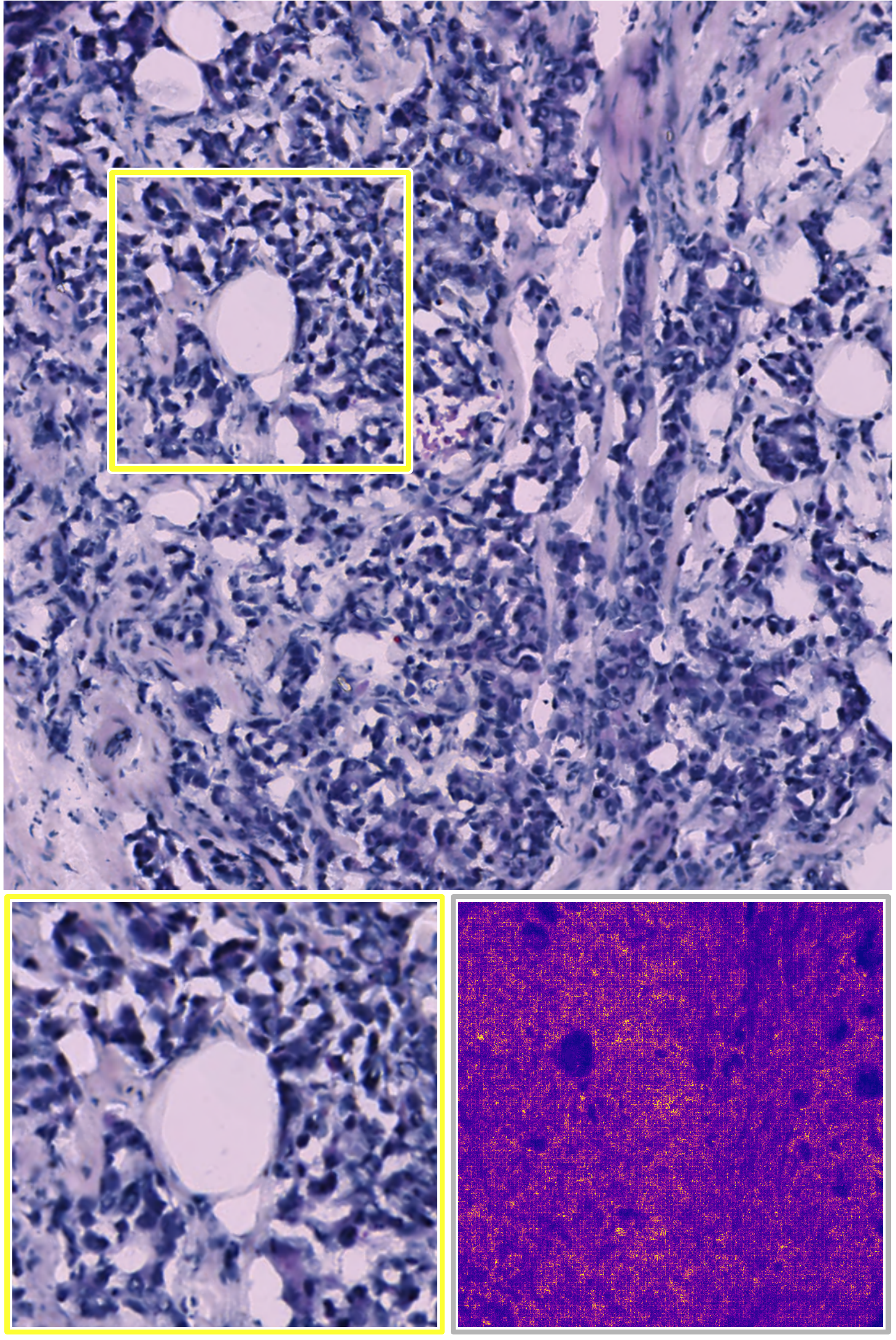}\\

\mbox{\smallfont } & 
\mbox{\smallfont 36.83dB/21.07MB}&
\mbox{\smallfont 35.83dB/106.81MB} &
\mbox{\smallfont 27.28dB/23.37MB}&
\mbox{\smallfont 34.81dB/9.20MB}&
\mbox{\smallfont 38.16dB/15.34MB}\\
\mbox{\smallfont (a): GT} &
\mbox{\smallfont (b): I-NGP}&
\mbox{\smallfont (c): LIG} &
\mbox{\smallfont (d): GaussianImage} &
\mbox{\smallfont (e): SGI (low-rate)}&
\mbox{\smallfont (f): SGI (high-rate) }\\

\end{array}$
\end{center}
\vspace{-.15in} 
\caption{{\bf Visual comparisons on STimage (w/ zoom-in cases and error maps).}
Zoom-in regions highlight perceptual differences. 
The third row shows per-pixel reconstruction error heatmaps, where warmer colors (e.g., yellow) indicate larger deviations from the ground truth.
PSNR (dB) and storage size (MB) for each method are shown below the visualizations.}
\label{fig:STimage-comparison}
\vspace{-.1in}
\end{figure*}

\subsection{Analysis of Inference Time}
Without optimization, arithmetic coding (AC) adds $\sim$80\,ms latency for a 2K image with 50K Gaussians on an A10 GPU.
The total latency ($\sim$88\,ms) remains competitive with GS-based methods and much faster than INR-based methods (e.g., SIREN, $\sim$250\,ms).
Latency can be further reduced via parallel decoding across various attributes using tailored CUDA kernel designs.

\section{Future Directions}

While SGI establishes a robust baseline for seed-based Gaussian image representation, several promising avenues exist to further push the boundaries of compression efficiency and reconstruction fidelity.

\noindent \textbf{Content-Adaptive Representation.} 
Our current framework utilizes a fixed number of Gaussians $K$ per seed to ensure optimization stability and implementation simplicity. However, the SGI architecture is inherently compatible with content-adaptive strategies. A natural extension would be to dynamically optimize $K$ or the primitive allocation based on local texture complexity, allocating more expressive power to high-frequency details while maintaining sparsity in smooth regions. Such directions are orthogonal to our SGI and could be integrated with concurrent ideas, such as those in Image-GS~\cite{Zhang-SIGGRAPH25}, to further enhance fidelity at extremely low bitrates.

\noindent \textbf{Advanced Optimization and Quantization.} 
To maintain a lightweight and efficient training pipeline, SGI currently employs additive uniform noise to simulate quantization. Future research could explore more sophisticated quantization-aware training (QAT) techniques, such as stochastic Gumbel annealing (SGA) or learned rounding-to-nearest estimators.

\end{document}